\documentclass[twoside,11pt,table,usenames,dvipsnames]{article}
\usepackage{jair, theapa, rawfonts}
\usepackage{comment}
\usepackage[implicit=false]{hyperref}

\usepackage{booktabs}

\usepackage{tikz}
\usepackage{tikz-qtree-compat}
\usetikzlibrary{shapes, fit, arrows.meta}
\usepackage{lingmacros}
\usepackage{colortbl} 
\usepackage{tabularx}
\usepackage{flafter} 
\usepackage{float}
\usepackage{graphicx}
\tikzset{>=latex}
\usepackage{footmisc}

\usepackage{xspace} 
\usepackage{fancyvrb} 
\usepackage{caption} 
\usepackage[color=red!5]{todonotes}

\usepackage{xstring}
\usepackage{soul}
\usepackage[normalem]{ulem} 
\usepackage{paralist} 
\usepackage{algpseudocode,algorithm} 
\usepackage{tikz}
\usepackage{tikz-qtree-compat}
\usetikzlibrary{shapes, fit, arrows.meta}
\usepackage{lingmacros}
\usepackage{colortbl} 
\usepackage{tabularx}
\usepackage{flafter} 
\usepackage{adjustbox}

\algnewcommand{\LineComment}[1]{\State \(\triangleright\) #1} 

\algrenewcommand\alglinenumber[1]{\scriptsize #1:} 

\tikzset{>=latex}

\usepackage{xcolor}
\usepackage[skins]{tcolorbox}
\usepackage{hyperref}
\usepackage{csquotes}
\usepackage{booktabs}
\usepackage{subcaption}
\usepackage{comment}
\usepackage{caption}
\usepackage{subcaption}

\newsavebox{\fmbox}

\newenvironment{definitionbox}[1]{\begin{tcolorbox}[title=#1, fonttitle=\bfseries, colback=Peach!5!white, colframe=Peach, colbacktitle=Peach]}{\end{tcolorbox}}
\newenvironment{examplebox}[1]{\begin{tcolorbox}[title=#1, fonttitle=\bfseries, colback=CadetBlue!5!white, colframe=CadetBlue, colbacktitle=CadetBlue]}{\end{tcolorbox}}

\usepackage{multirow}
\usepackage{amsmath}
\usepackage{verbatim}
\usepackage{longtable}

\usepackage{tabularx,lipsum,environ,amsmath,amssymb}

\jairheading{1}{1993}{1-15}{6/91}{9/91}
\ShortHeadings{Discourse-Aware Text Simplification}
{Niklaus, Cetto, Freitas, \& Handschuh}
\firstpageno{1}

\begin{document}

\title{Discourse-Aware Text Simplification:\\ From Complex Sentences to Linked Propositions}

\author{\name Christina Niklaus \email christina.niklaus@unisg.ch \\
       \name Matthias Cetto \email matthias.cetto@unisg.ch \\
       \addr School of Computer Science, University of St.Gallen, St.Gallen, Switzerland
       \AND
       \name André Freitas \email andre.freitas@idiap.ch \\
       \addr Idiap Research Institute, Martigny, Switzerland\\
       Department of Computer Science, University of Manchester, Manchester, UK
       \AND
       \name Siegfried Handschuh \email siegfried.handschuh@unisg.ch \\
       \addr School of Computer Science, University of St.Gallen, St.Gallen, Switzerland\\
       University of Passau, Passau, Germany}


\maketitle

\begin{abstract}

Sentences that present a complex syntax act as a major stumbling block for downstream Natural Language Processing applications whose predictive quality deteriorates with sentence length and complexity. The task of Text Simplification (TS) may remedy this situation. It aims to modify sentences in order to make them easier to process, using a set of rewriting operations, such as reordering, deletion, or splitting. These transformations are executed with the objective of converting the input into a simplified output while preserving its main idea and keeping it grammatically sound. 

State-of-the-art syntactic TS approaches suffer from two major drawbacks: first, they follow a very conservative approach in that they tend to retain the input rather than transforming it, and second, they ignore the cohesive nature of texts, where context spread across clauses or sentences is needed to infer the true meaning of a statement. To address these problems, we present a discourse-aware TS approach that splits and rephrases complex English sentences within the semantic context in which they occur. Based on a linguistically grounded transformation stage that uses clausal and phrasal disembedding mechanisms, complex sentences are transformed into shorter utterances with a simple canonical structure that can be easily analyzed by downstream applications. With sentence splitting, we thus address a TS task that has hardly been explored so far. Moreover, we introduce the notion of minimality in this context, as we aim to decompose source sentences into a set of self-contained minimal semantic units. To avoid breaking down the input into a disjointed sequence of statements that is difficult to interpret because important contextual information is missing, we incorporate the semantic context between the split propositions in the form of hierarchical structures and semantic relationships. In that way, we generate a semantic hierarchy of minimal propositions that leads to a novel representation of complex assertions that puts a semantic layer on top of the simplified sentences. As a proof of concept, we develop a reference TS implementation. 
In a comparative analysis, we demonstrate that our approach outperforms the state of the art in structural TS both in an automatic and a manual analysis.

\end{abstract}

\section{Introduction}
Sentences that present a complex linguistic structure can be \textit{hard to comprehend by human readers}, as well as \textit{difficult to analyze by natural language processing (NLP) applications}. Identifying grammatical complexities in a sentence and transforming them into simpler structures - using a set of text-to-text rewriting operations - is the goal of syntactic Text Simplification (TS). The major types of operations that are used to perform this rewriting step are \textit{deletion} and \textit{sentence splitting}. The goal of the deletion task is to remove parts of a sentence that contain only peripheral information in order to produce an output proposition that is more succinct and contains only the key information of the input. The splitting operation, on the contrary, aims at preserving the complete information of the input. Therefore, it divides a sentence into several shorter components, with each of them presenting a simpler and more regular structure that is easier to process by both humans and machines and may support a faster generalization in machine learning (ML) tasks.

Syntactic TS with a focus on the task of sentence splitting has been attracting growing interest in the NLP community within the past few years. It has been shown that sentence splitting may benefit both NLP and societal applications. One line of work encompasses syntactic TS approaches that serve as \textbf{assistive technology for specific target reader populations} \cite{Ferres2016,Siddharthan2014,Saggion:2015:MSI:2775084.2738046}. The main goal of such approaches is to improve the readability of a text, i.e. to enhance the ease with which it can be understood.
Thus, it aims to make information easier to comprehend for people 
with reading difficulties that may arise from, for example, aphasia \cite{carroll1999simplifying}, autism \cite{evans-etal-2014-evaluation} or deafness \cite{Inui:2003:TSR:1118984.1118986}, as well as for those with reduced reading levels, including children \cite{debelder2010text} and non-native speakers of the respective language \cite{angrosh-etal-2014-lexico}. In that way, information is made available to a broader audience.
Apart from substituting a word or phrase that is hard to understand with a more comprehensible synonym (\textit{lexical simplification}), the most effective operations to improve the reading comprehension for language-impaired humans are splitting long sentences, making discourse relations explicit, avoiding pre-posed adverbial clauses and presenting information in cause-effect order \cite{siddharthan2014survey}. 

The second line of work aims at \textbf{generating an intermediate representation that is easier to analyze by downstream semantic tasks whose predictive quality deteriorates with sentence length and structural complexity}, posing problems due to their potential high level of ambiguity. Prior work has established that shortening sentences by dropping constituents or splitting components, 
and thus operating on smaller units of text, facilitates and improves the performance 
of a variety of applications, including Machine Translation  (MT) \cite{stajner2016can,stajner2018improvingMT}, Open Information Extraction (IE) \cite{cetto2018graphene,Swarnadeep2018}, Text Summarization \cite{siddharthan2004syntactic,bouayad2009improving}, Relation Extraction \cite{Miwa2010simplificationRE}, Semantic Role Labeling \cite{Vickrey2008}, Question Generation \cite{heilman2010extracting,bernhard2012question}, Sentence Fusion \cite{filippova-strube-2008-sentence} and Parsing \cite{Chandrasekar1996,jonnalagadda2009towards}.

Many different methods for addressing the task of TS have been proposed so far. As noted in \citeauthor{stajner2017leveraging} \citeyear{stajner2017leveraging}, data-driven approaches outperform rule-based systems in the area of lexical simplification, which aims at substituting a difficult word or phrase with a more comprehensible synonym \cite{glavas2015,Paetzold:2016:ULS:3016387.3016433,nisioi2017exploring,Zhang2017}. In contrast, the state-of-the-art structural simplification approaches are rule-based \cite{Siddharthan2014,Ferres2016,Saggion:2015:MSI:2775084.2738046}, providing more grammatical output and covering a wider range of syntactic transformation operations, however, at the cost of being very conservative, often to the extent of not making any changes at all.

\begin{figure}[!htb]
\centering
\tikzstyle{myblock} = [rectangle, draw, minimum height=2.5cm, rounded corners] 
\begin{tikzpicture}[scale=0.95, transform shape]
    \node (1)[myblock, text width=4cm,align=center, label=below:{\textbf{Proposition 1}}]{\small{A fluoroscopic study is typically the next step in management.}};
    \node (2)[myblock,right =of 1,xshift=0cm, text width=4cm,align=center, label=below:{\textbf{Proposition 2}}]{\small{This fluoroscopic study is known as an upper gastrointestinal series.}};
    \node (3)[myblock,right =of 2,xshift=0cm, text width=4cm,align=center, label=below:{\textbf{Proposition 3}}]{\small{Volvulus is suspected.}};
    \node (4)[myblock,below =of 1,xshift=0cm, text width=4cm, align=center, label=below:{\textbf{Proposition 4}}]{\small{Caution with non water soluble contrast is mandatory.}};
    \node (5)[myblock,below =of 2,xshift=0cm, text width=4cm, align=center, label=below:{\textbf{Proposition 5}}]{\small{The usage of barium can impede surgical revision.}};
    \node (6)[myblock,below =of 3,xshift=0cm, text width=4cm, align=center, label=below:{\textbf{Proposition 6}}]{\small{The usage of barium can lead to increased post operative complications.}};
    
\end{tikzpicture}
\caption{Minimal propositions resulting from the sentence splitting subtask on our running example from Figure \ref{intro_example1}. The input is decomposed into a loose sequence of minimal semantic units that lacks coherence.}
\label{fig:minimal_propositions}
\end{figure}
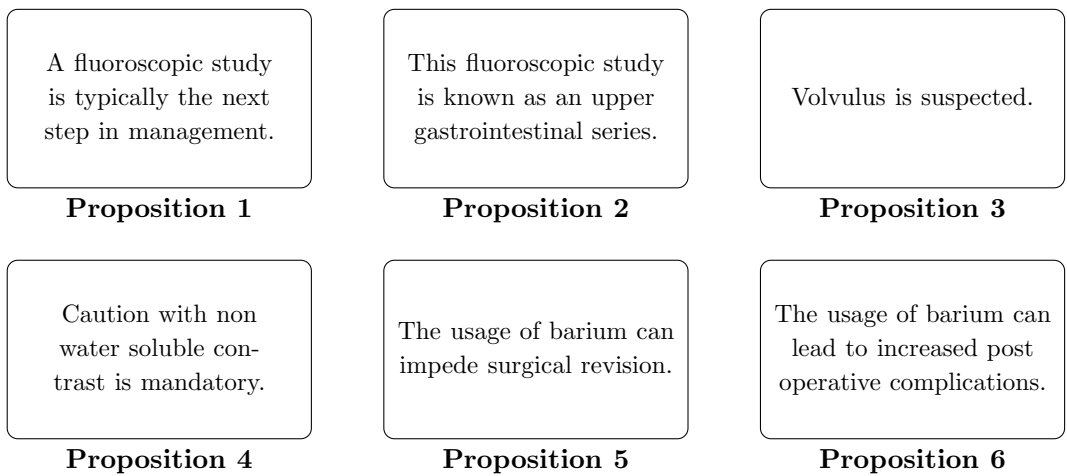

In order to overcome the conservatism exhibited by state-of-the-art syntactic TS approaches, i.e. their tendency to retain the input sentence rather than transforming it, we propose a novel sentence splitting approach that \textbf{breaks down a complex English sentence into a set of minimal propositions}, i.e. a sequence of sound, self-contained utterances, with each of them presenting a minimal semantic unit that cannot be further decomposed into meaningful propositions \cite{bast2013open} (for an example, see Figure \ref{fig:minimal_propositions}). Thus, we augment the Split-and-Rephrase task that was proposed in \citeauthor{Narayan2017} \citeyear{Narayan2017} by the notion of \textit{minimality}. 
In the development of our approach, we followed a principled and systematic procedure. First, we performed an in-depth study of the literature on syntactic sentence simplification, followed by a thorough linguistic analysis of the syntactic phenomena that need to be tackled in the sentence splitting task. Next, we materialized our findings into a small set of 35 hand-crafted transformation rules that decompose each source sentence into minimal semantic units and turn them into self-contained propositions that present a regular and grammatically sound structure. In that way, nested structures are removed and related pieces of information are brought closer together, resulting in a fine-grained representation of the input that is easier to process and thus leverages downstream Open IE applications, improving their performance in terms of accuracy and coverage.
Prior work has also shown that sentence splitting may be used as a preprocessing step to facilitate and boost the performance of further artificial intelligence (AI) tasks, including, for instance, Text Summarization \cite{bouayad2009improving} and MT \cite{stajner2018improvingMT}.

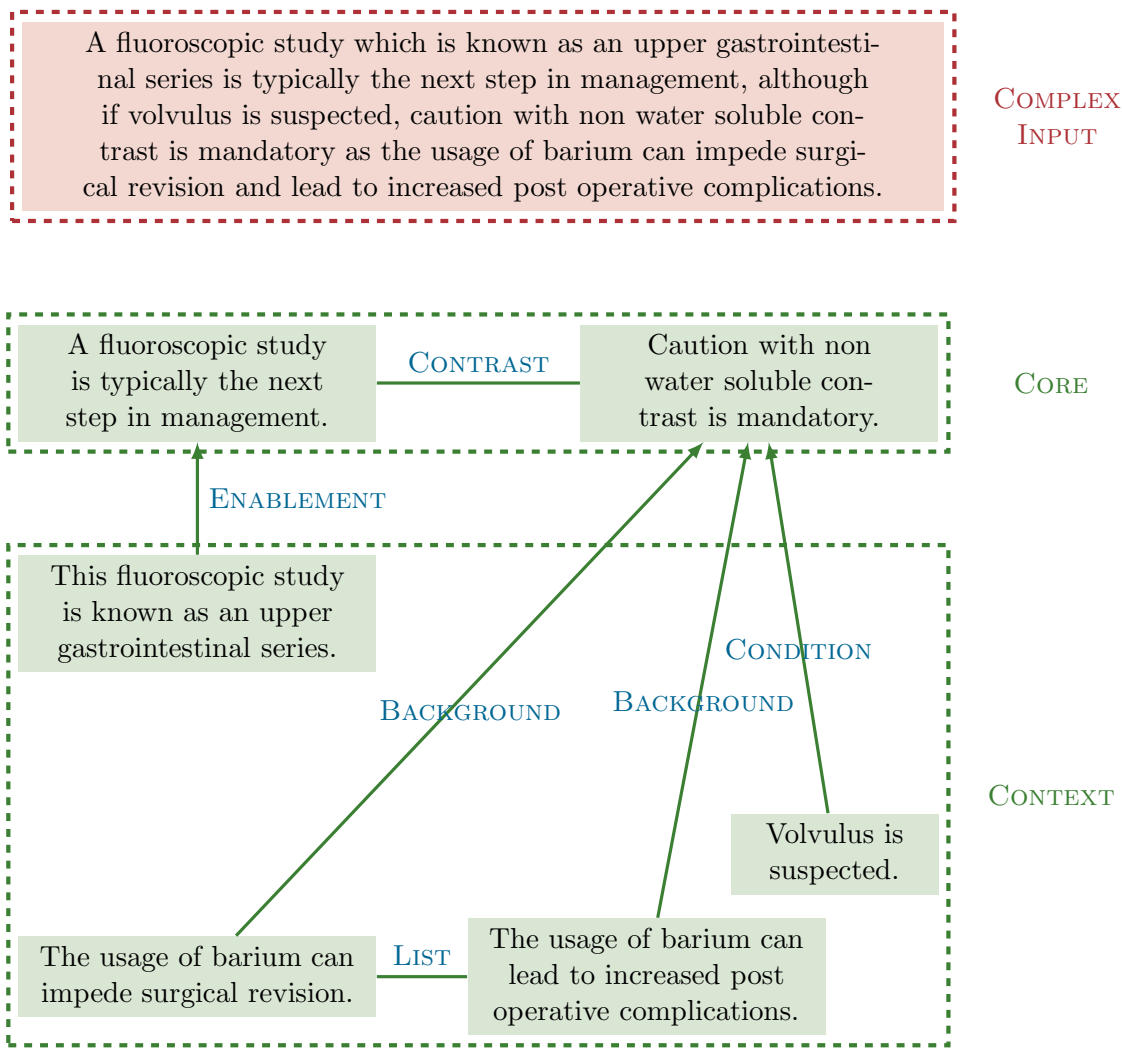
\begin{figure}[!htb]
\centering
\tikzset{node distance = 1.5cm and 0cm}
\tikzstyle{myblock} = [rectangle, minimum height=0.6cm] 

\begin{tikzpicture}
    \node (8)[myblock,fill=Maroon!15, text width=12cm,align=center]{A fluoroscopic study which is known as an upper gastrointestinal series is typically the next step in management, although if volvulus is suspected, caution with non water soluble contrast is mandatory as the usage of barium can impede surgical revision and lead to increased post operative complications.};
    \node (0)[myblock, fill=OliveGreen!15, xshift=-3.8cm, text width=4.5cm,align=center, below =of 8]{A fluoroscopic study is typically the next step in management.};
    \node (1)[myblock, fill=OliveGreen!15, xshift=2.7cm, text width=4.5cm,align=center, right =of 0]{Caution with non water soluble contrast is mandatory.};
    \node (2)[myblock, fill=OliveGreen!15, text width=4.5cm,align=center, below =of 0]{This fluoroscopic study is known as an upper gastrointestinal series.};
    \node (3)[myblock, fill=OliveGreen!15, text width=4.5cm,align=center, yshift=-2cm, below =of 2]{The usage of barium can impede surgical revision.};
    \node (4)[myblock, fill=OliveGreen!15, xshift=1.2cm, text width=4.5cm,align=center, right =of 3]{The usage of barium can lead to increased post operative complications.};
    \node (5)[myblock, fill=OliveGreen!15, text width=2.5cm,align=center, yshift=-1.2cm, xshift=2.5cm, above =of 4]{Volvulus is suspected.};

    \node (6)[draw, ultra thick, color=OliveGreen, dashed,fit=(0) (1)] {};
    \node (7)[draw, ultra thick, color=OliveGreen, dashed,fit=(2) (3) (4) (5)] {};
    \node (12)[draw, ultra thick, color=Maroon, dashed,fit=(8)] {};
    
    \node (9)[text width = 2cm, align=center, right =of 6, xshift=0.2cm]{\textcolor{OliveGreen}{\textsc{Core}}};
    \node (11)[text width = 2cm, align=center, right =of 7, xshift=0.2cm]{\textcolor{OliveGreen}{\textsc{Context}}};
    \node (13)[text width = 2cm, align=center, right =of 12, xshift=0.2cm]{\textcolor{Maroon}{\textsc{Complex\\Input}}};
    
    \draw[-,OliveGreen,very thick] (0) -- node[above, align=center] {\textcolor{MidnightBlue}{\textsc{Contrast}}} (1);
    \draw[->,OliveGreen,very thick] (2) -- node[right, align=center] {\textcolor{MidnightBlue}{\textsc{Enablement}}} (0);
    \draw[->,OliveGreen,very thick] (3) -- node[below, align=center] {\textcolor{MidnightBlue}{\textsc{Background}}} (1);
    \draw[->,OliveGreen,very thick] (4) -- node[below, align=center] {\textcolor{MidnightBlue}{\textsc{Background}}} (1);
    \draw[-,OliveGreen,very thick] (4) -- node[above, align=center] {\textcolor{MidnightBlue}{\textsc{List}}} (3);
    \draw[->,OliveGreen,very thick] (5) -- node[below, align=center] {\textcolor{MidnightBlue}{\textsc{Condition}}} (1);

\end{tikzpicture}
\caption{Example of the output that is generated by our proposed discourse-aware TS approach. A complex input sentence is transformed into a semantic hierarchy of simplified sentences in the form of minimal, self-contained propositions that are linked via rhetorical relations. The output presents a regular, fine-grained structure that is easy to process, while still preserving the coherence and, hence, the interpretability of the output.}
	\label{intro_example1}
\end{figure}



However, any sound and coherent text is not simply a loose arrangement of self-contained units, but rather a logical structure of utterances that are semantically connected \cite{siddharthan2014survey}. Consequently, when carrying out syntactic simplification operations without considering discourse implications, the rewriting may easily result in a disconnected sequence of simplified sentences, making the text harder to interpret (see the output of the example given in Figure \ref{fig:minimal_propositions}). 
Still, the vast majority of existing structural TS approaches do not take into account discourse level effects that arise from applying syntactic transformations. I.e., they split the input into simplified sentences without considering and preserving their semantic context. Accordingly, the resulting simplified text is prone to lack coherence, consisting of a set of semantically unrelated utterances that miss important contextual information. Therefore, the interpretability of the output for downstream Open IE tasks may be hampered. 
Thus, in order to \textbf{preserve the coherence structure} and, hence, the interpretability of the input, we propose a discourse-aware TS approach based on Rhetorical Structure Theory (RST) \cite{mann1988rhetorical} that establishes a semantic hierarchy on the simplified sentences. For this purpose, we first set up a contextual hierarchy between the split components, distinguishing \textit{core} sentences that contain the key information of the input from \textit{contextual} sentences that disclose less relevant, supplementary information. Then, the semantic relationship that holds between the decomposed spans is identified and classified. An example of the resulting context-preserving output is displayed in Figure \ref{intro_example1}.

The contributions of our work can be summarised as:

\begin{itemize}
\item[(i)] The proposal of a discourse-aware syntactic TS approach which transforms English sentences that present a complex linguistic structure into a semantic hierarchy of simplified sentences in the form of minimal, self-contained propositions.
\item[(ii)] The extension of the task of sentence splitting by the notion of minimality.
\item[(iii)] The introduction of a two-layered hierarchical representation of complex text data in the form of core sentences and accompanying contexts.
\item[(iv)] The use of rhetorical relations to preserve the coherence structure of the input, resulting in a novel contextual representation that puts a semantic layer on top of the simplified sentences. 
\item[(v)] A systematic analysis of the syntactic phenomena involved in complex sentence constructions whose findings were materialized in a small set of 35 hand-crafted simplification rules.
\item[(vi)] A comprehensive evaluation against state-of-the-art syntactic TS baselines.
\item[(vii)] An extrinsic evaluation demonstrating the merits of our approach for downstream Open IE applications. 
\item[(viii)] A supporting system (\textsc{DisSim}).
\item[(ix)] The publication of all the experimental code to support transparency and reproducibility.\footnote{\url{https://github.com/Lambda-3/DiscourseSimplification}}
\end{itemize}

This manuscript is organised as follows: in Section \ref{sec:related_work}, we provide an overview of the state of the art in structural TS. In addition, we discuss approaches addressing the problem of preserving coherence structures in texts.
In Section \ref{sec:approach}, we describe our discourse-aware TS approach that transforms complex input sentences into a semantic hierarchy of minimal propositions, resulting in a novel contextual representation of complex assertions in the form of hierarchically ordered and semantically interconnected sentences. 
In Section \ref{sec:evaluation}, we present the results of the experiments that we conducted to compare the performance of our proposed approach with state-of-the-art syntactic TS baselines. Moreover, we report the results of the evaluation of the semantic hierarchy, as well as the extrinsic evaluation assessing the effect on the performance of state-of-the-art Open IE systems when incorporating our approach as a preprocessing step.
Finally, in Section \ref{sec:conclusion}, we conclude our findings and present directions for future work.

This article substantially extends the work described in \citeauthor{niklaus-etal-2019-transforming} \citeyear{niklaus-etal-2019-transforming}. The \textbf{main novel contributions} of the work reported in this manuscript, as well as the \textbf{conclusions} that can be drawn from it are as follows:
\begin{enumerate}
    \item[(i)] \textbf{We emphasise and extend the work in the direction of the identification of the discourse structures} (see Section \ref{sec:subtask_2_description}). 
    While our previous paper is restricted to briefly outlining the core idea of this step, we here present the details that are necessary to fully capture and reproduce
    this part of our approach, describing for example the process of selecting the rhetorical relations included in our approach, the final set of rhetorical relations that we apply, and the logic behind differentiating core from context sentences, including important exceptions. 
    \item[(ii)] \textbf{We conduct a comprehensive evaluation from which we can conclude that our proposed discourse-aware TS approach not only succeeds in transforming syntactically complex sentences into a set of simplified propositions, but also in establishing a semantic hierarchy between them} (see Section \ref{sec:establishing_semantic_hierarchy}). Thus, it generates a fine-grained representation of complex assertions in the form of 
    \begin{itemize}
        \item[(a)] \textbf{hierarchically ordered} and
        \item[(b)] \textbf{semantically interconnected sentences}.
    \end{itemize}
    \item[(iii)] We present \textbf{additional detail about the sentence splitting subtask} by performing a systematic analysis of the syntactic phenomena involved in complex sentence constructions. The findings are materialized in a small set of transformation patterns whose details are presented in this article for the first time. 
    \textbf{It may both serve as a reference and support the re-implementation of the proposed approach.}
     \item[(iv)] \textbf{We extend the evaluation of the sentence splitting subtask by providing a detailed qualitative analysis of the transformation patterns}, including an in-depth analysis of the errors made by the proposed approach (see Section \ref{sec:eval_subtask_1}). These examinations provide further insights into the quality of the specified simplification rules and may serve as a starting point for future improvements.
    \item[(v)] While \citeauthor{niklaus-etal-2019-transforming} \citeyear{niklaus-etal-2019-transforming} is limited to a brief outline of the general approach, 
    thereby focusing on the sentence splitting subtask, this work provides a both \textbf{complete and detailed end-to-end narrative} of our proposed approach. 
\end{enumerate}

To summarise, this article articulates new empirical analyses and communicates the contribution with a level of detail and rigour which are not comparable to previous work. From our perspective this work has the chance of becoming a canonical reference in this specific subfield and in our minds, it constitutes a distinct contribution.

\section{Related Work}
\label{sec:related_work}

The following section summarizes previous work in the area of syntactic TS. We discuss the strengths and weaknesses of existing approaches, thereby demonstrating the effectiveness of our proposed context-preserving TS framework. In addition, we give a comprehensive overview of approaches that operate on the level of discourse by taking into account the coherence structure of texts.

\subsection{Sentence Splitting}

To date, three main classes of techniques for syntactic TS with a focus on the task of sentence splitting have been proposed. The first uses a set of syntax-based hand-crafted transformation rules to perform structural simplification operations, while the second exploits ML techniques where the model learns simplification rewrites automatically from examples of aligned complex source and simplified target sentences. In addition, approaches based on the idea of decomposing a sentence into its main semantic constituents using a semantic parser were described.

\subsubsection{Syntax-driven Rule-based Approaches}
The line of work on structural TS starts with \citeauthor{Chandrasekar1996} \citeyear{Chandrasekar1996}, who manually define a set of rules to detect points where sentences may be split, such as relative pronouns or conjunctions, based on chunking and dependency parse representations. \citeauthor{siddharthan2002architecture} \citeyear{siddharthan2002architecture} presents a pipelined architecture for a simplification framework that extracts a variety of clausal and phrasal components from a source sentence and transforms them into stand-alone sentences using a set of hand-written grammar rules based on shallow syntactic features.

More recently, \citeauthor{Siddharthan2014} \citeyear{Siddharthan2014} propose RegenT, a hybrid TS approach that combines an extensive set of 136 hand-written grammar rules defined over dependency tree structures for tackling 7 types of linguistic constructs with a much larger set of automatically acquired rules for lexical simplification. Taking a similar approach, \citeauthor{Ferres2016} \citeyear{Ferres2016} describe a linguistically-motivated rule-based TS approach called YATS, which relies on part-of-speech tags and syntactic dependency information to simplify a similar set of linguistic constructs, using a set of only 76 hand-crafted transformation patterns in total. A vector space model is applied to achieve lexical simplification. 
These two 
rule-based structural TS approaches primarily target reader populations with reading difficulties, such as people suffering from dyslexia, aphasia or deafness. According to \citeauthor{siddharthan2014survey} \citeyear{siddharthan2014survey}, those groups most notably benefit from substituting difficult words, making discourse relations explicit, reordering sentence components and splitting long sentences that contain clausal constructs. 
Therefore, TS approaches that are aimed at human readers commonly represent hybrid systems that operate both on the syntactic and lexical level, with the decomposition of clausal elements representing the main focus of the structural simplification component. 

\begin{table}[!htb]
\small
\centering
  \begin{tabular}{ | p{4.7cm} | c | c | c | }
    \hline
     \cellcolor{gray!25}{\textsc{Linguistic construct}} & \multicolumn{3}{|c|}{\cellcolor{gray!25}{\textsc{\#rules}}} \\ \hline 
    & \cellcolor{gray!25}{\textbf{YATS}}  & \cellcolor{gray!25}{\textbf{RegenT}} & \cellcolor{gray!25}{\textbf{\textsc{DisSim}}}  \\ \hline\hline
    Correlatives & 4 & 0 & 0 \\ \hline
    
    Coordinate clauses & \multirow{2}{*}{10} & \multirow{4}{*}{85 (lexicalised on conjunctions)} &  1 \\ \cline{1-1}\cline{4-4}
    
    Coordinate verb phrases &  & & 1 \\ \cline{1-2}\cline{4-4}
    
    Adverbial clauses & 12 &  & 6  \\ \hline
    
    Reported speech & 0 & 14 & 4  \\ \hline
    
    
    
    Relative clauses & 17 & \multirow{2}{*}{26} & 9  \\ \cline{1-2}\cline{4-4}
    
    Appositions & 1 &  & 2  \\ \hline
    Coordinate noun phrases & 0 & 0 & 2  \\ \hline
    Participial phrases & 0 & 0 & 4 \\ \hline
    Prepositional phrases & 0 & 0 & 3  \\ \hline
    Adjectival and adverbial phrases & 0 & 0 & 2 \\ \hline
    Lead noun phrases & 0 & 0 & 1  \\ \hline
    
  
    Passive constructions & 14 & 11 & 0  \\ \hline \hline
    
    \cellcolor{gray!25}{Total} & \cellcolor{gray!25}{76} & \cellcolor{gray!25}{136} & \cellcolor{gray!25}{35} \\ \hline

  \end{tabular} 
  
  \caption{Comparison of the linguistic constructs that are addressed by the syntactic TS approaches YATS, RegenT and our proposed approach \textsc{DisSim} based on the number of hand-crafted grammar rules for simplifying these structures.}
  \label{tab:overview_linguistic_constructs}
\end{table} 



In contrast to above-mentioned TS frameworks, our proposed TS approach (\textsc{DisSim}) does not address a human audience, but rather aims to \textbf{produce an intermediate representation that presents a simple and more regular structure that is easier to process for downstream Open IE applications}. 
For this purpose, we \textit{cover a wider range of syntactic constructs}. In particular, our approach is not limited to breaking up clausal components, but also splits and rephrases a variety of phrasal elements, resulting in a much more fine-grained output where each proposition represents a minimal semantic unit that is typically composed of a simple subject-predicate-object structure. Though tackling a larger set of linguistic constructs, our framework operates on a \textit{much smaller set of only 35 manually defined rules}.
A detailed comparison of the linguistic constructs that are handled by the two state-of-the-art rule-based syntactic TS systems YATS and RegenT and our proposed approach \textsc{DisSim} is shown in Table~\ref{tab:overview_linguistic_constructs}.

 \begin{table}[!htb]
 \footnotesize
 \centering
  \begin{tabular}{  l | p{11.5cm} }
    \toprule
    \textsc{System} & \textsc{Output} \\ \hline\hline
    \cellcolor{gray!25}{Input} & \cellcolor{gray!25}{\textit{The house was once part of a plantation and it was the home of Josiah Henson, a slave who escaped to Canada in 1830 and wrote the story of his life.}} \\ \hline
    RegenT & The house was once part of a plantation. And it was the home of Josiah Henson, a slave. This slave escaped to Canada in 1830 and wrote the story of his life. \\ \hline
   YATS & The house was once part of a plantation. And it was the home of Josiah Henson. Josiah Henson was a slave who escaped to Canada in 1830 and wrote the story of his life.\\ \hline
    \textsc{DisSim} & \begin{itemize}
    \item[\#1] 0 \textbf{The house was once part of a plantation.}
    \begin{itemize}
    \item[LIST] \#2
    \end{itemize}
    \item[\#2] 0 \textbf{It was the home of Josiah Henson.}
    \begin{itemize}
    \item[ELABORATION] \#3
    \item[LIST] \#1
    \end{itemize}
    \item[\#3] 1 \textbf{Josiah Henson was a slave.}
        \begin{itemize}
    \item[ELABORATION] \#4
    \item[ELABORATION] \#6
    \end{itemize}
    \item[\#4] 2 \textbf{This slave escaped to Canada.}
    \begin{itemize}
    \item[TEMPORAL] \#5
    \item[LIST] \#6
    \end{itemize}
    \item[\#5] 3 \textbf{This was in 1830.}
    \item[\#6] 2 \textbf{This slave wrote the story of his life.}
    \begin{itemize}
    \item[LIST] \#4
    \end{itemize}
    \end{itemize} \\ \bottomrule

\end{tabular} 
  
  \caption{Simplification example (from Newsela).}
  \label{outputexampleRuleBased}
\end{table}

Table~\ref{outputexampleRuleBased} contrasts the output generated by RegenT and YATS on a sample sentence. As can be seen, RegenT and YATS break down the input into a sequence of sentences that present its message in a way that is easy to digest for human readers. However, the sentences are still rather long and present an irregular structure that mixes multiple semantically unrelated propositions, potentially causing problems for downstream tasks, such as Open IE. On the contrary, our fairly aggressive simplification strategy that splits a source sentence into a large set of very short sentences is not well suited for a human audience and may in fact even hinder reading comprehension. However, we were able to demonstrate that the transformation process we propose improves the accuracy and coverage of the tuples extracted by state-of-the-art Open IE systems, serving as an example of a downstream NLP application that benefits from using our approach as a pre-processing step (see Section \ref{sec:extrinsic_eval}). 

In addition, \citeauthor{stajner2017leveraging} \citeyear{stajner2017leveraging} present \textsc{LexEv} and \textsc{EvLex}. They are semantically-motivated, aiming to reduce the amount of descriptions in a text and keeping only event-related information. For this purpose, they make use of a state-of-the-art event extraction system \cite{glavas2015} to first extract event mentions from a given text. Then, a set of 11 syntax-based extraction rules is applied to extract arguments of four types: \textit{agent, target, location} and \textit{time}, including rules for splitting noun compounds, extracting prepositional objects, nominal subjects and direct objects. For lexical simplification, an unsupervised approach based on word embeddings is used \cite{glavas2015}.
In comparison to \textsc{LexEv} and \textsc{EvLex}, where \textit{content reduction} plays a major role, we are interested in \textit{preserving the full informational content} of an input sentence. 

Lately, based on the work described in \citeauthor{evans2011comparing} \citeyear{evans2011comparing}, \citeauthor{evansorasan2019} \citeyear{evansorasan2019} propose a sentence splitting approach that combines data-driven and rule-based methods. In a first step, they use a CRF tagger \cite{Lafferty2001} to identify and classify signs of syntactic complexity, such as coordinators, \textit{wh}-words or punctuation marks. Then, a small set of 28 manually defined rules for splitting coordinate clauses and a larger set of 125 rules for decomposing nominally bound relative clauses is applied to rewrite long sentences into sequences of shorter single-clause sentences. 


\subsubsection{Approaches based on Semantic Parsing}
While the TS approaches described above are based on syntactic information, there are a variety of methods that use semantic structures for sentence splitting, including the work of \citeauthor{narayan2014hybrid} \citeyear{narayan2014hybrid} and \citeauthor{Narayan2016} \citeyear{Narayan2016}, who propose a framework that 
takes semantically-shared elements as the basis for splitting and rephrasing a sentence. It first generates a semantic representation of the input (in the form of Discourse Semantic Representations \cite{Kamp1981-KAMATO-2}) 
to identify splitting points in the sentence. In a second step, the split components are then rephrased by completing them with missing elements in order to reconstruct grammatically sound sentences. Lately, with DSS, \citeauthor{sulemSystem} \citeyear{sulemSystem} describe another semantic-based structural simplification framework that follows a similar approach, making use of the semantic parser presented in \citeauthor{hershcovich2017a} \citeyear{hershcovich2017a} which supports the direct decomposition of a sentence into individual semantic elements. Furthermore, they propose SENTS, a hybrid TS system that adds a lexical simplification component to DSS by running the split sentences through the neural MT-based system proposed in \citeauthor{nisioi2017exploring} \citeyear{nisioi2017exploring}.

Though these approaches pursue a very similar goal as compared to our framework, namely the decomposition of complex source sentences into minimal semantic units, they generally suffer from a \textit{poor grammaticality and fluency} of the simplified output. 

\subsubsection{Data-driven Approaches}
Recently, data-driven approaches for the task of sentence splitting emerged. 
\citeauthor{Narayan2017} \citeyear{Narayan2017} observed that existing TS corpora, such as PWKP \cite{zhu2010monolingual}, EW-SEW \cite{Coster:2011:SEW:2002736.2002865}, Newsela \cite{Xu2015newsela} or WikiLarge \cite{Zhang2017}, are ill suited for learning to decompose sentences into shorter, syntactically simplified components, as they contain only a small number of split examples.\footnote{For instance, according to \citeauthor{Narayan2017} \citeyear{Narayan2017}, the average number of simple sentences per complex source is 1.06 in PWPK. In fact, only 6.1\% of the complex input sentences are split into two or more simplified counterparts \cite{narayan2014hybrid}.} Thus, they compiled the WebSplit corpus,\footnote{\url{https://github.com/shashiongithub/Split-and-Rephrase}} the first TS dataset that explicitly addresses the task of sentence splitting, while abstracting away from lexical and deletion-based operations. It was derived in a semi-automatic way from the WebNLG corpus \cite{gardent-etal-2017-webnlg}, which contains 13K pairs of items, with each of them consisting of a set of RDF triples and one or more texts verbalising them. The resulting WebSplit dataset is composed of over one million tuples that map a single complex sentence to a sequence of structurally simplified sentences expressing the same meaning. \citeauthor{Narayan2017} \citeyear{Narayan2017} describe a set of sequence-to-sequence models \cite{bahdanau2014neural} trained on this corpus for breaking up complex input sentences into shorter components that present a simpler structure. 
\citeauthor{aharoni2018split} \citeyear{aharoni2018split} further explore this idea, augmenting the presented neural models with a copy mechanism \cite{Gu2016,See2017}. Though outperforming the models used in \citeauthor{Narayan2017} \citeyear{Narayan2017}, they still perform poorly compared to previous state-of-the-art rule-based syntactic simplification approaches. 

\citeauthor{Botha2018} \citeyear{Botha2018} noticed that the sentences from the WebSplit corpus contain fairly unnatural linguistic expressions using only a small vocabulary and a rather uniform sentence structure. 
To overcome this limitation, they present a scalable, language-agnostic method for mining large-scale training data from Wikipedia edit histories. Based on this approach, they compiled WikiSplit\footnote{\url{https://github.com/google-research-datasets/wiki-split}}, a new sentence splitting dataset that provides a rich and varied vocabulary over naturally expressed sentences and their extracted splits. 
When training the best-performing model of \citeauthor{aharoni2018split} \citeyear{aharoni2018split} on this corpus, they achieve a strong improvement over their prior best results. However, 
\citeauthor{Botha2018}s' \citeyear{Botha2018} approach exhibits a \textit{strong conservatism} since each input sentence is split in exactly two output sentences only, due to the uniform use of a single split per source sentence in the training set.
Consequently, the resulting simplified sentences are still comparatively long and complex, mixing multiple, potentially semantically unrelated propositions that are difficult to handle for downstream Open IE tasks. Though, as opposed to the semantic-based approaches presented above, this approach commonly succeeds in producing split sentences that are grammatically sound.

\subsection{Text Coherence}

When simplifying the structure of sentences without considering discourse implications, the rewriting may easily result in a loss of cohesion, hindering the process of making sense of the text. Thus, preserving the coherence of the input is crucial to maintain its interpretability in downstream applications.

\subsubsection{Discourse-level Syntactic Simplification}

The vast majority of existing structural TS approaches do not take into account discourse level effects that arise from splitting long and syntactically complex sentences into a sequence of shorter utterances. 
However, not considering the semantic context of the output propositions, the simplified text is prone to lack coherence, resulting in a set of incoherent propositions that miss important contextual information that is needed to infer the true meaning of the simplified output.
Though, two notable exceptions have to be mentioned.

\citeauthor{siddharthan2006syntactic} \citeyear{siddharthan2006syntactic} was the first to use discourse-aware cues in the simplification process, with the goal of generating a coherent output that is accessible to a wider audience. For this purpose, he adds a regeneration stage after the transformation process where, amongst others, simplified sentences are reordered, appropriate determiners are chosen 
and cue words in the form of rhetorical relations are introduced to connect them, thus making information easier to comprehend for people with reduced literacy. Though, as opposed to our approach, where a semantic relationship is established for each output sentence, only a comparatively low number of sentences is linked by such cue words in \citeauthor{siddharthan2006syntactic}'s \citeyear{siddharthan2006syntactic} framework.

Another approach that simplifies texts on the discourse level was proposed by \citeauthor{stajner2017leveraging} \citeyear{stajner2017leveraging}. It presents a semantically motivated method for eliminating irrelevant information from the input by maintaining only those parts that belong to factual event mentions. Our approach, on the contrary, aims to preserve all the information contained in the source. 
However, by distinguishing core from contextual information, we are still able to extract only the key information given in the input.

\subsubsection{Discourse Parsing}

The challenge of uncovering coherence structures in texts is pursued in the area of Discourse Parsing, which aims to identify discourse relations that hold between textual units in a document \cite{marcu1997rhetorical}. 
The annotated corpora of the Penn Discourse TreeBank (PDTB) \cite{miltsakaki2004penn} and the RST Discourse Treebank (RST-DT) \cite{carlson2002rst} have established two major directions in this area. They are grounded on different taxonomies: whereas PDTB provides a more shallow specification of discourse structures, RST-DT analyzes the hierarchical structure of discourse in texts based on Rhetorical Structure Theory (RST) \cite{mann1988rhetorical}.

In this work, we will concentrate on the latter taxonomy. 
In RST, coherence in texts is explained by the existence of rhetorical relations that hold between non-overlapping text spans in a hierarchical structure. Each rhetorical relation is defined on the basis of the function that the connected text spans take up in order to transport a certain message to the reader. These spans are specified as either a \textit{nucleus} (N) or a \textit{satellite} (S). The nucleus span embodies the central piece of information, whereas the role of the satellite is to further specify the nucleus. Rhetorical relations that connect a nucleus span and a satellite span are called \textit{mono-nuclear}, whereas rhetorical relations between two nucleus spans are denoted \textit{multi-nuclear}. For instance, the rhetorical relation \textit{Evidence} is defined as follows \cite{mann1988rhetorical}:

\begin{itemize}
    \item Constraints on N: The reader might not believe N to a degree satisfactory to the writer.
    \item Constraints on S: The reader believes S or will find it credible.
    \item Constraints on the N + S combination: The reader's comprehension of S increases the reader's belief of N.
    \item The effect: The reader's belief of N is increased.
\end{itemize}

An application of the \textit{Evidence} relation is illustrated in the following excerpt:
\begin{displayquote}
\textit{[The program as published for calendar year 1980 really works.]\textsubscript{N} [In only a few minutes, I entered all the figures from my 1980 tax return and got a result which agreed with my hand calculations to the penny.]\textsubscript{S}} \cite{mann1988rhetorical}
\end{displayquote}

Here, a positive practical experience of an income tax program is presented in the satellite span as an evidence to the reader in order to increase her belief that the program really works (nucleus span).
Besides the set and characteristics of rhetorical relations, \citeauthor{mann1988rhetorical} \citeyear{mann1988rhetorical} also introduced schema types that define the structural constituency arrangements of texts. In this way, RST is able to identify semantic relations both locally between neighbored elementary textual units and for the full document by using a hierarchical representation of nested schema type applications that cover long-range dependencies up to the whole document.

Over time, the original set of 25 relations presented in \citeauthor{mann1988rhetorical} \citeyear{mann1988rhetorical} has been extended. RST-DT currently contains 53 mono-nuclear and 25 multi-nuclear relations. 
Approaches to identify and detect rhetorical structure arrangements in texts range from early rule-based approaches \cite{marcu2000rhetorical} to supervised data-driven models that were trained on annotated corpora such as the RST-DT \cite{feng2014linear,li2014recursive}. Most commonly, these approaches separate the task of extracting discourse structures into two major subtasks: (1) elementary discourse unit (EDU) segmentation, where the input text is partitioned into basic textual units, and (2) a tree-building step, where the identified EDUs are interconnected by rhetorical relations to form a discourse tree. %
Due to the simplicity of the first task and the high accuracy achieved (90\% by \citeauthor{fisher2007utility} \citeyear{fisher2007utility}), recent research focuses on the more challenging task of the discourse tree construction, which includes the task of determining rhetorical relations between two consecutive text spans \cite{feng2014linear,joty2015codra}. 
Recent approaches often solve the tree-building task by training two individual classifiers: a binary structure classifier for finding pairs of consecutive text spans that are to be joined together, and a multi-class relation classifier for identifying the rhetorical relation that holds between them \cite{hernault2010hilda,li2014recursive,feng2014linear,joty2015codra}.
The most prominent features that are used by state-of-the-art systems for the rhetorical relation identification step are a mix of lexical and syntactic features, including cue phrases, part-of-speech tags, and information derived from syntactic parse trees \cite{feng2014linear,li2014recursive}. 

Although the segmentation step is related to the task of sentence splitting, it is not possible to simply use an RST parser for the sentence splitting subtask in our TS approach. 
Such a parser does not return grammatically sound sentences, but rather just copies the components that belong to the corresponding EDUs from the source. In order to reconstruct proper sentences, rephrasing is required. For this purpose, amongst others, referring expressions have to be identified, and phrases have to be rearranged and inflected. Moreover, the textual units resulting from the segmentation process are too coarse-grained for our purpose, since RST parsers mostly operate on clausal level. The goal of our approach, though, is to split the input into minimal semantic units, which requires to go down to the phrasal level in order to produce a much more fine-grained output in the form of minimal propositions.

\section{Discourse-Aware Sentence Splitting}
\label{sec:approach}
We present \textsc{DisSim}, a discourse-aware sentence splitting approach that creates a semantic hierarchy of simplified sentences.\footnote{To support transparency and reproducibility, the source code of our framework \cite{niklaus-etal-2019-dissim} is publicly available: \url{https://github.com/Lambda-3/DiscourseSimplification}.}
It takes a sentence as input and performs a recursive transformation stage that is based upon a small set of 35 hand-crafted grammar rules, encoding syntactic and lexical features that are derived from a sentence's phrase structure. Each rule specifies (1) how to \textit{split up and rephrase} the input into structurally simplified sentences  (see Section \ref{splittingsubtask}) and (2) how to \textit{set up a contextual hierarchy} between the split components (see Section \ref{constituencytypeclassification}) and \textit{identify the semantic relationship} that holds between those elements (see Section \ref{rhetoricalrelationidentification}). In that way, a semantic hierarchy of simplified sentences is established, providing an intermediate representation of the input that can be used to facilitate and improve the performance of downstream Open IE tasks. 


The transformation patterns were heuristically determined in a rule engineering process that was carried out on the basis of an in-depth study of the literature on syntactic sentence simplification \cite{siddharthan2006syntactic,siddharthan2014survey,siddharthan2002architecture,Siddharthan2014,evans2011comparing,evansorasan2019,heilman2010extracting,Vickrey2008,shardlow2014survey,SaggionsurveyTextSimplification,mallinson2019controllable,suter2016rule,Ferres2016,Chandrasekar1996}, followed by a thorough linguistic analysis of the syntactic phenomena that need to be tackled in the sentence splitting task. Our main goal was to provide a best-effort set of rules, targeting the challenge of being applied in a recursive fashion and to overcome biased or incorrectly structured parse trees. We empirically determined a fixed execution order of the specified rules by examining which sequence achieved the best simplification results in a qualitative manual analysis conducted on a development test set of 100 randomly sampled Wikipedia sentences. The grammar rules are applied recursively in a top-down fashion on the source sentence, until no more simplification pattern matches (see Algorithm \ref{alg:transformation}). For reproducibility purposes, the complete set of transformation patterns is detailed in the appendix. 

\begin{table}[!htb]
\centering
  \begin{tabular}{ l }
    \toprule
   {ROOT $<<:$ (S $<$ (NP $\$..$ (VP $<+$(VP) \underline{(\textbf{SBAR} $<,$ (\textit{IN} $\$+$ (\fbox{S $<$ (NP $\$..$ VP}))))})))} \\ \bottomrule

    \end{tabular}
  
  \caption{Example of a transformation pattern (for decomposing adverbial clauses with post-posed subordinative clauses).
  They are specified in terms of Tregex patterns \cite{Levy2006}. 
  A boxed pattern represents the part of a sentence that is extracted from the input and transformed into a new stand-alone sentence. A pattern in bold is deleted from the source. The underlined part is labelled as a context sentence, while the remaining part represents core information. The italic pattern is used as a cue phrase for identifying the rhetorical relation that holds between the decomposed spans.}
  \label{examplePatterns}

\end{table}

An example of a transformation rule is provided in Table \ref{examplePatterns}. In total, we manually defined a set of 35 patterns that are applied in a pre-determined order on a given input sentence. By disembedding clausal and phrasal constituents that contain only supplementary information, source sentences that present a complex linguistic structure are recursively split into shorter, syntactically simplified components. These elements are then converted into self-contained sentences. Hereby, the input is reduced to its key information (\textit{``core sentence''}) and augmented with a number of associated \textit{contextual sentences} that disclose additional information about it, resulting in a novel hierarchical representation in the form of core sentences and accompanying contexts. Moreover, we determine the rhetorical relations by which the split sentences are linked in order to preserve their semantic relationship. The resulting representation consisting of a set of hierarchically ordered and semantically interconnected sentences (see Figure \ref{example_full_final_running_example}) that present a simplified syntax will be denoted a \textit{linked proposition tree} in the following.

\begin{algorithm}[!htb]
\scriptsize
\caption{Transform into Semantic Hierarchy}
\label{alg:transformation}
\begin{algorithmic}[1]
    \Require{complex source sentence $str$}
    \Ensure{linked proposition tree $tree$}
    \Statex
    \Function{Initialize}{$str$}
        \State $new\_leaves$ $\gets$ source sentence $str$
        \State $new\_node \gets$ create a new parent node for $new\_leaves$
        \State $new\_node.labels$ $\gets$ None
        \State $new\_node.rel$ $\gets$ ROOT
        \State linked proposition tree $tree$ $\gets$ initialize with $new\_node$
        \State \Return $tree$
    \EndFunction
    \Statex
    \Procedure{TraverseTree}{$tree$}
      \LineComment{Process leaves (i.e. propositions) from left to right}
      \For {$leaf$ in $tree.leaves$}
        \LineComment{Check transformation rules in fixed order}
        \For {$rule$ in $TRANSFORM\_RULES$}
       \If{$match$} 
          \LineComment{{}{\textbf{(a) Sentence splitting}}} 
          \State 
          {}{$simplified\_propositions \gets$ decompose $leaf$ into a }
          \State {}{set of simplified propositions}
          \State {}{$new\_leaves \gets$ convert $simplified\_propositions$}
          \State {}{into leaf nodes}
          \LineComment{{}{\textbf{(b) Constituency Type Classification}}} 
          \State 
          {}{$new\_node \gets$ create a new parent node for $new\_leaves$}
          \State {}{$new\_node.labels \gets$ link each leaf in $new\_leaves$ to}
          \State {{}}{ $new\_node$ and label each edge with the leaf's constituency} 
          \State {}{ type $c \in CL$}
          
          
          \LineComment{{}{\textbf{(c) Rhetorical Relation Identification}}}
          \State {}{$cue\_phrase \gets$ extract cue phrase from $leaf.parse\_tree$}
          \State {}{$new\_node.rel \in REL \gets$ match $cue\_phrase$ against a}
          \State {}{ predefined set of rhetorical cue words}
          
          \LineComment{Update Tree}
          \State $tree.replace(leaf, new\_node)$
          \LineComment{Recursion}
          \State \Call{TraverseTree}{$tree$}
         
        \EndIf
   \EndFor
   \EndFor
   \State \Return $tree$
    \EndProcedure
\end{algorithmic}
\end{algorithm}

\subsection{Subtask 1: Splitting into Minimal Propositions}
\label{splittingsubtask}
In a first step, source sentences that present a complex linguistic form are transformed into regular, compact structures by disembedding clausal and phrasal components that contain only supplementary information. For this purpose, the transformation rules encode both the \textit{splitting points} and \textit{rephrasing procedure} for reconstructing proper sentences. 
Each rule takes a sentence's phrasal parse tree\footnote{generated by Stanford's pre-trained lexicalized parser \cite{Socher2013}} as input and encodes a pattern that, in case of a match, will extract textual parts from the tree. The decomposed text spans, as well as the remaining constituents are then transformed into proper stand-alone sentences. In order to ensure that the resulting simplified output is grammatically sound, the extracted spans are combined with their corresponding referents from the main sentence or appended to a simple canonical phrase (e.g. \textit{``This is''}), if necessary.
Table \ref{rulesAndFrequency} provides an overview of the linguistic constructs that are tackled by our approach, including the number of patterns that were specified for the respective syntactic phenomenon.

\begin{table}[!htb]
\small
\centering
  \begin{tabular}{ c  l  c   c  }
    \toprule
    & \textsc{Clausal/Phrasal type} & \textsc{Hierarchy} & \textsc{\# rules}  \\ \hline 
    \multicolumn{4}{c} {\textbf{Clausal disembedding}} \\ 
    1 & Coordinate clauses & coordinate & 1 \\ 

    2 & Adverbial clauses & subordinate & 6  \\ 
    
    
    
    3a & Relative clauses (non-restrictive) & subordinate & 5  \\ 
    
    3b & Relative clauses (restrictive) & subordinate & 4  \\ 
    
    4 & Reported speech & subordinate & 4  \\ \hline
    
    \multicolumn{4}{c} {\textbf{Phrasal disembedding}} \\ 
    5 & Coordinate verb phrases & coordinate & 1 \\ 
    6 &  Coordinate noun phrases & coordinate & 2  \\ 
     6 &  Participial phrases & subordinate & 4  \\ 
    8a &  Appositions (non-restrictive) & subordinate & 1  \\ 
    8b & Appositions (restrictive) & subordinate & 1  \\ 
    
    9 & Prepositional phrases & subordinate & 3  \\ 
    10 & Adjectival and adverbial phrases & subordinate & 2 \\ 
    11 & Lead NPs & subordinate & 1  \\ \hline 
    
    
    & Total & & 35 \\ \bottomrule

  \end{tabular} 
  
  \caption{Linguistic constructs addressed by \textsc{DisSim}, including both the hierarchical level and the number of simplification rules that were specified for each syntactic construct.}
  \label{rulesAndFrequency}
\end{table}

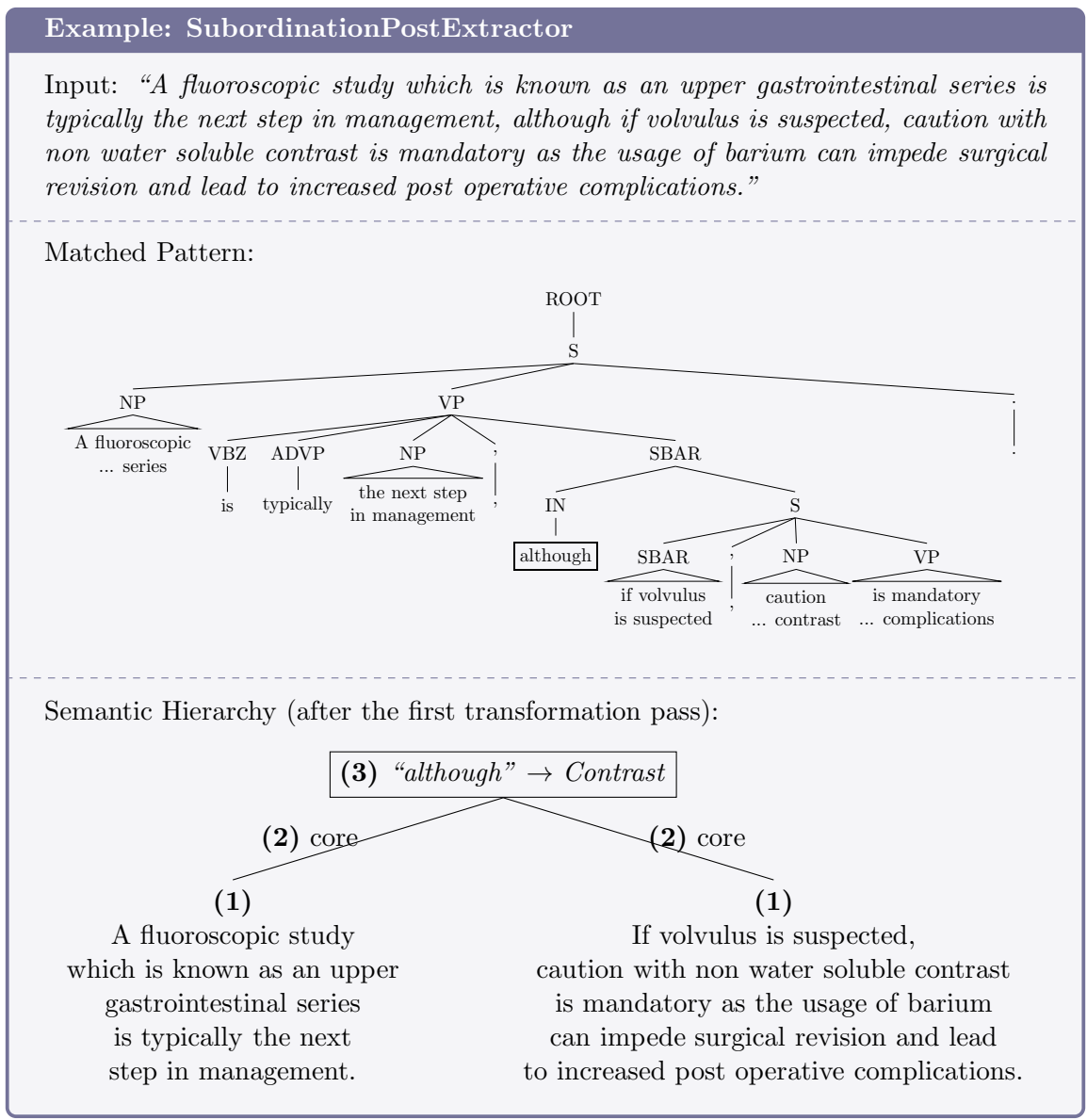
\begin{figure}[!ht]
\centering
\begin{examplebox}{Example: \textsc{SubordinationPostExtractor}}
Input: \textit{``A fluoroscopic study which is known as an upper gastrointestinal series is typically the next step in management, although if volvulus is suspected, caution with non water soluble contrast is mandatory as the usage of barium can impede surgical revision and lead to increased post operative complications.''}
\tcbline
Matched Pattern:
\begin{center}
\begin{tikzpicture}[scale=0.69, every tree node/.style={align=center}]
\Tree [.ROOT
        [.S
            [.NP \edge[roof]; {A fluoroscopic\\ ... series} ]
            [.VP
                [.VBZ is ]
                [.ADVP typically ]
                [.NP \edge[roof]; {the next step\\ in management} ]
                [., , ]
                [.SBAR
                    [.IN \framebox{although} ]
                    [.S 
                        [.SBAR \edge[roof]; {if volvulus\\ is suspected} ]
                        [., , ]
                        [.NP \edge[roof]; {caution\\ ... contrast} ]
                        [.VP \edge[roof]; {is mandatory\\ ... complications} ]
                    ]
                ] 
            ]
            [.. . ]
        ]
      ]
\end{tikzpicture}
\end{center}
\tcbline
Semantic Hierarchy (after the first transformation pass):
\begin{center}
\begin{tikzpicture}[scale=1, level distance=3cm, sibling distance=1.5cm, every tree node/.style={align=center}]
\Tree [.\node[style={draw,rectangle}] {\textbf{(3)} \textit{``although''} $\rightarrow$ \textit{Contrast}}; 
  \edge node[midway, left] {\textbf{(2)} core}; {\textbf{(1)}\\ A fluoroscopic study \\ which is known as an upper \\ gastrointestinal series \\is typically the next \\step in management.}
  \edge node[midway, right] {\textbf{(2)} core}; {\textbf{(1)}\\If volvulus is suspected, \\ caution with non water soluble contrast\\ is mandatory as the  usage of barium\\ can impede surgical revision  and lead\\ to increased post operative complications.}
]
\end{tikzpicture}
\end{center}
\end{examplebox}

\caption{\textbf{(Subtask 1)} The source sentence is split up and rephrased into a set of syntactically simplified sentences. \textbf{(Subtask 2)} Then, the split sentences are connected with information about their constituency type to establish a contextual hierarchy between them. \textbf{(Subtask 3)} Finally, by identifying and classifying the rhetorical relation that holds between the simplified sentences, their semantic relationship is restored which can be used to inform downstream Open IE applications.}
\label{fig:subordination_post_example_approach}
\end{figure}

For a better understanding of the splitting procedure, Figure \ref{fig:subordination_post_example_approach} visualizes the application of the first grammar rule that matches the input sentence from our running example. 
The upper part of the box represents the complex source sentence, which is matched against the specified simplification pattern (see Table \ref{examplePatterns}). The lower part then depicts the result of the corresponding transformation operation.

\subsection{Subtask 2: Establishing a Semantic Hierarchy}
\label{sec:subtask_2_description}
Each split will create two or more sentences with a simplified syntax. In order to establish a semantic hierarchy between them, two subtasks are carried out: constituency type classification and rhetorical relation identification.

\subsubsection{Constituency Type Classification}
\label{constituencytypeclassification}
First, we set up a contextual hierarchy between the split sentences. For this purpose, we connect them with information about their constituency type. According to \citeauthor{collinsgrammar} \citeyear{collinsgrammar}, clauses can be related to one another in two ways: first, there are parallel clauses that are linked by coordinating conjunctions, and second, clauses may be embedded inside another, introduced by subordinating conjunctions. The same applies to phrasal elements. Since subordinations commonly express less relevant information, we denote them ``context sentences''. In contrast, coordinations are of equal status and typically depict the key information contained in the input. Therefore, they are called ``core sentences'' in our approach. To differentiate between those two types of constituents, the transformation patterns encode a simple syntax-based approach where subordinate clauses and subordinate phrasal elements are classified as context sentences, while their superordinate counterparts as well as coordinate clauses and coordinate phrases are labelled as core (see Table \ref{rulesAndFrequency}).

However, two exceptions have to be noted. In case of an attribution, the subordinate clause expressing the actual statement (e.g. \textit{``Economic activity continued to increase.''}) is assigned a core tag, while the superordinate clause containing the fact that it was uttered by some entity (e.g. \textit{``This was what the Federal Reserve noted.''}) is labelled as contextual information. The reason for this is that the latter is considered less relevant as compared to the former, which holds the key information of the source sentence. Moreover, when disembedding adverbial clauses of contrast (as determined in the rhetorical relation identification step, see Section \ref{rhetoricalrelationidentification}), both the superordinate and the subordinate clause are labelled as a core sentence (see the example in Figure \ref{fig:subordination_post_example_approach}), in accordance with the theory of RST, where text spans that present a ``Contrast'' relationship are considered both as nuclei.

This approach allows for the distinction of core information capturing the key message of the input from contextual information that provides only supplementary material, resulting in a two-layered hierarchical representation in the form of core facts and accompanying contextual information. 
This directly relates to the concept of nuclearity in RST, which specifies each text span as either a nucleus or a satellite. The nucleus span embodies the central piece of information and is comparable to what we denote a core sentence, whereas the role of the satellite is to further specify the nucleus, corresponding to a context sentence in our case.

\subsubsection{Rhetorical Relation Identification}
\label{rhetoricalrelationidentification}
Second, we aim to preserve the semantic relationship between the decomposed spans. For this purpose, we identify and classify the rhetorical relations that hold between the simplified sentences, making use of both syntactic and lexical features that are encoded in the transformation patterns. While syntactic features are manifested in the phrasal composition of a sentence's parse tree, lexical features are extracted from the parse tree in the form of cue phrases. The determination of potential cue words and their positions in specific syntactic environments is based on the work of \citeauthor{knott1994using} \citeyear{knott1994using}. The extracted cue phrases are then used to infer the type of rhetorical relation. For this task, we utilize a predefined list of rhetorical cue words adapted from the work of \citeauthor{Taboada13} \citeyear{Taboada13}, which assigns them to the relation that they most likely trigger.\footnote{The full list of cue phrases that serve as lexical features for the identification of rhetorical relations in our approach, as well as the corresponding relations to which they are mapped,
is provided in the appendix.} For example, the transformation rule in Table \ref{examplePatterns}, which is the first pattern that matches our running example, specifies that the phrase \textit{``although''} is the cue word here, which is mapped to a ``Contrast'' relationship. The semantic hierarchy that results from the first transformation pass is depicted in the lower part of Figure \ref{fig:subordination_post_example_approach}.

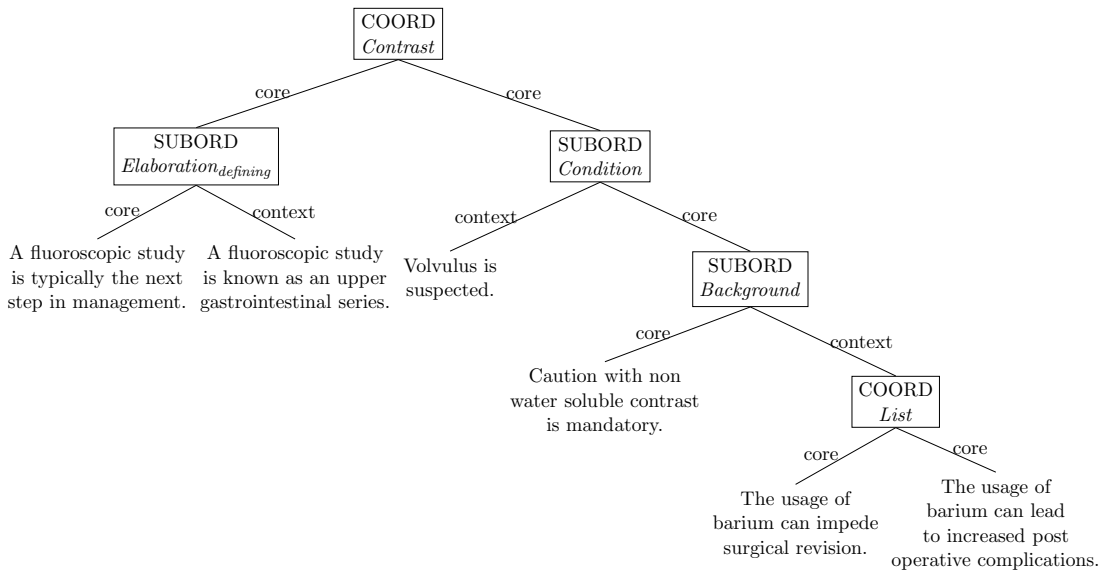
\begin{figure}[!ht]
\centering
\begin{tikzpicture}[scale=0.68, level distance=2.4cm, sibling distance=0cm, every tree node/.style={align=center, transform shape}]
\Tree [
      .\node [style={draw,rectangle}] {COORD\\\textit{Contrast}};
            \edge node[midway, left] {core}; [
                .\node [style={draw,rectangle}] {SUBORD\\\textit{Elaboration\textsubscript{defining}}};
                    \edge node[midway, left] {core}; [.\node(a){A fluoroscopic study\\ is typically the next \\step in management.};]
                    \edge node[midway, right] {context}; [.\node(b){A fluoroscopic study\\ is known as an upper\\ gastrointestinal series.};]
            ]
            \edge node[midway, right] {core}; [
                .\node [style={draw,rectangle}] {SUBORD\\\textit{Condition}};
                     \edge node[midway, left] {context}; [.\node(c){Volvulus is\\ suspected.};]
                    \edge node[midway, right] {core};[.\node [style={draw,rectangle}] {SUBORD\\\textit{Background}};
                        \edge node[midway, left] {core}; [.\node(d){Caution with non\\ water soluble contrast\\ is mandatory.};]
                        \edge node[midway, right] {context}; [.\node [style={draw,rectangle}] {COORD\\\textit{List}};
                            \edge node[midway, left] {core}; [.\node(e){The usage of\\ barium can impede\\ surgical revision.};]
                            \edge node[midway, right] {core}; [.\node(f){The usage of\\ barium can lead\\ to increased post\\ operative complications.};]
                        ]
                    ]
            ]
        ]
]
\end{tikzpicture}
\caption{Final linked proposition tree of the example sentence.}
\label{example_full_final_running_example}
\end{figure}

The process of selecting the rhetorical relations to include in our approach was guided by the following two questions:
\begin{itemize}
    \item[(i)] Which rhetorical relations are the most relevant for downstream Open IE applications?
    \item[(ii)] Which of them are likely to be recognized by syntactic and lexical features according to the work of \citeauthor{Taboada13} \citeyear{Taboada13}?
\end{itemize}

Based on the above considerations, we employ a subset of the classical set of RST relations defined in \citeauthor{mann1988rhetorical} \citeyear{mann1988rhetorical}. This comprises the \textit{Contrast, List, Disjunction, Cause, Result, Background, Condition, Elaboration} and \textit{Purpose} relationships. Additionally, we adopt a number of relations from the extended set of rhetorical relations defined in the RST-DT, including \textit{Temporal-After, Temporal-Before} and \textit{Attribution}. In order to deal with the problem of linking split sentences whose connecting rhetorical relation could not be identified, the custom relations \textit{Unknown-Coordination} and \textit{Unknown-Subordination} are introduced. Moreover, representing classes of context that were frequently encountered in our example sentences, we add the relations \textit{Spatial} and \textit{Temporal} for the identification of semantic relationships that provide local or temporal information about a presented situation.

Accordingly, we end up with the following set of rhetorical relations:
\begin{quote}
    \textit{\{Contrast, List, Disjunction, Cause, Result,
Background, Condition, Elaboration, Attribution, Purpose, Temporal-Before, Temporal-After, Temporal, Spatial, Unknown-Coordination, Unknown-Subordination\}}\footnote{Following the approach described in \citeauthor{Taboada13} \citeyear{Taboada13}, we later map the \textit{List} and \textit{Disjunction} relationships to a common \textit{Joint} class and integrate
the \textit{Result} relation into the semantic category of \textit{Cause} relationships. Moreover, both \textit{Temporal-After} and \textit{Temporal-Before} become part of the \textit{Temporal} class of rhetorical relations. Finally, \textit{Purpose} is embedded in the class of \textit{Enablement} relationships.}
\end{quote}

The lower part of Figure \ref{fig:subordination_post_example_approach} depicts the result of the first transformation pass on the example sentence. 
Afterwards, the resulting leaf nodes are recursively simplified in a top-down fashion. When no more rule matches, the algorithm stops and returns the generated linked proposition tree,\footnote{A linked proposition tree is a labeled binary tree whose leaf nodes represent the minimal propositions, whereas its inner nodes represent rhetorical relations that hold between a pair of associated split sentences. Every node of the tree is connected to its parent by an edge that presents either a core or a context label. For more details on the data model, see \citeauthor{niklaus-etal-2022-shallow} \citeyear{niklaus-etal-2022-shallow}.} representing the semantic hierarchy of simplified sentences for the given input sentence (see Figure \ref{example_full_final_running_example}).\footnote{For a detailed step-by-step example demonstrating the complete transformation process, the interested reader may refer to the appendix.} 

\section{Evaluation}
\label{sec:evaluation}
To evaluate our proposed discourse-aware TS approach, we first compare its performance with state-of-the-art syntactic simplification systems with respect to the sentence splitting task (subtask 1). Next, we assess the accuracy of the generated semantic hierarchy (subtask 2) on the basis of the RST-DT corpus. In addition, we extrinsically evaluate the usefulness of our approach as a preprocessing step for downstream semantic applications in the context of the task of Open IE. 

\subsection{Experimental Setup}

In the following, we present the experimental setup for evaluating the performance of our reference implementation \textsc{DisSim} with regard to the two subtasks of (1) splitting and rephrasing syntactically complex input sentences into a set of \textit{minimal propositions}, and (2) setting up a \textit{semantic hierarchy} between the split components, based on constituency type classification and rhetorical relation identification. In addition, we describe the setting for assessing the effectiveness of our approach for downstream Open IE tasks.


\subsubsection{Subtask 1: Splitting into Minimal Propositions}

\paragraph{Datasets}
To evaluate the performance of our discourse-aware TS framework with regard to the sentence splitting subtask, we conducted experiments on three commonly applied TS corpora from two different domains. The first dataset we used was \textit{WikiLarge} \cite{xu2016optimizing}. It consists of 359 sentences from the PWKP corpus \cite{zhu2010monolingual}, which is made up of aligned complex English Wikipedia sentences and Simple English Wikipedia counterparts. In WikiLarge, each complex sentence from PWKP is paired with eight simplification variants that were collected by crowdsourcing. Moreover, to demonstrate domain independence, we compared the output generated by our TS approach with that of the various baseline systems on the \textit{Newsela} corpus \cite{Xu2015newsela}, which is composed of 1077 sentences from newswire articles. In addition, we assessed the performance of our simplification system using the 5000 test sentences from the \textit{WikiSplit} benchmark \cite{Botha2018}, which was mined from Wikipedia edit histories. 


\paragraph{Baselines}
We compared our reference implementation \textsc{DisSim} against several state-of-the-art TS baseline systems that have a strong focus on syntactic transformations through explicitly modeling splitting operations. For WikiLarge, these include (i) the recent semantic-based DSS model that decomposes a sentence into its main semantic constituents using a semantic parser; (ii) SENTS, which is an extension of DSS that runs the split sentences through the NTS system proposed in \citeauthor{nisioi2017exploring} \citeyear{nisioi2017exploring}; (iii) \textsc{Hybrid} \cite{narayan2014hybrid}, a further TS approach that takes semantically-shared elements as the basis for splitting and rephrasing a sentence; (iv) the hybrid TS system YATS; and (v) RegenT, another hybrid state-of-the-art simplification approach for combined lexical and syntactic simplification. In addition, we report evaluation scores for both the complex input sentences, which allows for a better judgment of system conservatism, and the corresponding simple reference sentences. With respect to the Newsela dataset, we considered the same baseline systems, with the exceptions of DSS and SENTS, whose outputs were not available. Finally, regarding the WikiSplit corpus, we restricted the comparison of our TS approach to the best-performing system in \citeauthor{Botha2018} \citeyear{Botha2018}, Copy512, which is a sequence-to-sequence neural model augmented with a copy mechanism and trained over the WikiSplit dataset.

\paragraph{Automatic metrics} The automatic metrics that were calculated in the evaluation procedure comprise a number of basic statistics, including (i) the average sentence length of the simplified sentences in terms of the average number of tokens per output sentence (\#T/S); (ii) the average number of simplified output sentences per complex input (\#S/C); (iii) the percentage of sentences that are copied from the source without performing any simplification operation (\%SAME), serving as an indicator for system conservatism, i.e. the tendency of a system to retain the input sentence rather than transforming it; and (iv) the averaged Levenshtein distance from the input (LD\textsubscript{SC}), which provides further evidence for a system's conservatism. Furthermore, in accordance with prior work on TS, we report average BLEU \cite{papineni2002bleu} and SARI \cite{xu2016optimizing} scores for the rephrasings of each system. Finally, we computed the SAMSA and SAMSA\textsubscript{abl} scores of each system, which are the first metrics that explicitly target syntactic aspects of TS \cite{sulemsemantic}. The SAMSA metric is based on the idea that an optimal split of the input is one where each predicate-argument structure is assigned its own sentence in the simplified output and measures to what extent this assertion holds for the input-output pair under consideration. Accordingly, the SAMSA score is maximized when each split sentence represents exactly one semantic unit in the
input. SAMSA\textsubscript{abl}, in contrast, does not penalize cases where the number of sentences in the simplified output is lower than the number of events contained in the input, indicating separate semantic units that should be split into individual target sentences for obtaining minimal propositions.

\paragraph{Human evaluation} Human evaluation was carried out on a subset of 50 randomly sampled sentences per corpus by 2 in-house non-native, but fluent English speakers who rated each input-output pair for the different systems according to three parameters: grammaticality, meaning preservation and structural simplicity. The detailed guidelines presented to the annotators are given in 
the supplementary material. Regarding the grammaticality and meaning preservation dimensions, we adopted the guidelines from \citeauthor{stajner2017leveraging} \citeyear{stajner2017leveraging}, with some minor deviations to better reflect our goal of simplifying the \textit{structure} of the input sentences, while \textit{retaining their full informational content}. Besides, since the focus of our work is on structural rather than lexical simplification, we followed the approach taken in \citeauthor{sulemSystem} \citeyear{sulemSystem} in terms of the third parameter, simplicity, and neglected the lexical simplicity of the output sentences. Instead, we restricted our analysis to the syntactic complexity of the resulting sentences, which was measured on a scale that ranges from -2 to 2 in accordance with \citeauthor{nisioi2017exploring} \citeyear{nisioi2017exploring}.

\paragraph{Manual analyses} In order to get further insights into the quality of our implemented simplification patterns, we performed an extensive qualitative analysis of the 35 hand-crafted transformation rules, including an examination of the frequency distribution of the syntactic constructs that are addressed by our TS approach, a manual recall-based analysis of the simplification patterns that we defined, and a detailed error analysis. For this purpose, we compiled a dataset consisting of 100 Wikipedia sentences per syntactic phenomenon tackled by our TS approach.\footnote{available under \url{https://github.com/Lambda-3/DiscourseSimplification/blob/master/supplemental_material/dataset_pattern_analysis.zip}} In the construction of this corpus we ensured that the collected sentences exhibit a great syntactic variability to allow for a reliable predication about the coverage and accuracy of the specified transformation rules.

\subsubsection{Subtask 2: Establishing a Semantic Hierarchy}

We evaluated the constituency type classification and rhetorical relation identification steps by mapping the minimal propositions that were generated in the previous subtask to the EDUs of the RST-DT corpus, a collection of 385 Wall Street Journal articles annotated for rhetorical relations.

\paragraph{Constituency type classification} To determine whether the hierarchical relationship that was assigned by our framework between a pair of simplified sentences is correct, we check if the hierarchy of the contextual layers corresponds to the nuclearity of the aligned text fragments from RST-DT. 

\paragraph{Rhetorical relation identification} For each matching sentence pair, we examine whether the rhetorical relation assigned by our approach equates the relation that connects the corresponding sentences in the RST-DT dataset. For this purpose, we use the more coarse-grained classification scheme from \citeauthor{Taboada13} \citeyear{Taboada13}, who group the full set of 78 rhetorical relations that are contained in the RST-DT corpus into 19 classes of relations that share rhetorical meaning.

\subsubsection{Extrinsic Evaluation} 

The fine-grained representation of complex sentences in the form of hierarchically ordered and semantically interconnected propositions may serve as an intermediate representation for downstream tasks.
An application area that may benefit greatly from our approach 
as a preprocessing step is the task of Open IE \cite{Banko07}.
The goal of this task is
to turn unstructured information expressed in natural language text into a structured representation in the form of relational tuples that consist of a set of arguments and a phrase denoting a relation between them, e.g. $\langle$\textit{arg\textsubscript{1}:} Barack Obama; \textit{rel:} served as; \textit{arg\textsubscript{2}:} the 44th President of the US$\rangle$. In contrast to traditional IE systems, Open IE approaches allow for a domain-independent discovery of relations in large amounts of text, as they do not depend on any relation-specific human input \cite{niklaus-etal-2018-survey}.

In order to demonstrate the merits of our approach for the task of Open IE, we integrated it as a preprocessing step into a variety of state-of-the-art Open IE systems, including OpenIE-4 \cite{Mausam16}, ClausIE \cite{DelCorro13}, Stanford Open IE \cite{Angeli15}, \textsc{ReVerb} \cite{Fader11} and \textsc{Ollie} \cite{Mausam12}. We then investigated whether these systems profit from our proposed discourse-aware TS approach along two different dimensions.
In a first step, we measured if their performance was improved in terms of \textit{recall} and \textit{precision} of the generated relational tuples when splitting complex source sentences into a set of syntactically simplified minimal propositions with the help of our TS approach before performing the extraction step. In a second step, we examined whether the semantic hierarchy generated by our approach was beneficial in \textit{preserving the interpretability} of the output.
\subsection{Results and Discussion}

Below, we present and discuss the results of our evaluation procedure.

\subsubsection{Subtask 1: Splitting into Minimal Propositions}
\label{sec:eval_subtask_1}
In the following, we describe the evaluation results with respect to the first subtask of our approach, whose goal is to decompose sentences that present a complex linguistic structure into a set of syntactically simplified propositions, with each of them representing a minimal semantic unit.

\paragraph{Automatic metrics}

The upper part of Table \ref{resultsAutomaticEval} reports the results that were achieved on the 359 sentences from the WikiLarge corpus, using a set of automatic metrics. Transforming each sentence of the dataset, our reference implementation \textsc{DisSim} reaches the highest splitting rate among the TS systems under consideration, together with \textsc{Hybrid}, DSS and SENTS. With 2.82 split sentences per input on average, our framework outputs by a large margin the highest number of structurally simplified sentences per source. Moreover, consisting of 11.01 tokens on average, the \textsc{DisSim} approach returns the shortest sentences of all systems, on an average halving the length of the input. 
The relatively high word-based Levenshtein distance of 11.90 confirms previous findings, suggesting that our structural TS approach tends to extensively split the source sentences instead of retaining the input by simply copying the input without performing any simplification operations. SENTS (13.79) is the only system that shows an even higher Levenshtein distance to the source.

\begin{table}[!htb]
\centering
\small
  \begin{tabular}{  p{2.4cm} | c  c  c  c | c  c | c  c  }
    \toprule
    & \#T/S & \#S/C & \%\newline SAME & LD\textsubscript{SC} & BLEU & SARI & SAM\-SA & SAM\-SA\textsubscript{abl} \\ \hline \hline
    \multicolumn{9}{c}{\textbf{359 test sentences from the Wikilarge corpus}} \\\hline
    \cellcolor{gray!25}{Complex source} & \cellcolor{gray!25}{22.06} & \cellcolor{gray!25}{1.03} & \cellcolor{gray!25}{100} & \cellcolor{gray!25}{0.00}  & \cellcolor{gray!25}{94.25} & \cellcolor{gray!25}{32.53} & \cellcolor{gray!25}{0.59} & \cellcolor{gray!25}{0.96}\\ 
    \cellcolor{gray!25}{Simple reference} & \cellcolor{gray!25}{20.19} & \cellcolor{gray!25}{1.14} & \cellcolor{gray!25}{0.00} & \cellcolor{gray!25}{7.14}  & \cellcolor{gray!25}{99.48} & \cellcolor{gray!25}{43.09} & \cellcolor{gray!25}{0.48} & \cellcolor{gray!25}{0.78}  \\ 
    \textsc{DisSim} & \textbf{11.01} & \textbf{2.82} & \textbf{0.00} & 11.90           & 63.03          & \textbf{35.05} & \textbf{0.67} & 0.84 \\ 
    
    DSS                      & 12.91          & 1.87          & \textbf{0.00} & 8.14             & 74.42          & 34.32 & 0.64          & 0.75\\ 
    
    SENTS               & 14.17          & 1.09          & \textbf{0.00} & \textbf{13.79}       & 54.37          & 29.76 & 0.40          & 0.58   \\ 
    
    \textsc{Hybrid}               &    13.44       &    1.03      &  \textbf{0.00} &  13.04      &    48.97      & 26.19  &    0.47     & 0.76   \\ 
    
    YATS                     & 18.83          & 1.40          & 18.66        & 4.44          & 73.07          & 33.03 & 0.56          & 0.80  \\ 
    
    RegenT                   & 18.20          & 1.45          & 41.50        & 3.77     & \textbf{82.49}          & 32.41 & 0.61          & \textbf{0.85}\\ \hline\hline

    \multicolumn{9}{c}{\textbf{1077 test sentences from the Newsela corpus}} \\\hline
    \cellcolor{gray!25}{Complex source} & \cellcolor{gray!25}{23.34} & \cellcolor{gray!25}{1.01} & \cellcolor{gray!25}{100} & \cellcolor{gray!25}{0.00}  & \cellcolor{gray!25}{20.91} & \cellcolor{gray!25}{9.84} & \cellcolor{gray!25}{0.49} & \cellcolor{gray!25}{0.96} \\ 
    \cellcolor{gray!25}{Simple reference} & \cellcolor{gray!25}{12.81} & \cellcolor{gray!25}{1.01} & \cellcolor{gray!25}{0.00} & \cellcolor{gray!25}{16.25}  & \cellcolor{gray!25}{100} & \cellcolor{gray!25}{91.13} & \cellcolor{gray!25}{0.25} & \cellcolor{gray!25}{0.46}  \\ 
    
    \textsc{DisSim} & \textbf{11.20} & \textbf{2.96} & \textbf{0.00} & 13.00           & 14.54          & \textbf{49.00} & \textbf{0.57} & 0.84 \\ 
    
     \textsc{Hybrid}   &   12.49         &     1.02      &     \textbf{0.00}     & \textbf{13.46} & 14.42       &   40.34       & 0.38 &       0.74        \\ 
    
    YATS                     & 18.71          & 1.42         & 16.16       & 5.03      & 17.51          & 36.88 & 0.50          & 0.83  \\ 
    
    RegenT                   & 16.74          & 1.61          & 33.33       & 5.03   & \textbf{18.96}         & 32.83 & 0.55         & \textbf{0.85} \\ \hline\hline

    \multicolumn{9}{c}{\textbf{5000 test sentences from the WikiSplit corpus}} \\\hline
    \cellcolor{gray!25}{Complex source} & \cellcolor{gray!25}{32.01} & \cellcolor{gray!25}{1.10} & \cellcolor{gray!25}{100} & \cellcolor{gray!25}{0.00}  & \cellcolor{gray!25}{74.28} & \cellcolor{gray!25}{29.91} & \cellcolor{gray!25}{0.37} & \cellcolor{gray!25}{0.95} \\ 
    \cellcolor{gray!25}{Simple reference} & \cellcolor{gray!25}{18.14} & \cellcolor{gray!25}{2.08} & \cellcolor{gray!25}{0.00} & \cellcolor{gray!25}{7.48}  & \cellcolor{gray!25}{100} & \cellcolor{gray!25}{94.71} & \cellcolor{gray!25}{0.49} & \cellcolor{gray!25}{0.75}  \\ 
    
    \textsc{DisSim} & \textbf{11.91} & \textbf{4.09} & \textbf{0.76} & \textbf{19.10} &  51.96          & 39.33 & \textbf{0.54} & \textbf{0.84} \\ 
    
    Copy512                     & 16.55          & 2.08         & 13.30        & 2.39     & \textbf{76.42}          & \textbf{61.51} & 0.51         & 0.78  \\ \bottomrule
   
  \end{tabular} 
  
  \caption{Automatic evaluation results. We report evaluation scores for a number of basic statistics. These include the average sentence length of the simplified sentences in terms of the average number of tokens per output sentence (\#T/S) and the average number of simplified output sentences per complex input (\#S/C). Furthermore, in order to assess system conservatism, we measure the percentage of sentences that are copied from the source without performing any simplification operation (\%SAME) and the averaged Levenshtein distance from the input (LD\textsubscript{SC}). Moreover, we calculate the average BLEU and SARI scores for the rephrasings of each system, as well as their SAMSA and SAMSA\textsubscript{abl} scores, which are the first metrics targeting syntactic aspects of TS. The highest score by each evaluation criterion is shown in bold.}
  \label{resultsAutomaticEval}
\end{table}


With regard to SARI, our \textsc{DisSim} framework (35.05) again outperforms the baseline systems.\footnote{Note that SARI is a lexical simplicity metric that measures the goodness of words that are added, deleted and kept in the simplified output generated by the TS systems \cite{xu2016optimizing}. As detailed in the appendix, regarding SARI, splitting-only approaches have a natural advantage over systems that perform lexical simplification operations, as well as simple syntactic transformations in the form of deletions. This explains \textsc{DisSim}'s comparatively high SARI score.}
However, it is among the systems with the lowest BLEU score (63.03).
Though, \citeauthor{sulemBLEU2018} \citeyear{sulemBLEU2018} recently demonstrated that BLEU is ill suited for the evaluation of TS approaches when sentence splitting is involved, since in this case
it negatively correlates with structural simplicity, thus penalizing sentences that present a simplified syntax. Moreover, they showed that the BLEU score presents no correlation with the grammaticality and meaning preservation dimensions. For this reason, we only report these scores for the sake of completeness and to match past work.

According to \citeauthor{sulemsemantic} \citeyear{sulemsemantic}, the recently proposed SAMSA and SAMSA\textsubscript{abl} metrics are better suited for the evaluation of the sentence splitting task. With a score of 0.67, the \textsc{DisSim} framework shows the best performance for SAMSA, while its score of 0.84 for SAMSA\textsubscript{abl} is just below the one obtained by the RegenT system (0.85). As reported in \citeauthor{sulemsemantic} \citeyear{sulemsemantic}, SAMSA highly correlates with human judgments for structural simplicity and grammaticality, while SAMSA\textsubscript{abl} achieves the highest correlation for meaning preservation.

The results on the Newsela dataset, depicted in the middle part of Table \ref{resultsAutomaticEval}, support our findings on the WikiLarge corpus, indicating that our TS approach can be applied in a domain independent manner.
The lower part of Table \ref{resultsAutomaticEval} illustrates the numbers achieved on the WikiSplit dataset. Though the Copy512 system beats our approach in terms of BLEU and SARI, the remaining scores are clearly in favour of the \textsc{DisSim} system. With an average length of 11.91 tokens per simplified sentence, it reduces the source sentences to about one third of its original length. The tendency to perform a large amount of splitting operations is underlined by both the high number of simple sentences per complex input (4.09) and the large Levenshtein distance from the source (19.10), as well as the low percentage of unmodified input sentences (0.76\%). The SAMSA (0.54) and SAMSA\textsubscript{abl} scores (0.84) convey further evidence that our approach succeeds in producing structurally simplified sentences that are grammatically sound and preserve the original meaning of the input.

\paragraph{Manual analyses}
In the following, we report the results of the human judgments and the manual qualitative analysis of the transformation patterns.

\begin{table}[!htb]
\centering
\small
\begin{tabular}{  l | c  c  c | c }
    \toprule
    & G & M & S & average \\ \bottomrule 
    \multicolumn{5}{c}{\textbf{WikiLarge test set}} \\ \hline
    
    \cellcolor{gray!25}{Simple reference} & \cellcolor{gray!25}{4.70} & \cellcolor{gray!25}{4.56} & \cellcolor{gray!25}{-0.2} & \cellcolor{gray!25}{3.02} \\ 
    
    \textsc{DisSim} & 4.36 & 4.50 & \textbf{1.30} & \textbf{3.39} \\ 
    
    DSS & 3.44 & 3.68 & 0.06 & 2.39 \\ 
    
    SENTS & 3.48 & 2.70 & -0.18 & 2.00 \\ 
    
    \textsc{Hybrid} & 3.16 & 2.60 & 0.86 & 2.21 \\ 
    
    YATS & 4.40 & \textbf{4.60} & 0.22 & 3.07 \\ 
    
    RegenT & \textbf{4.64} & 4.56 & 0.28 & 3.16 \\ \bottomrule

    \multicolumn{5}{c}{\textbf{Newsela test set}} \\ \hline
    
    \cellcolor{gray!25}{Simple reference} & \cellcolor{gray!25}{4.92} & \cellcolor{gray!25}{2.94} & \cellcolor{gray!25}{0.46} & \cellcolor{gray!25}{2.77} \\ 
    
    \textsc{DisSim} & 4.44 & 4.60 & \textbf{1.38} & \textbf{3.47} \\ 
    
    \textsc{Hybrid} & 2.97 & 2.35 & 0.93 & 2.08 \\ 
    
    YATS & 4.26 & 4.42 & 0.32 & 3.00 \\ 
    
    RegenT & \textbf{4.54} & \textbf{4.70} & 0.62 & 3.29 \\ \bottomrule

    \multicolumn{5}{c}{\textbf{WikiSplit test set}} \\ \hline
    
    \cellcolor{gray!25}{Simple reference} & \cellcolor{gray!25}{4.72} & \cellcolor{gray!25}{4.32} & \cellcolor{gray!25}{0.44} & \cellcolor{gray!25}{3.16} \\ 
    
    \textsc{DisSim} & 4.36 & 4.36 & \textbf{1.66} & \textbf{3.46} \\ 
    
    Copy512 & \textbf{4.72} & \textbf{4.72} & 0.92 & 3.45 \\ \bottomrule

  \end{tabular} 
  
  \caption{Human evaluation ratings on a random sample of 50 sentences from each dataset. Grammaticality (G) and meaning preservation (M) are measured using a 1 to 5 scale. A -2 to 2 scale is used for scoring the \textit{structural} simplicity (S) of the output relative to the input sentence. The last column \textit{(average)} presents the average score obtained by each system with regard to all three dimensions. The highest score by each evaluation criterion is shown in bold.} 
  \label{resultsHumanEval}
    
\end{table}

\subparagraph{Human judgments}
The results of the human evaluation are displayed in Table~\ref{resultsHumanEval}. 
The inter-annotator agreement was calculated using Cohen's $\kappa$ \cite{cohen1968weighted}, resulting in rates of 0.72 (G), 0.74 (M) and 0.60 (S).
The assigned scores demonstrate that our \textsc{DisSim} approach outperforms all other TS systems in the simplicity dimension (S). With a score of 1.30 on the WikiLarge sample sentences, it is far ahead of the baseline approaches, with \textsc{Hybrid} (0.86) coming closest. However, this system receives the lowest scores for grammaticality (G) and meaning preservation (M). RegenT obtains the highest score for G (4.64), while YATS is the best-performing approach in terms of M (4.60). With a rate of only 0.22, though, it achieves a low score for S, indicating that the high score in the M dimension is due to the conservative approach taken by YATS, resulting in only a small number of simplification operations. This explanation also holds true for RegenT's high mark for G. Still, our \textsc{DisSim} approach follows closely, with a score of 4.50 for M and 4.36 for G, suggesting that it obtains its goal of returning fine-grained simplified sentences that achieve a high level of grammaticality and preserve the meaning of the input. Considering the average scores of all systems under consideration, our approach is the best-performing system (3.39), followed by RegenT (3.16). The human evaluation ratings on the Newsela and WikiSplit sentences show similar results, again supporting the domain independence of our proposed approach.

\subparagraph{Qualitative analysis of the transformation rule patterns}
Table \ref{tab:recall} shows the results of the recall-based qualitative analysis of the transformation patterns. With an average rate of more than 80\% for the syntactic phenomena under consideration, the overall ratio of the simplification rules being fired whenever the corresponding clausal or phrasal component is present in the given input sentence is very high. However, the patterns for the extraction of coordinate noun phrases, as well as prepositional phrases exhibit a much lower recall ratio. Regarding the former, the main reason is that we follow a very conservative approach when splitting lists of noun phrases, since this type of syntactic construct is prone to parsing errors and hard to distinguish from appositions. When decomposing prepositional phrases, we are confronted with the complex problem of resolving attachment ambiguities that is pervasive when dealing with this kind of phrasal element \cite{Gelbukh2018,PP2015}. 

\begin{table}[!htb]
\centering
\small
  \begin{tabular}{  l | c  c  c }
    \toprule
    & frequency & \%fired & \%correct transformations  \\ \bottomrule 
    \multicolumn{4}{c} {\textbf{Clausal disembedding}} \\ \hline
    Coordinate clauses & 113 & 93.8\% & 99.1\% \\ 
    
    Adverbial clauses & 113 & 84.1\% & 96.8\% \\ 
    
    Relative clauses (non-defining) & 108 & 88.9\% & 70.8\% \\ 
    
    Relative clauses (defining) & 103 & 86.4\% & 75.3\% \\ 
    
    Reported speech & 112 & 82.1\% & 75.0\% \\ \bottomrule
    
    \multicolumn{4}{c} {\textbf{Phrasal disembedding}} \\ \hline 
    Coordinate verb phrases & 109 & 85.3\% & 89.2\% \\ 
    Coordinate noun phrases & 115 & 48.7\% & 82.1\% \\ 
    Participial phrases & 111 & 76.6\% & 72.9\% \\ 
    Appositions (non-restrictive) & 107 & 86.0\% & 83.7\%\\ 
    Appositions (restrictive) & 122 & 87.7\% & 72.0\% \\ 
    
    Prepositional phrases & 163 & 68.1\% & 75.7\%  \\ \bottomrule
    
    Total & 1276 & 80.7\% & 81.1\% \\
     \bottomrule
    
  \end{tabular} 
  
  \caption{Recall-based qualitative analysis of the transformation rule patterns. This table presents the results of a manual analysis of the performance of the hand-crafted simplification patterns. The first column lists the syntactic phenomena under consideration, the second column indicates its frequency in the dataset, the third column displays the percentage of the grammar fired, and the fourth column reveals the percentage of sentences where the transformation operation results in a correct split.}
  \label{tab:recall}
\end{table}

With respect to the accuracy of the specified transformation patterns in terms of the percentage of input sentences that were correctly split into syntactically simplified output sentences, the rules for disembedding coordinate and adverbial clauses perform remarkably well, approaching an accuracy rate of almost 100\%. On average, correct transformations are carried out in over 80\% of the cases. The syntactic construct that provides the lowest accuracy are non-restrictive relative clauses, which are prone to missing some essential part from the source sentence or assigning the wrong attachment phrase to the extracted clausal component, as revealed by the error analysis. 

\begin{table}[!htb]
\centering
\small
  \begin{tabular}{  l | c  c  c  c  c  c }
    \toprule
    & Error 1 & Error 2 & Error 3 & Error 4 & Error 5 & Error 6 \\ \bottomrule 
    \multicolumn{7}{c} {\textbf{Clausal disembedding}} \\ \hline
    Coordinate clauses & 1 & 0 & 0 & 0 & 0 & 0\\ 
    
    Adverbial clauses & 1 & 1 & 0 & 1 & 0 & 0 \\ 
    
    Relative clauses (non-defining)  & 5 & 8 & 0 & 0 & 14 & 1 \\ 
    
    Relative clauses (defining)  & 8 & 8 & 2 & 0 & 5 & 1 \\ 
    
    Reported speech & 5 & 1 & 13 & 1 & 2 & 1 \\ \bottomrule
    
    \multicolumn{7}{c} {\textbf{Phrasal disembedding}} \\ \hline Coordinate verb phrases & 4 & 3 & 2 & 1 & 0 & 0 \\ 
    Coordinate noun phrases & 3 & 3 & 0 & 3 & 1 & 0\\ 
    Participial phrases & 2 & 2 & 4 & 2 & 13 & 0\\ 
    Appositions (non-restrictive) & 0 & 5 & 3 & 0 & 7 & 0 \\ 
    Appositions (restrictive)  & 1 & 21 & 3 & 0 & 0 & 0 \\ 
    
    Prepositional phrases & 3 & 11 & 4 & 6 & 4 & 0 \\ \bottomrule
    
    Total & 33  & 63  & 31  & 14 & 46 & 3 \\
    & (17\%) & (33\%) & (16\%) & (7\%) & (24\%) & (2\%) \\ \bottomrule
    
  \end{tabular} 
  
  \caption{Results of the error analysis. Six types of errors were identified (Error 1: additional parts; Error 2: missing parts; Error 3: morphological errors; Error 4: wrong split point; Error 5: wrong referent; Error 6: wrong order of the syntactic elements).}
  \label{tab:error_analysis}
\end{table}

\subparagraph{Error analysis}
Representing about 60\% of the erroneous simplifications, missing elements of the input when constructing the simplified sentences and allocating the wrong referent are the most frequently experienced problems that were identified in the qualitative analysis of the performance of the hand-crafted simplification rules. In total, six types of errors were identified, as detailed in Table \ref{tab:error_analysis}. With the help of an example sentence, each error class is illustrated in the appendix. 

To give a clearer picture of the source of the errors, we complemented the analysis described above with a tracing where the errors start. We found that they can be attributed to two causes: either the parse tree constructed by the constituency parser is erroneous or the simplification rules we specified fail to capture slight nuances or variations of the input presenting a rarely occurring sentence structure, as illustrated in the following examples:
\begin{itemize}
    \item \textit{``Chun was later pardoned by President Kim Young-sam on the advice of then President-elect Kim Dae-jung.''}: Here, the adverb modifier \textit{``then''} is missed when decomposing the appositive phrase \textit{``then President-elect''}.
    \item \textit{``Mars has two Moons, Phobos and Deimos.''}: Here, the appositive phrase \textit{``Phobos and Deimos''} is extracted and transformed into a stand-alone sentence. However, the plural form is not recognized by our approach, resulting in a simplified sentence with an incorrectly inflected verb (\textit{``is''} instead of \textit{``are''}).
    \item \textit{``She is the adoptive mother of actor Dylan McDermott, whom she adopted when he was 18.''}: The simplified sentence resulting from the given relative clause presents an unusual arrangement of its syntactic elements. This is due to the fact that the specified transformation rules capture only simple pronoun-subject-verb constructs in the case of relative clauses that are introduced by the pronoun \textit{``whom''}.
\end{itemize}

As Table \ref{tab:error_analysis2} shows, morphological errors and wrong orderings of the syntactic elements in the simplified output mostly appear due to particular structures that are not covered by the specified transformation patterns. In contrast, additional parts in the simplified sentences, wrong split points and wrong referents can typically be traced back to erroneous parses:
\begin{itemize}
    \item \textit{``The largest populations of Mennonites are in the Democratic Republic of Congo and the United States.''}: In this example, the noun phrase \textit{``the United States''} is mistakenly attributed as a prepositional phrase referring to \textit{``the Democratic Republic''} due to an erroneous parse tree, leading to a simplified sentence that incorrectly includes an additional noun phrase.
    \item \textit{``Peter Ermakov killed her with a gun shot to the left side of her head.''}: This input is split at the wrong point due to an erroneous parse tree that mistakenly attributes the prepositional phrase \textit{``to the left side of her head''} to the pronoun \textit{``her''} instead of the noun phrase \textit{``a gun shot''} to which it actually refers. Therefore, it is transformed into a separate simplified sentence that lacks coherence with the surrounding simplified sentences.
    \item \textit{``Pushkin and his wife Natalya Goncharova, whom he married in 1831, later became regulars of the court society.''}: Here, the parser fails to identify the correct referent of the relative clause (\textit{``Natalya Goncharova''}). Instead, it mistakenly recognizes the full coordinate noun phrase of the superordinate clause as its antecedent.
\end{itemize}

Errors that can be attributed to erroneous parse trees will diminish as the performance of the underlying parser increases. Thus, more attention should be paid to errors caused by underspecified simplification rules. According to Table \ref{tab:error_analysis2}, there is the most potential for improvement with respect to reported speech and restrictive appositive phrases.

\begin{table}[!htb]
\centering
\small
  \begin{tabular}{  l | c  c  c  c  c  c | c}
    \toprule
    & Error 1 & Error 2 & Error 3 & Error 4 & Error 5 & Error 6 & Total \\ \bottomrule 
    \multicolumn{8}{c} {\textbf{Clausal disembedding}} \\ \hline
    Coordinate clauses & 0 - 1 & --- & --- & --- & --- & --- & 0.00 \\ 
    Adverbial clauses & 0 - 1 & 0 - 1 & --- & 1 - 0 & --- & --- & 0.33 \\ 
    Relative clauses (non-def.)  & 4 - 1 & 1 - 7 & --- & --- & 8 - 6 & 0 - 1 & 0.46 \\ 
    Relative clauses (def.)  & 8 - 0 & 8 - 0 & 1 - 1 & --- & 5 - 0 & 0 - 1 & 0.92 \\ 
    Reported speech & 2 - 3 & 0 - 1 & 0 - 13 & 1 - 0 & 2 - 0 & 1 - 0 & 0.26 \\ \bottomrule
    \multicolumn{8}{c} {\textbf{Phrasal disembedding}} \\ \hline 
    Coordinate verb phrases & 3 - 1 & 0 - 3 & 0 - 2 & 1 - 0 & --- & --- & 0.40 \\ 
    Coordinate noun phrases & 3 - 0 & 0 - 3 & --- & 2 - 1 & 0 - 1 & --- & 0.50 \\ 
    Participial phrases & 0 - 2 & 0 - 2 & 4 - 0 & 0 - 2 & 9 - 4 & --- & 0.57 \\ 
    Appositions (non-restr.) & --- & 5 - 0 & 1 - 2 & --- & 5 - 2 & --- & 0.73 \\ 
    Appositions (restr.)  & 0 - 1 & 7 - 14 & 0 - 3 & --- & --- & --- & 0.28 \\ 
    Prepositional phrases & 0 - 3 & 7 - 4 & 0 - 4 & 4 - 2 & 2 - 2 & --- & 0.46   \\ \bottomrule
    Total & 0.61 & 0.44 & 0.19 & 0.64 & 0.67 & 0.33 & 0.50 \\
     \bottomrule
  \end{tabular}
  \caption{Total number of errors that can be attributed to wrong constituency parses \textit{(first number)} compared to rules not covering the respective sentence's structure \textit{(second number)} (Error 1: additional parts; Error 2: missing parts; Error 3: morphological errors; Error 4: wrong split point; Error 5: wrong referent; Error 6: wrong order of the syntactic elements). The last row and column \textit{(``Total'')} indicate the percentage of errors that can be attributed to erroneous constituency parse trees.}
  \label{tab:error_analysis2}
\end{table}





\begin{table}[!htb]
  \centering
    \small
  \begin{tabular}{  p{2.6cm}  c | c  c  c | c }
    \toprule 
    
    \textsc{Type} & \textsc{Rule} & \textsc{Wiki\-Large} & \textsc{New\-sela} & \textsc{Wiki\-Split} & \textsc{Total} \\ \hline\hline 
    Coordinate clauses & \textsc{Rule \#1} 
    & \textbf{38} & \textbf{123} & \textbf{1388} & \textbf{1549 (8.77\%)}\\ \hline
    
    \multirow{6}{\linewidth}{Adverbial clauses} & \textsc{Rule \#2} 
    & \textbf{40} & \textbf{123} & \textbf{832} & \textbf{995 (5.63\%)}\\ \cline{2-6}
                      & \textsc{Rule \#3} 
                      & 6 & 59 & 189 & 254 (1.44\%)\\  \cline{2-6}
                      & \textsc{Rule \#4} 
                      & 10 & 49 & 221 & 280 (1.59\%) \\  \cline{2-6}
                      & \textsc{Rule \#5} 
                      & 0 & 6 & 8 & 14 (0.08\%) \\  \cline{2-6}
                      & \textsc{Rule \#6} 
                      & 0 & 0 & 3 & 3 (0.02\%)\\  \cline{2-6}
                      & \textsc{Rule \#7} 
                      & 0 & 0 & 2 & 2 (0.01\%)\\ \hline
    
     \multirow{5}{\linewidth}{Relative clauses (non-restrictive)} & \textsc{Rule \#8} 
     & 1 & 2 & 33 & 36 (0.20\%)\\ \cline{2-6}
   & \textsc{Rule \#9} 
   & 6 & 16 & 127 & 149 (0.84\%)\\ \cline{2-6}
   & \textsc{Rule \#10} 
   & 1 & 0 & 6 & 7 (0.04\%)\\ \cline{2-6}
   & \textsc{Rule \#11} 
   & 0 & 0 & 15 & 15 (0.08\%)\\ \cline{2-6}
   & \textsc{Rule \#12} 
   & 10 & 55 & 377 & 442 (2.50\%)\\ \hline
    
    \multirow{4}{\linewidth}{Relative clauses (restrictive)} & \textsc{Rule \#13} 
    & 0 & 0 & 0 & 0 (0.00\%)\\ \cline{2-6}
    & \textsc{Rule \#14} 
    & 0 & 0 & 17 & 17 (0.10\%)\\ \cline{2-6}
    & \textsc{Rule \#15} 
    & \textbf{24} & \textbf{121} & \textbf{612} & \textbf{757 (4.29\%)}\\ \cline{2-6}
    & \textsc{Rule \#16} 
    & 7 & 35 & 254 & 296 (1.68\%)\\ \hline

     \multirow{4}{\linewidth}{Reported speech} & \textsc{Rule \#17} 
     & 2 & 33 & 38 & 73 (0.41\%)\\ \cline{2-6}
    & \textsc{Rule \#18} 
    & 0 & 4 & 0 & 4 (0.02\%)\\ \cline{2-6}
    & \textsc{Rule \#19} 
    & 0 & 22 & 0 & 22 (0.12\%) \\ \cline{2-6}
    & \textsc{Rule \#20} 
    & 3 & \textbf{99} & 140 & 242 (1.37\%) \\ \hline
    

    
    

    Coordinate verb phrases & \textsc{Rule \#21} 
    & \textbf{38} & \textbf{107} & \textbf{1268} & \textbf{1413 (8.00\%)} \\ \hline
    
     \multirow{2}{\linewidth}{Coordinate noun phrases} & \textsc{Rule \#22} 
     & \textbf{51} & \textbf{81} & \textbf{959} & \textbf{1091 (6.18\%)}\\ \cline{2-6}
    & \textsc{Rule \#23} 
    & 7 & 32 & 168 & 207 (1.17\%) \\ \hline
    
    \multirow{4}{\linewidth}{Participial phrases} & \textsc{Rule \#24} 
    & 8 & 11 & 266 & 285 (1.61\%)\\ \cline{2-6}
    & \textsc{Rule \#25} 
    &  \textbf{40} & \textbf{76} & \textbf{652} & \textbf{768 (4.35\%)} \\ \cline{2-6}
    & \textsc{Rule \#26} 
    & 23 & 38 & 405 & 466 (2.64\%)\\ \cline{2-6}
    & \textsc{Rule \#27} 
    & 4 & 7 & 156 & 167 (0.95\%)\\ \hline

    Appositions (non-restrictive) & \textsc{Rule \#28} 
    & \textbf{38} & \textbf{147} & \textbf{844} & \textbf{1029 (5.83\%)}\\ \hline
    Appositions (restrictive) & \textsc{Rule \#29} 
    & \textbf{32} & 66 & \textbf{863} & \textbf{961 (5.44\%)}\\ \hline
    
     \multirow{3}{\linewidth}{Prepositional phrases} & \textsc{Rule \#30} 
     & \textbf{117} & \textbf{421} & \textbf{2965} & \textbf{3503 (19.83\%)}\\ \cline{2-6}
    & \textsc{Rule \#31}
    & \textbf{65} & \textbf{152} & \textbf{1314} & \textbf{1531 (8.67\%)} \\ \cline{2-6}
    & \textsc{Rule \#32} 
    & 21 & 66 & 490 & 570 (3.23\%)\\ \hline

    \multirow{2}{\linewidth}{Adjectival/adverbial phrases} & \textsc{Rule \#33} 
    & 17 & 46 & 194 & 257 (1.46\%) \\ \cline{2-6}
    & \textsc{Rule \#34} 
    & 14 & 22 & 212 & 248 (1.40\%) \\ \hline

    Lead noun phrases & \textsc{Rule \#35} 
    & 0 & 0  & 2 & 2 (0.01\%) \\ \hline \hline
     
    
    Total & & 623 & 2019 & 15020 & 17662 \\ \bottomrule 

  \end{tabular}
  
  
  \caption{Rule application statistics.}
  \label{tab:rule_application_stats}
\end{table}

\subparagraph{Rule application statistics}

In addition, we calculated a frequency distribution of the grammar rules that were triggered by our \textsc{DisSim} approach during the transformation of the sentences from WikiLarge, Newsela and WikiSplit. The results are displayed in Table \ref{tab:rule_application_stats}. The ten most matched rules are shown in bold for each dataset. It becomes apparent that the transformation patterns that are applied the most often are the same for all three corpora, with the notable exception of the \textsc{SubordinationPostAttributionExtractor} rule, which is among the top ten rules for Newsela, but not for the other two datasets. This pattern encodes a form of reported speech. Since this type of speech is far more prevalent in newswire texts than in Wikipedia articles, this represents a result that was to be expected. Furthermore, the table shows that the ten most matched simplification rules (i.e. less than 30\% of the 35 manually defined grammar rules) account for about 75\% of the rule applications.

\subsubsection{Subtask 2: Establishing a Semantic Hierarchy}
\label{sec:establishing_semantic_hierarchy}

In the following section, we describe the evaluation results with regard to the second subtask, whose goal is to set up a semantic hierarchy between the split sentences.

\paragraph{Constituency type classification}

In 88.88\% of the matched sentence pairs, the hierarchical relationship that was allocated between a pair of simplified sentences by our TS approach corresponds to the nuclearity status of the aligned EDUs from RST-DT, i.e. in case of a nucleus-nucleus relationship in RST-DT, both output sentences from \textsc{DisSim} are assigned to the same context layer, while in case of a nucleus-satellite relationship the sentence mapped to the nucleus EDU is allocated to the context layer \textit{cl}, whereas the sentence mapped to the satellite span is assigned to the subordinate context layer \textit{cl}+1. 
For comparison, \textsc{PAR-s} \cite{joty2015codra}, a state-of-the-art sentence-level discourse parser, reaches a precision of 75.2\% on the RST-DT test set in the nuclearity labeling task when automatic discourse segmentation is used, i.e. when it is fed the output of its discourse segmenter. Thus, the performance of our approach with respect to the constituency type classification outperforms current intra-sentential discourse parsers.

The majority of the cases where our TS approach assigns a hierarchical relationship that differs from the nuclearity in the RST-DT corpus can be attributed to relative clauses. 
In RST-DT, given a sentence that contains a relative clause, the EDUs are typically assigned a nucleus-nucleus relationship. In contrast, we regard the information contained in relative clauses as background information that further describes the entity to which it refers.  
Therefore, we classify a simplified sentence that originates from such a type of clause as a contextual sentence that contributes additional information about its referent contained in the superordinate clause. 

\paragraph{Rhetorical relation identification}

Table \ref{tab:RST_counts_RST-DT} displays the frequency distribution of the 19 classes of rhetorical relations that were specified in \citeauthor{Taboada13} \citeyear{Taboada13} over the RST-DT corpus. The ten most frequently occurring classes make up for 89.45\% of the relations that are present in the dataset. We decided to limit ourselves to these classes in the evaluation of the rhetorical relation identification step, with two exceptions.

First, we did not take into account two of the classes included in this set, namely ``Topic-change'' and ``Same-unit''. The former encompasses relations that connect large sections of text when there is an abrupt change between topics \cite{schrimpf-2018-using}. Accordingly, this type of relation is relevant when examining larger paragraphs, but of no importance when considering intra-sentential relationships only, as is the case in our approach. The latter is not a true coherence relation. In RST-DT, it is used to link parts of units separated by embedded units
or spans \cite{Taboada13}.\footnote{For example, in the sentence \textit{``\uline{The petite, 29-year-old Ms. Johnson}, dressed in jeans and a sweatshirt, \uline{is a claims adjuster with Aetna Life Camp Casualty}.''}, a ``Same-unit'' relation holds between the two underlined text spans \cite{kibrik2005corpus}.} Second, similar to \citeauthor{benamara2015mapping} \citeyear{benamara2015mapping}, we merged the two highly related classes of ``Cause'' and ``Explanation'' into a single category, since they both indicate a causal relationship. Consequently, we ended up with a set of seven rhetorical relations, aggregating 75.74\% of the rhetorical relations included in RST-DT.

\begin{table}[!htb]
\centering
  \begin{tabular}{  l  c  c  c  }
    \toprule
    \textsc{Rhetorical Relation} & \textsc{Count} & \textsc{Percentage} & \textsc{Precision}\\ \hline
    \textbf{Elaboration} & 7,675 & 25.65\% & 0.5550 \\ 
    \textbf{Joint} & 7,116 & 23.78\% & 0.6673 \\ 
    \textbf{Attribution} & 2,984 & 9.97\% & 0.9601 \\ 
    Same-unit & 2,788 & 9.32\%  & ---\\ 
    \textbf{Contrast} & 1,522 & 5.09\% & 0.7421 \\ 
    Topic-change & 1,315 & 4.39\% & --- \\ 
    \textbf{Explanation} & 966 & 3.21\% &  \\ 
    \textbf{Cause} & 754 & 2.52\% & \multirow{-2}{*}{{0.7037}}\\ 
    \textbf{Temporal} & 964 & 3.22\% & 0.7895 \\ 
    \textbf{Background} & 897 & 2.30\% & 0.4459 \\  \hline
    
     &   &  & avg.: 0.6948\\ \bottomrule 
     
     \multicolumn{4}{p{0.96\linewidth}}{{{Evaluation (2.0\%), Enablement (1.8\%), Comparison (1.5\%), Textual organization (1.2\%), Condition (1.1\%), Topic-comment (0.9\%), Manner-means (0.7\%), Summary (0.7\%), Span (0.0\%).} Note that the \textit{Spatial} relation is not included in the RST-DT gold annotations. Therefore, it is not possible to include an analysis of the performance of our approach with respect to this type of rhetorical relation.}} \\
     
  \end{tabular} 
  
  \caption{Frequency distribution of the 19 classes of rhetorical
  relations from that are distinguished in \citeauthor{Taboada13} \citeyear{Taboada13} over the RST-DT corpus, 
  and the precision of \textsc{DisSim}'s rhetorical relation identification step.}
  \label{tab:RST_counts_RST-DT}
\end{table}

The right column in Table \ref{tab:RST_counts_RST-DT} displays the precision of our TS approach for each class of rhetorical relation when run over the sentences from RST-DT. 
 With a score of 96.01\%, the ``Attribution'' relation reaches by far the highest precision. The remaining relations, too, show decent scores, with a precision of around 70\%. The only exception is ``Background''. 
The difficulty with this type of relationship is that it signifies a very broad category that is not signalled by discourse markers and therefore hard to detect by our approach \cite{Taboada13}. 

With an average precision of 68.5\%,\footnote{The average precision refers to the scores of the selected subset of rhetorical relations only, i.e. the ones that are printed in bold in Table \ref{tab:RST_counts_RST-DT}.} our approach surpasses \textsc{PAR-s} in the relation labeling task, which achieves a precision of 66.1\% on the RST-DT corpus based on automatic discourse segmentation.\footnote{Note that \textsc{PAR-s}' precision score is computed based on the full set of 19 relation classes that are used for the annotation of the RST-DT. Consequently, its average precision score is not directly comparable to the score achieved by our approach, as they do not include exactly the same set of rhetorical relations. However, as the relations that are not considered in the evaluation of our approach make up for less than a quarter of the relations occurring in the RST-DT corpus, we assume that it still allows for a reliable approximation of the performance of the \textsc{PAR-s} baseline as compared to the approach we propose.}

\subsubsection{Extrinsic Evaluation}
\label{sec:extrinsic_eval}

In a comparative analysis, we demonstrated that the semantic hierarchy of minimal propositions benefits state-of-the-art Open IE systems in two dimensions:
    \textit{\begin{compactitem}
        \item[(a)] The canonical subject-predicate-object structure of the simplified sentences reduces the complexity of the relation extraction step, leading to an improved precision (up to 32\%) and recall (up to 30\%) of the extracted relational tuples on a large Open IE benchmark corpus \cite{Stanovsky2016benchmark}. 
        \item[(b)] The semantic hierarchy 
        in terms of intra-sentential rhetorical structures and hierarchical relationships can be leveraged to \textit{extract relational tuples within their semantic context}, resulting in more informative and coherent predicate-argument structures.
    \end{compactitem}}
In that way, the shallow semantic representation of state-of-the-art Open IE systems is transformed into a canonical context-preserving representation of relational tuples, resulting in a novel standardized output scheme for complex text data. For details, see \citeauthor{niklaus2023canonical} \citeyear{niklaus2023canonical}.

\section{Conclusion}
\label{sec:conclusion}
In this work, we present a context-preserving sentence splitting approach that transforms syntactically complex sentences into a novel two-layered representation in the form of core sentences and accompanying contexts that are semantically linked by rhetorical relations. In that way, a semantic hierarchy of propositions is created, with each of them presenting a minimal semantic unit that cannot be further decomposed into meaningful utterances.

\subsection{Contributions}
The goal of our proposed discourse-aware TS approach is to break up sentences with a complex syntax into a set of minimal propositions by decomposing clausal and phrasal elements in a recursive top-down fashion. In that way, we generate a fine-grained intermediate representation that presents a canonical structure which is easier to process for downstream Open IE applications. 

However, when splitting complex input sentences into a sequence of self-contained propositions without preserving the semantic relationship between the individual components, we end up with a set of incoherent utterances which lack important contextual information that is needed for a proper interpretation of the output. Hence, such isolated propositions easily mislead to incorrect reasoning in downstream Open IE applications. To address this challenge, we developed a novel contextual representation that puts a semantic layer on top of the simplified sentences. 

The proposed framework was built upon a transformation stage where syntactically complex sentences are recursively transformed into a hierarchical representation of simplified core and contextual sentences. For this purpose, we defined a set of transformation rules that decompose clausal and phrasal elements from a given source sentence and turn them into stand-alone sentences that present a simplified structure. Moreover, distinguishing constructs that provide key information from those that contribute only some piece of background information, we set up a contextual hierarchy between the split components. Finally, we identify the rhetorical relations that connect them using a cue phrase mapping. In that way, we establish a semantic hierarchy of simplified sentences. The framework that we presented was evaluated within our reference implementation \textsc{DisSim}. To enable reproducible research, all code is provided online.

\subsection{Summary of the Results}

Both a thorough manual analysis and an automatic evaluation across three TS datasets from two different domains demonstrate that our proposed framework outperforms the state of the art in structural TS. In a comparative analysis, we demonstrate that our TS approach achieves the highest scores on all three simplification corpora with regard to SAMSA (0.67, 0.57, 0.54), a metric targeted at automatically measuring the syntactic complexity of sentences which highly correlates with human judgments on structural simplicity and grammaticality. In addition, it comes no later than a close second in terms of SAMSA\textsubscript{abl} (0.84, 0.84, 0.84), which shows a high correlation with human ratings on meaning preservation. These findings are supported by the other scores of the automatic evaluation, as well as the human annotation ratings, suggesting that our proposed approach succeeds in splitting complex input sentences into minimal semantic units, producing a sequence of output sentences that present a simpler and more regular structure that is easy to process. 

Moreover, a detailed qualitative analysis of the 35 hand-crafted transformation patterns shows that syntactically complex sentences are split into well-formed simplified sentences in almost 80\% of the cases under consideration on average. Together with the error analysis, these investigations allow for determining starting points for future research on how to improve the performance of our approach. 
Furthermore, a comparative analysis with the annotations contained in the RST Discourse Treebank reveals that we are able to capture the contextual hierarchy between the split sentences with a precision of 89\% and reach an average precision of 69\% for the classification of the rhetorical relations that hold between them.

In an extrinsic evaluation, we investigate how the performance of Open IE systems is affected when making use of our proposed approach. For this purpose, we integrate our discourse-aware TS framework into state-of-the-art Open IE systems as a preprocessing step. In a first step, we are able to show that splitting syntactically complex sentences into minimal semantic units that present a canonical structure improves their performance by up to 32\% in precision and 30\% in recall. On top of that, we confirm that the generated relational tuples are enriched with important contextual information that helps in preserving the meaning of the input sentence in the extracted propositions. Thus, we demonstrate that the proposed fine-grained semantic representation in terms of a contextual hierarchy of simplified sentences benefits downstream Open IE tasks.

\subsection{Future Work}

In the future, we aim to improve our proposed discourse-aware TS approach with regard to previously mentioned limitations and to extend the experimental analysis that was carried out for assessing its performance. For instance, as of now, our framework operates sentence-by-sentence, i.e. when creating the semantic hierarchy of minimal propositions, it only considers intra-sentential relationships. A future research direction will be to examine how this approach can be extended to include inter-sentence relationships, resulting in a discourse tree that covers not only a single source sentence, but rather a paragraph or even a full document, similar to rhetorical structure trees generated by RST parsers.

Furthermore, we plan to develop a data-driven approach for the task of rhetorical relation identification. Following recent methods in the area of discourse parsing \cite{feng2014linear,joty2015codra}, we suggest to use a supervised learning approach that applies a rich set of features to complement the lookup of cue phrases. This may also help in identifying implicit relations that are not explicitly signalled by cue phrases. 
In addition, we intend to augment the analysis of the second subtask, i.e. the constituency type classification and rhetorical relation identification, by comparing the performance of our approach with state-of-the-art discourse parsers.

As the use of a lexicalized parser ties our approach to a specific textual domain, further research will be carried out to evaluate if our proposed framework performs well in other domains, too, such as biomedical or legal texts. 
Moreover, we plan to examine to what extent other NLP tasks, such as MT or Text Summarization, may benefit from the output produced by our approach. 
Finally, we intend to port the idea of creating a hierarchical representation of semantically-linked minimal propositions to languages other than English.

\appendix

\section*{Appendix A. Transformation Patterns}

During the transformation process,  
we operate both on the clausal and phrasal level of a sentence by recursively

\begin{itemize}
    \item \textit{splitting and rephrasing complex multi-clause sentences} into sequences of simple sentences that each contain exactly one independent clause, and
    \item \textit{extracting selected phrasal components} into stand-alone sentences. 
\end{itemize}
By connecting the decomposed sentences with information about their constituency type, a two-layered hierarchical representation in the form of core sentences and accompanying contexts is created. 
Finally, in order to preserve their semantic relationship, the rhetorical relations that hold between the split components are identified and classified. 
In that way, complex input sentences are transformed into a semantic hierarchy of minimal propositions that present a simple and regular structure. 
Thus, we create an intermediate representation which is targeted at supporting machine processing in downstream Open IE tasks whose predictive quality deteriorates with sentence length and structural complexity. 
In the following, we will discuss in detail the 35 patterns that we specified to carry out this transformation process.


\subsection*{Clausal Disembedding}
\label{sec:clausal_disembedding}

\citeauthor{collinsgrammar} \citeyear{collinsgrammar} distinguishes four central functional categories of clauses which can be divided into two combination methods: \textit{coordination} and \textit{subordination}. A \textbf{coordinate clause} is a clause that is connected to another clause with a coordinating conjunction, such as \textit{``and''} or \textit{``but''}. In contrast, a subordinate clause is a clause that begins with a subordinating conjunction, e.g. \textit{``because''} or \textit{``while''}, and which must be used together with a main clause.
The class of subordinate clauses can be subdivided into the following three groups:
\begin{itemize}
    \item[(1)] \textbf{adverbial clauses},
    \item[(2)] \textbf{relative clauses} and
    \item[(3)] \textbf{reported speech}.
\end{itemize}

All four clause types are handled by our proposed discourse-aware TS approach, as detailed below.


\subsubsection*{Coordinate Clauses} 
Coordinate clauses are two or more clauses in a sentence that have the same status and are joined by coordinating conjunctions, e.g. \textit{``and''}, \textit{``or''} or \textit{``but''} \cite{collinsgrammar}.
When given such a syntactic structure, each clause is extracted from the input and transformed into a separate simplified sentence according to the rule depicted in Figure \ref{fig:coordination_definition}. To avoid important contextual references from being lost (that may be contained, for example, in prepositional phrases at the beginning or end of the source sentence), each of the separated clauses is concatenated with the textual spans \textit{z\textsubscript{1}} and \textit{z\textsubscript{2}} that precede or follow, respectively, the coordinate clauses. In the event of a pair of coordinate clauses, the span that connects them acts as the signal span (\textit{x}) which is mapped to its corresponding rhetorical relation to infer the type of semantic relationship that holds between the decomposed elements. Since coordinate clauses commonly are of equal importance, each simplified output is labelled as a core sentence.

\begin{table}[!htb]
\footnotesize
\centering
  \begin{tabular}{  p{2cm} | p{11.5cm} }
    \toprule 
    
    \textsc{Tregex Pattern} & \cellcolor{Peach!50}{ROOT $<<:$ (S $<$ (S ?$\$..$ \underline{\dotuline{CC}} \& $\$..$ \textbf{S}))}\\ \hline
    \textsc{Extracted Sentence} & \cellcolor{Peach!25}{\textbf{S}.}\\ \hline
    
    \textsc{Example Input} & Many consider the flavor to be very agreeable, \underline{\dotuline{but}} \textbf{it is generally bitter if steeped in boiling water}.\\ \hline
    
    \textsc{Semantic Hierarchy} &
\vspace{0.25cm}
    \begin{minipage}[c]{\linewidth}
\hspace{2cm}
\begin{tikzpicture}[scale=0.9, level distance=2.4cm, sibling distance=2cm, every tree node/.style={align=center, transform shape}]
\Tree [
      .\node [style={draw,rectangle}, fill=gray!10] {COORD\\\textit{Contrast}};
            \edge node[midway, left] {core}; [.\node(a){Many consider the\\ flavor to be\\ very agreeable.};]
            \edge node[midway, right] {core}; [.\node(b){It is generally\\ bitter if steeped\\ in boiling water.};]
        ]
]
\end{tikzpicture}

\end{minipage}
    \\

    \bottomrule
    \end{tabular}
  
  \caption[The transformation rule pattern for splitting coordinate clauses.]{The transformation rule pattern for splitting coordinate clauses (\textsc{Rule \#1}). The pattern in bold represents the part of a sentence that is extracted from the source and turned into a stand-alone sentence. The underlined pattern (solid line) will be deleted from the remaining part of the input. The pattern that is underlined with a dotted line serves as a cue phrase for the rhetorical relation identification step.}
  \label{tab_pattern_coord_clauses}

\end{table}

\begin{figure}[H]
\centering
\small
\begin{examplebox}{Example: Coordinate Clauses}
Sentence: \textit{``Many consider the flavor to be very agreeable, but it is generally bitter if steeped in boiling water.''}
\tcbline
Matched Pattern:
\begin{center}
\begin{tikzpicture}[scale=1, sibling distance = 1cm, level distance = 1.3cm, every tree node/.style={align=center}]
\Tree [.ROOT
        [.S
          [.S \edge[roof]; {Many consider the\\ flavor to be very agreeable} ]
          [.CC \framebox{but} ]
          [.S \edge[roof]; {it is generally bitter if\\ steeped in boiling water} ]
          [.. . ] ] ]
\end{tikzpicture}
\end{center}
\tcbline
Extraction:
\begin{center}
\begin{tikzpicture}[scale=1, level distance=2cm, sibling distance=2.5cm, every tree node/.style={align=center}]
\Tree [.\node[style={draw,rectangle}] {\textbf{(3)}  \textit{``but''} $\rightarrow$ \textit{Contrast}}; 
  \edge node[midway, left] {\textbf{(2)} core}; {\textbf{(1)} \\Many consider the  flavor\\ to be very agreeable.}
  \edge node[midway, right] {\textbf{(2)} core}; {\textbf{(1)} \\It is generally bitter if\\ steeped in boiling water.}
]
\end{tikzpicture}
\end{center}
\end{examplebox}

\caption[Example for splitting coordinate clauses.]{Example for splitting coordinate clauses (\textsc{Rule \#1}).}
\label{fig:coordination_example}
\end{figure}
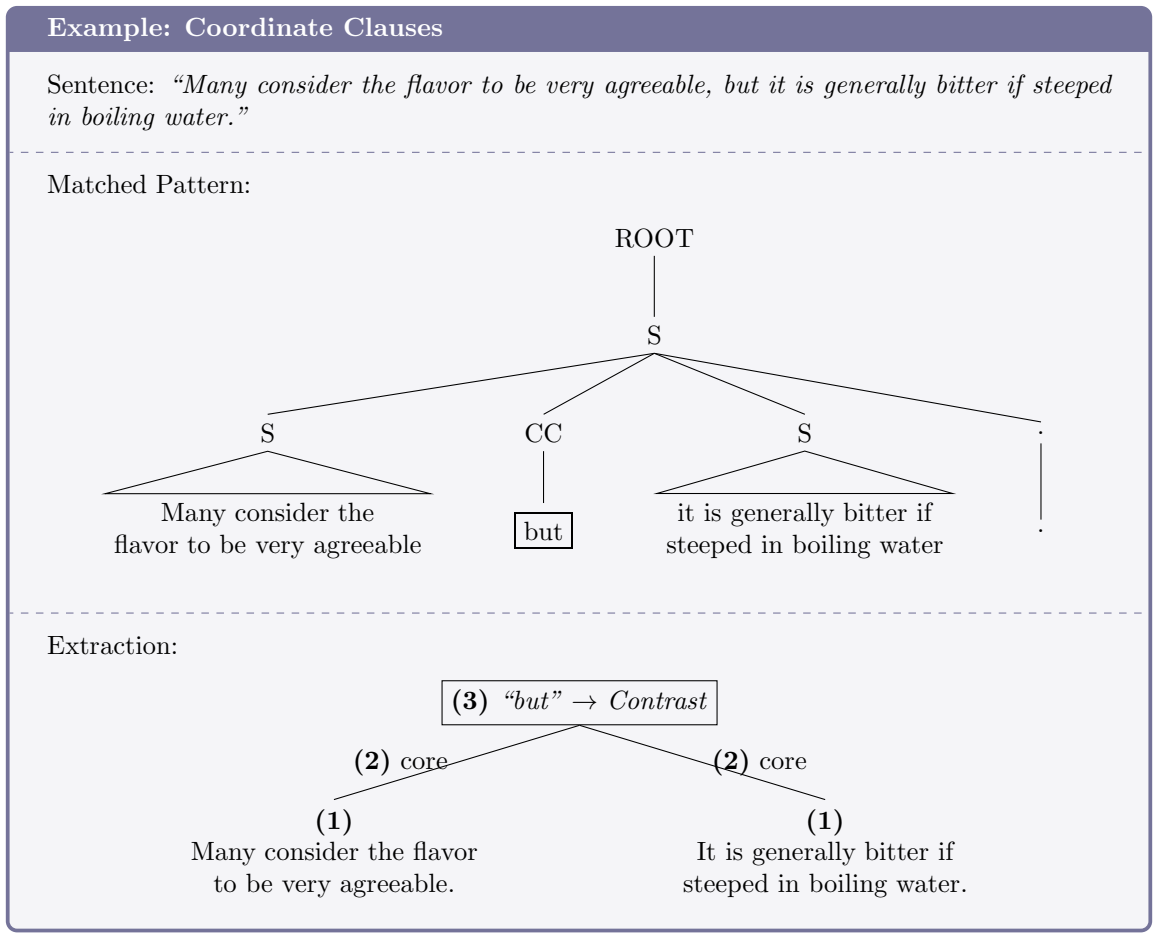

The simplification rule that defines how to split up and transform coordinate clauses is specified in terms of Tregex patterns\footnote{For details on the rule syntax, see \citeauthor{Levy2006} \citeyear{Levy2006}.} in Table \ref{tab_pattern_coord_clauses}. For a better understanding, Figure \ref{fig:coordination_definition} illustrates this rule in the form of a phrasal parse tree, and Figure \ref{fig:coordination_example} shows how an example sentence is mapped to this pattern and converted into a semantic hierarchy by (1) breaking up the input into a sequence of simplified sentences, (2) setting up a contextual hierarchy between the split components, and (3) identifying the semantic relationship that holds between them.\footnote{In the discussion of the remaining linguistic constructs, we will restrict ourselves to reporting the Tregex patterns that we specified for carrying out the transformation process.} 

\begin{figure}[H]
\centering
\small
\begin{definitionbox}{Rule: Coordinate Clauses}
Phrasal Pattern:
\begin{center}
\begin{tikzpicture}[scale=1, sibling distance = 0.5cm, level distance = 1.5cm, every tree node/.style={align=center}]
\Tree [.ROOT
        [.S
          \edge[roof, dotted]; {$z_1$}
          [.S_1 ]
          \edge[roof, dotted]; {\framebox{$x$}}
          [.S_2 ]
          \edge[roof, dotted]; {}
          [.S_n ]
          \edge[roof, dotted]; {$z_2$}
        ] 
 	]
\end{tikzpicture}
\end{center}
\tcbline
Extraction:
\begin{center}
\begin{tikzpicture}[scale=1, level distance=1.75cm, sibling distance=0.5cm, every tree node/.style={align=center}]
\Tree [.\node[style={draw,rectangle}] {Signal span: $x$ if $n=2$ else $\emptyset$}; 
  \edge node[midway, left] {}; {$z_1$ $\mathbin{\|}$ S_1 $\mathbin{\|}$ $z_2$}
  \edge node[midway, right] {}; {$z_1$ $\mathbin{\|}$ S_2 $\mathbin{\|}$ $z_2$}
  \edge[draw=none]; {...}
  \edge node[midway, right] {}; {$z_1$ $\mathbin{\|}$ S_n $\mathbin{\|}$ $z_2$}
]
\end{tikzpicture}
\end{center}
\end{definitionbox}

\caption{Rule for splitting coordinate clauses.}
\label{fig:coordination_definition}
\end{figure}
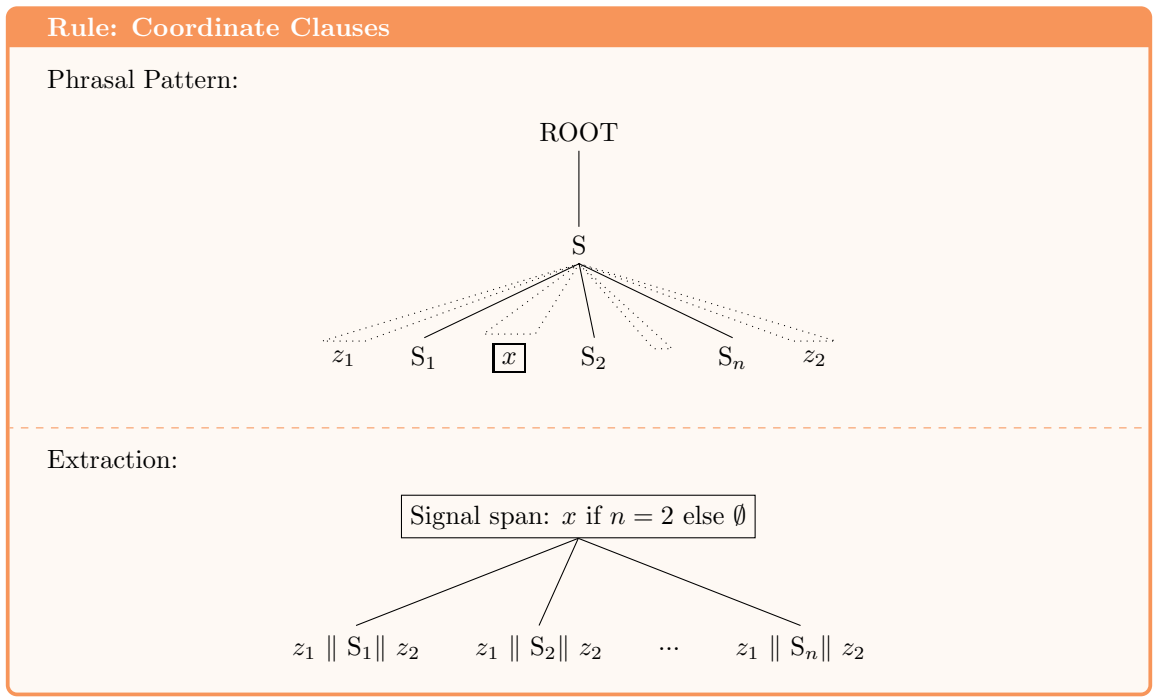

\subsubsection*{Adverbial Clauses} 
One major category of subordinate clauses are adverbial clauses. An adverbial clause is a dependent clause that functions as an adverb, adding information that describes how an action took place (e.g., when, where and how). It is fronted by a subordinating conjunction, such as \textit{``since''}, \textit{``while''}, \textit{``after''}, \textit{``if''} or \textit{``because''}. Depending on the actions or senses of their conjunctions, adverbial clauses are grouped into different classes, including \textit{time}, \textit{place}, \textit{condition}, \textit{contrast} and \textit{reason} \cite{Quir85}.

\begin{table}[!htb]
    \begin{subtable}[h]{\textwidth}
    \centering
    \footnotesize
     \begin{tabular}{  p{2cm} | p{11.5cm}  }
     \toprule
  
  \textsc{Tregex Pattern} & \cellcolor{Peach!50}{ROOT $<<:$ (S $<$ (NP $\$..$ (VP $<+$(VP) (\underline{\dotuline{SBAR}} $<$ (\textit{\textbf{S $<$ (NP $\$..$ VP)})))))}} \\ \hline
  
  \textsc{Extracted Sentence} & \cellcolor{Peach!25}{\textit{\textbf{S $<$ (NP $\$..$ VP)}}.}\\ \hline
   
  \textsc{Example Input} & Donald Trump was elected over Democratic nominee Hillary Clinton, \underline{\dotuline{although}} \textbf{he lost the popular vote}.\\ \hline
  
  \textsc{Semantic Hierarchy} & 
  \vspace{0.25cm}
    \begin{minipage}[c]{\linewidth}
\hspace{0.8cm}
\begin{tikzpicture}[scale=0.9, level distance=2.4cm, sibling distance=2cm, every tree node/.style={align=center, transform shape}]
\Tree [
      .\node [style={draw,rectangle}, fill=gray!10] {COORD\\\textit{Contrast}};
            \edge node[midway, left] {core}; [.\node(a){Donald Trump was\\ elected over Democratic\\ nominee Hillary Clinton.};]
            \edge node[midway, right] {core}; [.\node(b){He lost the popular vote.};]
        ]
]
\end{tikzpicture}

\end{minipage}
  
  \\ \bottomrule
  \end{tabular}
  \caption{\textsc{Rule \#2}: Postposed adverbial clauses.}
       \label{examplePatterns1}
  \end{subtable}
  \par\bigskip
  
  \end{table}
  
  \begin{table}[!htb]
\ContinuedFloat
  
  \begin{subtable}[h]{\textwidth}
  \centering
    \footnotesize
     \begin{tabular}{  p{2cm} | p{11.5cm} }
     \toprule
    
    \textsc{Tregex Pattern} & \cellcolor{Peach!50}{ROOT $<<:$ (S $<$ (\underline{\dotuline{SBAR}} $<$ (\textit{\textbf{S $<$ (NP $\$..$ VP)}) $\$..$ (NP $\$..$ VP)))}} \\ \hline
  
  \textsc{Extracted Sentence} & \cellcolor{Peach!25}{\textit{\textbf{S $<$ (NP $\$..$ VP)}.}}\\ \hline
  
  \textsc{Example Input} & \underline{\dotuline{Though}} \textbf{shorts are an option for many casual occasions,} they may also be inappropriate for more formal occasions. \\ \hline
  
  \textsc{Semantic Hierarchy} & 
  
  \vspace{0.25cm}
    \begin{minipage}[c]{\linewidth}
\hspace{0.5cm}
\begin{tikzpicture}[scale=0.9, level distance=2.4cm, sibling distance=2cm, every tree node/.style={align=center, transform shape}]
\Tree [
      .\node [style={draw,rectangle}, fill=gray!10] {COORD\\\textit{Contrast}};
            \edge node[midway, left] {core}; [.\node(a){They may also be inappropriate\\ for more formal occasions.};]
            \edge node[midway, right] {core}; [.\node(b){Shorts are an option\\ for many casual occasions.};]
        ]
]
\end{tikzpicture}

\end{minipage}

  \\ \bottomrule
                       \end{tabular}
  \caption{\textsc{Rule \#3}: Preposed adverbial clauses.}
  \end{subtable}
  \par\bigskip
  
  \end{table}
  
  \begin{table}[!htb]
\ContinuedFloat
  
  \begin{subtable}[h]{\textwidth}
  \centering
    \footnotesize
     \begin{tabular}{  p{2cm} | p{11.5cm}  }
     \toprule
                       
       
       \textsc{Tregex Pattern} & \cellcolor{Peach!50}{ROOT $<<:$ (S $<$ (NP $\$..$ (VP $<+$(VP) (NP$|$PP $\$..$ (\textbf{\textit{S $<<,$ (VP $<<,$ \dotuline{/(T$|$t)o/})}})))))}\\ \hline
  
  \textsc{Extracted Sentence} & \cellcolor{Peach!25}{\textit{``This'' + \textsc{Be} + \textbf{S $<<,$ (VP $<<,$ /(T$|$t)o/)}.}}\\ \hline
  
  \textsc{Example Input} & He was sent to Geneva in 1929 \textbf{\textit{\dotuline{to} act as Ireland's representative to the League of Nations.}}\\ \hline
  
  \textsc{Semantic Hierarchy} & 
  \vspace{0.25cm}
    \begin{minipage}[c]{\linewidth}
\hspace{1.6cm}
\begin{tikzpicture}[scale=0.9, level distance=2.4cm, sibling distance=2cm, every tree node/.style={align=center, transform shape}]
\Tree [
      .\node [style={draw,rectangle}, fill=gray!10] {SUBORD\\\textit{Purpose}};
            \edge node[midway, left] {core}; [.\node(a){He was sent to\\ Geneva in 1929.};]
            \edge node[midway, right] {context}; [.\node(b){This was to act as\\ Ireland's representative to\\ the League of Nations.};]
        ]
]
\end{tikzpicture}

\end{minipage}
  
  \\ \bottomrule

       \end{tabular}
  \caption{\textsc{Rule \#4}: Postposed adverbial clauses of purpose introduced by the phrase \textit{``to do''}.}
  \end{subtable}
  \par\bigskip
  \end{table}
  
  \begin{table}[!htb]
\ContinuedFloat
  
  \begin{subtable}[h]{\textwidth}
  \centering
    \footnotesize
     \begin{tabular}{  p{2cm} | p{11.5cm}  }
     \toprule

          
          \textsc{Tregex Pattern} & \cellcolor{Peach!50}{ROOT $<<:$ (S $<$ (\textbf{\textit{S $<<,$ (VP $<<,$ \dotuline{/(T$|$t)o/)}}} $\$..$ (NP $\$..$ VP)))}\\ \hline
  
  \textsc{Extracted Sentence} & \cellcolor{Peach!25}{\textit{``This'' + \textsc{Be} + \textbf{S $<<,$ (VP $<<,$ /(T$|$t)o/)}.} }\\ \hline
  
  \textsc{Example Input} & \textit{\textbf{\dotuline{To} support this claim,}} he points out the actor is wearing a very large fake moustache.\\ \hline
  
  \textsc{Semantic Hierarchy} & 
  
   \vspace{0.25cm}
    \begin{minipage}[c]{\linewidth}
\hspace{1.6cm}
\begin{tikzpicture}[scale=0.9, level distance=2.4cm, sibling distance=2cm, every tree node/.style={align=center, transform shape}]
\Tree [
      .\node [style={draw,rectangle}, fill=gray!10] {SUBORD\\\textit{Purpose}};
            \edge node[midway, left] {core}; [.\node(a){He points out the\\ actor is wearing a very\\ large fake moustache.};]
            \edge node[midway, right] {context}; [.\node(b){This is to\\ support this claim.};]
        ]
]
\end{tikzpicture}

\end{minipage}

  \\ \bottomrule

       \end{tabular}
  \caption{\textsc{Rule \#5}: Preposed adverbial clauses of purpose introduced by the phrase \textit{``to do''}.}
  \end{subtable}
  \par\bigskip
  
  \end{table}
  
  \begin{table}[!htb]
\ContinuedFloat
  
  \begin{subtable}[h]{\textwidth}
  \centering
    \footnotesize
     \begin{tabular}{  p{2cm} | p{11.5cm} }
     \toprule
                       
      
      \textsc{Tregex Pattern} & \cellcolor{Peach!50}{ROOT $<<:$ (S $<$ (NP $\$..$ (VP $<+$(VP) (\textbf{\textit{\dotuline{SBAR} $<$ (S $<<,$ (VP $<<,$ \dotuline{/(T$|$t)o/})}})))))}\\ \hline
  
  \textsc{Extracted Sentence} & \cellcolor{Peach!25}{\textit{``This'' + \textsc{Be} + \textbf{SBAR $<$ (S $<<,$ (VP $<<,$ /(T$|$t)o/))}.}}\\ \hline
  
  \textsc{Example Input} & A graduate student is researching the evolution of human eyes, \textbf{\textit{\dotuline{in order to} discredit creationists by proving that eyes have evolved.}}\\ \hline
  
  \textsc{Semantic Hierarchy} &

    \vspace{0.25cm}
    \begin{minipage}[c]{\linewidth}
\hspace{1.2cm}
\begin{tikzpicture}[scale=0.9, level distance=2.4cm, sibling distance=2cm, every tree node/.style={align=center, transform shape}]
\Tree [
      .\node [style={draw,rectangle}, fill=gray!10] {SUBORD\\\textit{Purpose}};
            \edge node[midway, left] {core}; [.\node(a){A graduate student\\ is researching the\\ evolution of human eyes.};]
            \edge node[midway, right] {context}; [.\node(b){This is to discredit\\ creationists by proving\\ that eyes have evolved.};]
        ]
]
\end{tikzpicture}

\end{minipage}
  
  \\ \bottomrule

      \end{tabular}
  \caption{\textsc{Rule \#6}: Postposed adverbial clauses of purpose introduced by the phrase \textit{``in order to do''}.}
  \end{subtable}
  \par\bigskip
  \end{table}
  
  \begin{table}[!htb]
\ContinuedFloat
  \begin{subtable}[h]{\textwidth}
  \centering
    \footnotesize
     \begin{tabular}{  p{2cm} | p{11.5cm} }
     \toprule
                       
      
      \textsc{Tregex Pattern} & \cellcolor{Peach!50}{ROOT $<<:$ (S $<$ (\textbf{\textit{\dotuline{SBAR} $<$ (S $<<,$ (VP $<<,$ \dotuline{/(T$|$t)o/}))}} $\$..$ (NP $\$..$ VP)))} \\ \hline
  
  \textsc{Extracted Sentence} & \cellcolor{Peach!25}{\textit{``This'' + \textsc{Be} + \textit{\textbf{SBAR $<$ (S $<<,$ (VP $<<,$ /(T$|$t)o/))}}.}} \\ \hline
  
  \textsc{Example Input} & \textit{\textbf{\dotuline{In order to} cater for large items and fast loading,}} the entire tail section was hinged.\\ \hline
  
  \textsc{Semantic Hierarchy} &

    \vspace{0.25cm}
    \begin{minipage}[c]{\linewidth}
\hspace{1.2cm}
\begin{tikzpicture}[scale=0.9, level distance=2.4cm, sibling distance=2cm, every tree node/.style={align=center, transform shape}]
\Tree [
      .\node [style={draw,rectangle}, fill=gray!10] {SUBORD\\\textit{Purpose}};
            \edge node[midway, left] {core}; [.\node(a){The entire tail\\ section was hinged.};]
            \edge node[midway, right] {context}; [.\node(b){This was to cater for\\ large items and fast loading.};]
        ]
]
\end{tikzpicture}

\end{minipage}
  
  \\ \bottomrule

    \end{tabular}
    \caption{\textsc{Rule \#7}: Preposed adverbial clauses of purpose introduced by the phrase \textit{``in order to do''}.}
    \end{subtable}
  
  \caption[The transformation rule patterns addressing adverbial clauses.]{The transformation rule patterns addressing adverbial clauses. A pattern in bold represents the part of the input that is extracted from the source and rephrased into a self-contained sentence. An underlined pattern (solid line) will be deleted from the remaining part of the input. The pattern that is underlined with a dotted line serves as a cue phrase for the rhetorical relation identification step. An italic pattern will be labelled as a context sentence.}
  \label{examplePatterns_adverbial}

\end{table}

Sentences containing an adverbial clause are split into two simplified components. One of them corresponds to the superordinate statement, whereas the other embodies the subordinate clause. Syntactic variations of such structures cover the following linguistic expressions, differing in the order of their clausal components:
\begin{itemize}
    \item[(1)] the subordinate clause follows the superordinate span;
    \item[(2)] the subordinate clause precedes the superordinate span; or
    \item[(3)] the subordinate clause is positioned between discontinuous parts of the superordinate span.
\end{itemize}

The subordinate clause is always introduced with a discourse connective \cite{prasad2007penn}. It serves as a cue phrase that is mapped to its corresponding rhetorical relation in order to infer the type of semantic relationship that holds between the decomposed elements.\footnote{\textsc{Rule} \#4 to \#7 are lexicalized on the preposition \textit{``to''} and always mapped to a ``Purpose'' relationship.} Since the sentence originating from the adverbial clause typically presents a piece of background information, it is marked with a context label, while the one corresponding to its superordinate span is tagged as a core sentence. 
The full set of Tregex patterns that were specified in order to transform sentences containing adverbial clauses into a simplified hierarchical representation are displayed in Table \ref{examplePatterns_adverbial}.




\subsubsection*{Relative Clauses} 

A relative clause is a clause that is attached to its antecedent by a relative pronoun, such as \textit{``who''}, \textit{``which''} or \textit{``where''}. There are two types of relative clauses, differing in the semantic relation between the clause and the phrase to which it refers: restrictive and non-restrictive. In the former case, the relative clause is strongly connected to its antecedent, providing information that identifies the phrasal component it modifies (e.g., \textit{``Obama criticized \textbf{leaders} \uline{who refuse to step off}.''}) Hence, it supplies essential information and therefore cannot be eliminated without affecting the meaning of the sentence. In contrast, non-restrictive relative clauses are parenthetic comments which describe, but do not further define their antecedent (e.g., \textit{``Obama brought attention to the \textbf{New York City Subway System}, \uline{which was in a bad condition at the time}.''}), and thus can be left out without disrupting the meaning or structure of the sentence \cite{Quir85}. As non-restrictive relative clauses are set off by commas, unlike their restrictive counterparts, they can be easily distinguished from one another on a purely syntactic basis.


\begin{table}[!htb]
    \begin{subtable}[h]{\textwidth}
    \centering
    \footnotesize
     \begin{tabular}{  p{2cm} | p{11.5cm} }
     \toprule
  
  \textsc{Tregex Pattern} & \cellcolor{Peach!50}{ROOT $<<:$ (S $<<$ (NP $<,$ \fbox{NP} $\& <$ (\underline{/,/ $\$+$ (SBAR $<,$ (WHPP} $\$+$ \textbf{\textit{S $\& <,$ IN}} $\& <-$ \underline{WHNP) \& $?\$+$ /,/}))))} \\ \hline
  
  \textsc{Extracted Sentence} & \cellcolor{Peach!25}{\textit{\textbf{S} + \textbf{IN} + \fbox{NP}.}} \\ \hline
  
  \textsc{Example Input} & He had been enrolled into \fbox{Harvard University}\underline{,} \textit{\textbf{at} \underline{which} \textbf{he studied archaeology.}} \\ \hline
  
  \textsc{Semantic Hierarchy} &

    \vspace{0.25cm}
    \begin{minipage}[c]{\linewidth}
\hspace{1.1cm}
\begin{tikzpicture}[scale=0.9, level distance=2.4cm, sibling distance=2cm, every tree node/.style={align=center, transform shape}]
\Tree [
      .\node [style={draw,rectangle}, fill=gray!10] {SUBORD\\\textit{Elaboration\textsubscript{non-defining}}};
            \edge node[midway, left] {core}; [.\node(a){He had been enrolled\\ into Harvard University.};]
            \edge node[midway, right] {context}; [.\node(b){He studied archaeology.};]
        ]
]
\end{tikzpicture}

\end{minipage}

  \\ \bottomrule
       \end{tabular}
  \caption{\textsc{Rule \#8}: Non-restrictive relative clauses commencing with a preposition followed by a relative pronoun.}
  \end{subtable}
  \par\bigskip
  \end{table}
  
  \begin{table}[!htb]
\ContinuedFloat
  
  \begin{subtable}[h]{\textwidth}
  \centering
    \footnotesize
     \begin{tabular}{  p{2cm} | p{11.5cm} }
     \toprule
      
    
    \textsc{Tregex Pattern} & \cellcolor{Peach!50}{ROOT $<<:$ (S $<<$ (/.*/ $<$ (NP$|$PP $\$+$ (\underline{/,/ $\$+$ (SBAR $<,$ (WHADVP} $\$+$ \textit{\textbf{S}} \& $<<:$ \underline{WRB) \& $?\$+$ /,/})))))} \\ \hline
  
  \textsc{Extracted Sentence} & \cellcolor{Peach!25}{\textit{\textbf{S}}.} \\ \hline
  
  \textsc{Example Input} & The noted communist, Sakhavu Kurumpakara Thankappan, was born and raised in Kurumpakara\underline{, where} \textit{\textbf{a memorial dedicated to him is situated at the Udayonmuttam Junction.}} \\ \hline
  
  \textsc{Semantic Hierarchy} &
  
   \vspace{0.25cm}
    \begin{minipage}[c]{\linewidth}
\hspace{0.4cm}
\begin{tikzpicture}[scale=0.9, level distance=2.4cm, sibling distance=1cm, every tree node/.style={align=center, transform shape}]
\Tree [
      .\node [style={draw,rectangle}, fill=gray!10] {SUBORD\\\textit{Spatial}};
            \edge node[midway, left] {core}; [.\node(a){The noted communist,\\ Sakhavu Kurumpakara, was\\ born and raised in Kurumpakara.};]
            \edge node[midway, right] {context}; [.\node(b){A memorial dedicated\\ to him is situated\\ at the Udayonmuttam Junction.};]
        ]
]
\end{tikzpicture}

\end{minipage}

  \\ \bottomrule

       \end{tabular}
  \caption{\textsc{Rule \#9}: Non-restrictive relative clauses commencing with the relative pronoun \textit{``where''}.}
  \end{subtable}
  \par\bigskip
  
  \end{table}
  
  \begin{table}[!htb]
\ContinuedFloat
  
  \begin{subtable}[h]{\textwidth}
  \centering
    \footnotesize
     \begin{tabular}{  p{2cm} | p{11.5cm} }
     \toprule
    
    
    \textsc{Tregex Pattern} & \cellcolor{Peach!50}{ROOT $<<:$ (S $<<$ (NP $<,$ \fbox{NP} $\& <$ (\underline{/,/ $\$+$ (SBAR $<,$ (WHNP} $\$+$ (\textbf{\textit{S $<,$ NP=np\_rel $\& <-$ (VP=vp\_rel $?<+$(VP) PP)}}) \& $<<:$ \underline{(WP $<:$ whom)}) \& $?\$+$ \underline{/,/}))))} \\ \hline
  
  \textsc{Extracted Sentence} & \cellcolor{Peach!25}{\textit{np\_rel + vp\_rel + \fbox{NP} + PP.}} \\ \hline
  
  \textsc{Example Input} & He is best known for his work with \fbox{The Byrds}\underline{, whom} \textit{\textbf{he joined in September 1968.}} \\ \hline
  
  \textsc{Semantic Hierarchy} &

  \vspace{0.25cm}
    \begin{minipage}[c]{\linewidth}
\hspace{1.3cm}
\begin{tikzpicture}[scale=0.9, level distance=2.4cm, sibling distance=2cm, every tree node/.style={align=center, transform shape}]
\Tree [
      .\node [style={draw,rectangle}, fill=gray!10] {SUBORD\\\textit{Elaboration\textsubscript{non-defining}}};
            \edge node[midway, left] {core}; [.\node(a){He is best known for\\ his work with The Byrds.};]
            \edge node[midway, right] {context}; [.\node(b){He joined The Byrds\\ in September 1968.};]
        ]
]
\end{tikzpicture}

\end{minipage}
  
  \\ \bottomrule

       \end{tabular}
  \caption{\textsc{Rule \#10}: Non-restrictive relative clauses commencing with the relative pronoun \textit{``whom''}.}
  \end{subtable}
  \par\bigskip
  
  \end{table}
  
  \begin{table}[!htb]
\ContinuedFloat
  
  \begin{subtable}[h]{\textwidth}
  \centering
    \footnotesize
     \begin{tabular}{  p{2cm} | p{11.5cm} }
     \toprule
    
    
    \textsc{Tregex Pattern} & \cellcolor{Peach!50}{ROOT $<<:$ (S $<<$ (NP $<$ (\fbox{NP} $\$+$ (\underline{/,/ $\$+$ (SBAR $<,$ (WHNP} $\$+$ \textbf{\textit{ S}} $\& <,$ (\underline{/WP $\backslash\backslash\$$/} $\$+$ \textit{\textbf{/.*/=subject}})) \& $?\$+$ \underline{/,/})))))} \\ \hline
  
  \textsc{Extracted Sentence} & \cellcolor{Peach!25}{\textit{\fbox{NP} + ``'s'' + \textbf{/.*/=subject} + \textbf{S}.}} \\ \hline
  
  \textsc{Example Input} & Dunn comes to establish a strong bond with \fbox{Maggie}\underline{, whose} \textit{\textbf{own family cares little for her well-being.}} \\ \hline
  
  \textsc{Semantic Hierarchy} &

  \vspace{0.25cm}
    \begin{minipage}[c]{\linewidth}
\hspace{0.6cm}
\begin{tikzpicture}[scale=0.9, level distance=2.4cm, sibling distance=2cm, every tree node/.style={align=center, transform shape}]
\Tree [
      .\node [style={draw,rectangle}, fill=gray!10] {SUBORD\\\textit{Elaboration\textsubscript{non-defining}}};
            \edge node[midway, left] {core}; [.\node(a){Dunn comes to establish\\ a strong bond with Maggie.};]
            \edge node[midway, right] {context}; [.\node(b){Maggie's own family cares\\ little for her well-being.};]
        ]
]
\end{tikzpicture}

\end{minipage}
  
  \\ \bottomrule
    
       \end{tabular}
  \caption{\textsc{Rule \#11}: Non-restrictive relative clauses commencing with the relative pronoun \textit{``whose''}.}
  \end{subtable}
  \par\bigskip
  \end{table}
  
  \begin{table}[!htb]
\ContinuedFloat
  \begin{subtable}[h]{\textwidth}
  \centering
    \footnotesize
     \begin{tabular}{  p{2cm} | p{11.5cm} }
     \toprule
    
    
    \textsc{Tregex Pattern} & \cellcolor{Peach!50}{ROOT $<<:$ (S $<<$ (NP $<,$ \fbox{NP} \& $<$ (\underline{/,/ $\$+$ (SBAR $<,$ (WHNP} $\$+$ \textbf{\textit{S}} \& $<<:$ \underline{WP$|$WDT) \& $?\$+$ /,/}))))} \\ \hline
  
  \textsc{Extracted Sentence} & \cellcolor{Peach!25}{\textit{\fbox{NP} + \textbf{S}}.} \\ \hline
  
  \textsc{Example Input} & She met \fbox{her husband}\underline{, who} \textit{\textbf{was completing his doctorate in physics.}} \\ \hline
  
  \textsc{Semantic Hierarchy} &
  
  \vspace{0.25cm}
    \begin{minipage}[c]{\linewidth}
\hspace{1.8cm}
\begin{tikzpicture}[scale=0.9, level distance=2.4cm, sibling distance=2cm, every tree node/.style={align=center, transform shape}]
\Tree [
      .\node [style={draw,rectangle}, fill=gray!10] {SUBORD\\\textit{Elaboration\textsubscript{non-defining}}};
            \edge node[midway, left] {core}; [.\node(a){She met her\\ husband.};]
            \edge node[midway, right] {context}; [.\node(b){Her husband was completing\\ his doctorate in physics.};]
        ]
]
\end{tikzpicture}

\end{minipage}
  
  \\ \bottomrule

    \end{tabular}
    \caption{\textsc{Rule \#12}: Non-restrictive relative clauses commencing with either the relative pronoun \textit{``who''} or \textit{``which''}.}
    \end{subtable}
  
  \caption[The transformation rule patterns addressing non-restrictive relative clauses.]{The transformation rule patterns addressing non-restrictive relative clauses. A pattern in bold represents the part of the input that is extracted from the source and rephrased into a self-contained sentence. A boxed pattern refers to its referent. An underlined pattern (solid line) will be deleted from the remaining part of the input. An italic pattern will be labelled as a context sentence.}
  \label{examplePatterns_nonres_rel}

\end{table}

\begin{table}[!htb]
    \begin{subtable}[h]{\textwidth}
    \centering
    \footnotesize
     \begin{tabular}{  p{2cm} | p{11.5cm} }
     \toprule

     
     \textsc{Tregex Pattern} & \cellcolor{Peach!50}{ROOT $<<:$ (S $<<$ (NP $<,$ \fbox{NP} \& $<$ (\underline{SBAR $<,$ (WHNP} $\$+$ \textit{\textbf{(S $<,$ NP \& $<-$ (VP $?<+$(VP) PP))}} \& $<<:$ \underline{(WP $<:$ whom)}))))} \\ \hline
     
     \textsc{Extracted Sentence} & \cellcolor{Peach!25}{\textit{\textbf{(S $<,$ NP \& $<-$ (VP $?<+$(VP) PP))} + \fbox{NP}.}} \\ \hline
     
     \textsc{Example Input} & \fbox{The artist} \underline{whom} \textbf{\textit{she admires}} won an award. \\ \hline
     
     \textsc{Semantic Hierarchy} &
      
  \vspace{0.25cm}
    \begin{minipage}[c]{\linewidth}
\hspace{2.5cm}
\begin{tikzpicture}[scale=0.9, level distance=2.4cm, sibling distance=2cm, every tree node/.style={align=center, transform shape}]
\Tree [
      .\node [style={draw,rectangle}, fill=gray!10] {SUBORD\\\textit{Elaboration\textsubscript{defining}}};
            \edge node[midway, left] {core}; [.\node(a){The artist won\\ an award.};]
            \edge node[midway, right] {context}; [.\node(b){She admires\\ the artist.};]
        ]
]
\end{tikzpicture}

\end{minipage}
     
     \\  \bottomrule
       \end{tabular}
  \caption{\textsc{Rule \#13}: Restrictive relative clauses commencing with the relative pronoun \textit{``whom''}.}
  \end{subtable}
  \par\bigskip
  
  \end{table}
  
  \begin{table}[!htb]
\ContinuedFloat
  
  \begin{subtable}[h]{\textwidth}
  \centering
    \footnotesize
     \begin{tabular}{  p{2cm} | p{11.5cm} }
     \toprule

     
     \textsc{Tregex Pattern} & \cellcolor{Peach!50}{ROOT $<<:$ (S $<<$ (NP $<$ (\fbox{NP} $\$+$ (\underline{SBAR $<,$ (WHNP} $\$+$ \textit{\textbf{S}} \& $<,$ (\underline{/ WP$\backslash\backslash\$$/} $\$+$ \textbf{\textit{/.*/=subject}}))))))}\\ \hline
     
     \textsc{Extracted Sentence} & \cellcolor{Peach!25}{\textit{\fbox{NP} + ``'s'' + \textbf{subject} + \textbf{S}.}} \\ \hline
     
     \textsc{Example Input} & The rescue operation to reach Flight 608 was carried out by \fbox{the Canadian Forces} \underline{whose} \textbf{\textit{plane spotted the downed aircraft.}} \\ \hline
     
     \textsc{Semantic Hierarchy} &
      
      \vspace{0.25cm}
    \begin{minipage}[c]{\linewidth}
\hspace{1cm}
\begin{tikzpicture}[scale=0.9, level distance=2.4cm, sibling distance=2cm, every tree node/.style={align=center, transform shape}]
\Tree [
      .\node [style={draw,rectangle}, fill=gray!10] {SUBORD\\\textit{Elaboration\textsubscript{defining}}};
            \edge node[midway, left] {core}; [.\node(a){The rescue operation to\\ reach Flight 608 was carried\\ out by the Canadian Forces.};]
            \edge node[midway, right] {context}; [.\node(b){The Canadian Forces'\\ plane spotted the\\ downed aircraft.};]
        ]
]
\end{tikzpicture}

\end{minipage}
     
     \\  \bottomrule
     
       \end{tabular}
  \caption{\textsc{Rule \#14}: Restrictive relative clauses commencing with the relative pronoun \textit{``whose''}.}
  \end{subtable}
  \par\bigskip
  
  \end{table}
  
  \begin{table}[!htb]
\ContinuedFloat
  
  \begin{subtable}[h]{\textwidth}
  \centering
    \footnotesize
     \begin{tabular}{  p{2cm} | p{11.5cm} }
     \toprule

     
     \textsc{Tregex Pattern} & \cellcolor{Peach!50}{ROOT $<<:$ (S $<<$ (NP $<,$ (\fbox{NP} $\$++$ (\underline{SBAR $<,$ (WHNP} $\$+$ \textbf{\textit{S}} \& $<<:$ \underline{WP$|$WDT) \& $?\$+$ /,/}))))} \\ \hline
     
     \textsc{Extracted Sentence} & \cellcolor{Peach!25}{\textit{\fbox{NP} + \textit{\textbf{S}}}.} \\ \hline
     
     \textsc{Example Input} & Ishak Belfodil is \fbox{a Franco-Algerian football player} \underline{who} \textbf{\textit{currently plays for French club Olympique Lyonnais in Ligue 1.}} \\ \hline
     
     \textsc{Semantic Hierarchy} &
      \vspace{0.25cm}
    \begin{minipage}[c]{\linewidth}
\hspace{0.6cm}
\begin{tikzpicture}[scale=0.9, level distance=2.4cm, sibling distance=2cm, every tree node/.style={align=center, transform shape}]
\Tree [
      .\node [style={draw,rectangle}, fill=gray!10] {SUBORD\\\textit{Elaboration\textsubscript{defining}}};
            \edge node[midway, left] {core}; [.\node(a){Ishak Belfodil is a\\ Franco-Algerian\\ football player.};]
            \edge node[midway, right] {context}; [.\node(b){This Franco-Algerian football player\\ currently plays for French\\ club Olympique Lyonnais in Ligue 1.};]
        ]
]
\end{tikzpicture}

\end{minipage}
     
     \\  \bottomrule

       \end{tabular}
  \caption{\textsc{Rule \#15}: Restrictive relative clauses commencing with either the relative pronoun \textit{``who''}, \textit{``which''} or \textit{``that''}.}
  \end{subtable}
  \par\bigskip

  \end{table}
  
  \begin{table}[!htb]
\ContinuedFloat
  
  \begin{subtable}[h]{\textwidth}
  \centering
    \footnotesize
     \begin{tabular}{  p{2cm} | p{11.5cm} }
     \toprule

      \textsc{Tregex Pattern} & \cellcolor{Peach!50}{ROOT $<<:$ (S $<<$ (NP $<,$ (\fbox{NP} $\$++$ (\underline{SBAR} $<:$ \textit{\textbf{(S $<$ (VP $?<$ (PP $?<:$ IN)))}}))))}\\ \hline
     
     \textsc{Extracted Sentence} & \cellcolor{Peach!25}{\textit{\textbf{(S $<$ (VP $?<$ (PP $?<:$ IN)))} + \fbox{NP}.}} \\ \hline
     
     \textsc{Example Input} & \fbox{The novelist} \textit{\textbf{she adores}} published a new book.\\ \hline
     
     \textsc{Semantic Hierarchy} &
      
     \vspace{0.25cm}
    \begin{minipage}[c]{\linewidth}
\hspace{1.8cm}
\begin{tikzpicture}[scale=0.9, level distance=2.4cm, sibling distance=2cm, every tree node/.style={align=center, transform shape}]
\Tree [
      .\node [style={draw,rectangle}, fill=gray!10] {SUBORD\\\textit{Elaboration\textsubscript{defining}}};
            \edge node[midway, left] {core}; [.\node(a){The novelist published\\ a new book.};]
            \edge node[midway, right] {context}; [.\node(b){She adores\\ the novelist.};]
        ]
]
\end{tikzpicture}

\end{minipage}
     
     \\  \bottomrule
     
    
    \end{tabular}
    \caption{\textsc{Rule \#16}: Reduced relative clauses, which are not marked by an explicit relative pronoun.}
    \end{subtable}
  
  \caption[The transformation rule patterns addressing restrictive relative clauses.]{The transformation rule patterns addressing restrictive relative clauses. A pattern in bold represents the part of the input that is extracted from the source and rephrased into a self-contained sentence. A boxed pattern refers to its referent. An underlined pattern (solid line) will be deleted from the remaining part of the input. An italic pattern will be labelled as a context sentence.}
  \label{examplePatterns_res_rel}

\end{table}


In order to identify whether a given sentence includes a non-restrictive relative clause, we check if one of the Tregex patterns of Table \ref{examplePatterns_nonres_rel} matches its phrasal parse tree. If so, a self-contained context sentence providing additional information about the referred phrase is constructed by linking the relative clause (without the relative pronoun) to the phrase that has been identified as its antecedent. At the same time, the source sentence is reduced to its key information by dropping the extracted relative clause, and tagged as a core sentence. In this way, subordinations that are introduced by one of the relative pronouns \textit{``who''}, \textit{``whom''}, \textit{``whose''}, \textit{``which''} or \textit{``where''}, as well as a combination of a preposition and one of the pronouns mentioned before are handled by our approach. 

As opposed to non-restrictive relative clauses, their restrictive counterparts represent an integral part of the phrase to which they are linked. Regardless, according to our goal of splitting complex input sentences into minimal propositions, we decided to decompose the restrictive version of relative clauses as well, and transform them into stand-alone context sentences according to the rules listed in Table \ref{examplePatterns_res_rel}. To clarify that the detached components provide irremissible information about their referent, we insert the adjunct ``defining'' when we add the semantic relationship between the split sentences in the form of a rhetorical relation. Both types of relative clauses establish an ``Elaboration'' relation, since they present additional detail about the entity to which they refer.\footnote{The only exception are relative clauses commencing with the relative pronoun \textit{``where''} (\textsc{Rule \#9}). As they provide details about a location, we assign a ``Spatial'' relationship between the split sentences in this case.} For the sake of clarity, we append the adjunct ``non-defining'' in case of a non-restrictive relative clause.

\subsubsection*{Reported Speech}
Special attention was paid to attribution relationships expressed in reported speech, e.g. \textit{``Obama \uline{announced} [that he would resign his Senate seat.]$_{SBAR}$}'', which, too, fall into the syntactic category of subordinations. To distinguish this type of linguistic expression from adverbial clauses, we defined a set of additional rule patterns that target the detection of attributions. Similar to \citeauthor{Mausam12} \citeyear{Mausam12}, 
we identify attributions by matching the lemmatized version of the head verb of the sentence (here: \textit{``announce''}) against a list of verbs of reported speech and cognition \cite{carlson2001discourse}.


\begin{table}[!htb]
    \begin{subtable}[h]{\textwidth}
    \centering
    \footnotesize
     \begin{tabular}{  p{2cm} | p{11.5cm} }
     \toprule
    
    
    
    
    \textsc{Tregex Pattern} & \cellcolor{Peach!50}{ROOT $<<:$ (S $<$ (S$|$SBAR$|$SBARQ $\$..$ \textbf{\textit{(NP [\$,, \dotuline{VP=vp} $|$ $\$..$ \dotuline{VP=vp}])}}));
    
    \textit{vb} is an attribution-verb} \\ \hline
    
    \textsc{Extracted Sentence} & \cellcolor{Peach!25}{\textit{``This'' + \textsc{Be} + ``what'' + \textbf{(NP [\$,, \dotuline{VP=vp} $|$ $\$..$ \dotuline{VP=vp}])}.}} \\ \hline
    
    \textsc{Example Input} &  Witness memories don't get better with time\underline{,} \textit{\textbf{she \dotuline{said} in an interview with the International Herald Tribune}.} \\ \hline
    
    \textsc{Semantic Hierarchy} &
     
     \vspace{0.25cm}
    \begin{minipage}[c]{\linewidth}
\hspace{0.8cm}
\begin{tikzpicture}[scale=0.9, level distance=2.4cm, sibling distance=2cm, every tree node/.style={align=center, transform shape}]
\Tree [
      .\node [style={draw,rectangle}, fill=gray!10] {SUBORD\\\textit{Attribution}};
            \edge node[midway, left] {core}; [.\node(a){Witness memories don't\\ get better with time.};]
            \edge node[midway, right] {context}; [.\node(b){This is what she said\\ in an interview with the\\ International Herald Tribune.};]
        ]
]
\end{tikzpicture}

\end{minipage}

    \\ \bottomrule
       \end{tabular}
  \caption{\textsc{Rule \#17}: Reported speech with postposed attribution.}
  \end{subtable}
  \par\bigskip

  \end{table}
  
  \begin{table}[!htb]
\ContinuedFloat
  
  \begin{subtable}[h]{\textwidth}
  \centering
    \footnotesize
     \begin{tabular}{  p{2cm} | p{11.5cm} }
     \toprule
    
    
    
    
    \textsc{Tregex Pattern} &  \cellcolor{Peach!50}{ROOT $<<:$ (\textbf{\textit{S $<$ (NP $\$..$ (VP $<+$(VP)}} (\underline{SBAR} [\dotuline{,, /``/=start $|$ $<<,$ /``/=start}] [\dotuline{.. /''/=end $|$ $<<-$ /''/=end}]))))}\\ \hline
    
    \textsc{Extracted Sentence} & \cellcolor{Peach!25}{\textit{``This'' + \textsc{Be} + ``what'' + \textbf{S $<$ (NP $\$..$ (VP $<+$(VP)}}.} \\ \hline
    
    \textsc{Example Input} & \textit{\textbf{Pauli remarked sadly}} \dotuline{``}It is not even wrong\dotuline{''}. \\ \hline
    
    \textsc{Semantic Hierarchy} &
    
     \vspace{0.25cm}
    \begin{minipage}[c]{\linewidth}
\hspace{1.2cm}
\begin{tikzpicture}[scale=0.9, level distance=2.4cm, sibling distance=2cm, every tree node/.style={align=center, transform shape}]
\Tree [
      .\node [style={draw,rectangle}, fill=gray!10] {SUBORD\\\textit{Attribution}};
            \edge node[midway, left] {core}; [.\node(a){``It is not even wrong.''};]
            \edge node[midway, right] {context}; [.\node(b){This was what\\ Pauli remarked sadly.};]
        ]
]
\end{tikzpicture}

\end{minipage}
    
    \\ \bottomrule
       \end{tabular}
  \caption{\textsc{Rule \#18}: Direct speech with preposed attribution.}
  \end{subtable}
  \par\bigskip
  
  \end{table}
  
  \begin{table}[!htb]
\ContinuedFloat
  
  \begin{subtable}[h]{\textwidth}
  \centering
    \footnotesize
     \begin{tabular}{  p{2cm} | p{11.5cm} }
     \toprule
                       
    
    

    \textsc{Tregex Pattern} & \cellcolor{Peach!50}{ROOT $<<:$ (S $<$ (S$|$SBAR$|$SBARQ \dotuline{[,, /``/=start $|$ $<<,$ /``/=start] [.. /''/=end $|$ $<<-$ /''/=end]} $\$..$ \textit{\textbf{(NP [\$,, VP $|$ $\$..$ VP])}}))} \\ \hline
    
    \textsc{Extracted Sentence} & \cellcolor{Peach!25}{\textit{``This'' + \textsc{Be} + ``what'' + \textbf{(NP [\$,, VP $|$ $\$..$ VP])}.}}\\ \hline
    
    \textsc{Example Input} & \dotuline{``}I love you\dotuline{''}\underline{,} \textit{\textbf{he said}}.\\ \hline
    
    \textsc{Semantic Hierachy} & 
    
    \vspace{0.25cm}
    \begin{minipage}[c]{\linewidth}
\hspace{2.1cm}
\begin{tikzpicture}[scale=0.9, level distance=2.4cm, sibling distance=2cm, every tree node/.style={align=center, transform shape}]
\Tree [
      .\node [style={draw,rectangle}, fill=gray!10] {SUBORD\\\textit{Attribution}};
            \edge node[midway, left] {core}; [.\node(a){``I love you.''};]
            \edge node[midway, right] {context}; [.\node(b){This was what he said.};]
        ]
]
\end{tikzpicture}

\end{minipage}

    \\ \bottomrule
       \end{tabular}
  \caption{\textsc{Rule \#19}: Direct speech with postposed attribution.}
  \end{subtable}
  \par\bigskip\end{table}
  
  \begin{table}[!htb]
\ContinuedFloat

  \begin{subtable}[h]{\textwidth}
  \centering
    \footnotesize
     \begin{tabular}{  p{2cm} | p{11.5cm} }
     \toprule

    
    
    
    
    \textsc{Tregex Pattern} & \cellcolor{Peach!50}{ROOT $<<:$ (\textbf{\textit{S $<$ (NP $\$..$ (\dotuline{VP=vp} $<+$(VP)}} (\underline{SBAR} $<$ S))));
    
    \textit{vb} is an attribution verb} \\ \hline
    
    \textsc{Extracted Sentence} & \cellcolor{Peach!25}{\textit{``This'' + \textsc{Be} + ``what'' + \textbf{S $<$ (NP $\$..$ (\dotuline{VP=vp} $<+$(VP)}.}} \\ \hline
    
    \textsc{Example Input} & \textit{\textbf{Ellis \dotuline{claimed}}} \underline{that} the character was not based on his father. \\ \hline
    
    \textsc{Semantic Hierarchy} &
    
    \vspace{0.25cm}
    \begin{minipage}[c]{\linewidth}
\hspace{1.7cm}
\begin{tikzpicture}[scale=0.9, level distance=2.4cm, sibling distance=2cm, every tree node/.style={align=center, transform shape}]
\Tree [
      .\node [style={draw,rectangle}, fill=gray!10] {SUBORD\\\textit{Attribution}};
            \edge node[midway, left] {core}; [.\node(a){The character was not\\ based on his father.};]
            \edge node[midway, right] {context}; [.\node(b){This was what\\ Ellis claimed.};]
        ]
]
\end{tikzpicture}

\end{minipage}

    \\ \bottomrule
    \end{tabular}
    \caption{\textsc{Rule \#20}: Reported speech with preposed attribution.}
    \end{subtable}
  
  \caption[The transformation rule patterns addressing reported speech constructs.]{The transformation rule patterns addressing reported speech constructs. A pattern in bold represents the part of the input that is extracted from the source and rephrased into a self-contained sentence. An underlined pattern (solid line) will be deleted from the remaining part of the input. A pattern that is underlined with a dotted line serves as a cue phrase for the rhetorical relation identification step. An italic pattern will be labelled as a context sentence.}
  \label{examplePatterns_reportedSpeech}

\end{table}

The Tregex patterns for splitting up and turning reported speech constructs into a two-layered hierarchy of simplified sentences are listed in Table \ref{examplePatterns_reportedSpeech}. The decomposed parts are always connected via an ``Attribution'' relation. Moreover, note that there is a peculiarity here with regard to the contextual hierarchy. As opposed to previously mentioned cases, where subordinations are marked up as context sentences, while the corresponding superordinate spans are labelled as core sentences, this relationship is inverted now. Since we consider the actual statement (e.g., \textit{``He would resign his Senate seat.''}) to be more meaningful than the fact that this was pronounced by some entity (e.g., \textit{``This was what Obama announced.''}), we tag the subordinate clause containing the uttered proposition as core, while we classify the superordinate attribution assertion as context. 
Also note that the syntactic pattern of \textsc{Rule \#20} matches the rule that addresses postposed adverbial clauses, i.e. \textsc{Rule \#2}. Thus, to ensure that the attribution pattern is favoured in the presence of a verb of reported speech or cognition, \textsc{Rule \#20} is triggered before \textsc{Rule \#2} in the order of execution of the transformation patterns that we specified. 


\subsection*{Phrasal Disembedding}
\label{sec:phrasal_disembedding}


Clausal disembedding, which is achieved by decomposing coordinate clauses, adverbial clauses, relative clauses and reported speech constructs (see Section \ref{sec:clausal_disembedding}), results in simple sentences, i.e. sentences that each contain exactly one independent clause. However, such sentences may still present a rather complex structure that mixes multiple semantic units. For instance, when simplifying the following input sentence on the clausal level

\begin{quote}
    \textit{``After graduating from Columbia University in 1983 with a degree in political science, Barack Obama worked as a commmunity organizer in Chicago.''}
\end{quote}

we end up with two decomposed spans:
\begin{itemize}
    \item \textit{Barack Obama worked as a community organizer in Chicago.}
    \item \textit{Barack Obama was graduating from Columbia University in 1983 with a degree in political science.}
\end{itemize}

Hence, in order to split the input into sentences where each span represents an atomic unit that cannot be further decomposed into meaningful propositions, the TS approach that we propose incorporates phrasal disembedding. According to our goal of generating minimal semantic units, with each of them expressing a complete and indivisible thought, it ensures that phrasal components, too, are transformed into self-contained sentences, in addition to splitting and rephrasing clausal elements, as described above. The resulting semantic hierarchy of the example sentence is displayed in Figure \ref{fig:example_clausal_phrasal_disembedding}.

\begin{figure}[htb!]
  \flushleft
  \small
\begin{BVerbatim}
(1) #1    0    Barack Obama worked as a community organizer.
(1a)               S:TEMPORAL              This was in Chicago.
(1b)               L:TEMPORAL_BEFORE       #2
(2) #2    1    Barack Obama was graduating from Columbia University.
(2a)               S:TEMPORAL              This was in 1983.
(2b)               S:ELABORATION           This was with a degree in political science.
\end{BVerbatim}
  \caption{Semantic hierarchy of minimal propositions after clausal and phrasal disembedding.}
  \label{fig:example_clausal_phrasal_disembedding}
\end{figure}

In total, our TS approach addresses seven types of phrasal constructs, including
\begin{itemize}
    \item[(1)] \textbf{coordinate verb phrases}, 
    \item[(2)] \textbf{coordinate noun phrase lists}, 
    \item[(3)] \textbf{participial phrases}, 
    \item[(4)] \textbf{appositive phrases}, 
    \item[(5)] \textbf{prepositional phrases}, 
    \item[(6)] \textbf{adjectival and adverbial phrases}, as well as
    \item[(7)] \textbf{lead noun phrases}.
\end{itemize}
In the following, the transformation patterns that were specified for targeting aforementioned linguistic constructs are presented.

\subsubsection*{Coordinate Verb Phrases}

If a verb phrase is made up of multiple coordinate verb phrases, as is the case in the example sentence depicted in Table~\ref{examplePatterns_coordinateVPs}, each verb phrase is decomposed and appended to the shared noun phrase to which they refer, thus generating two or more simplified sentences with reduced sentence length. In this way, we aim to increase the minimality of the resulting propositions in the sense that each semantic unit that is expressed in the individual verb phrases ends up in a separate output sentence.
Analogous to the rule for splitting and rephrasing coordinate clauses, the textual span that links two coordinate verb phrases is considered as the signal span which is mapped to its corresponding rhetorical relation. Since coordinate verb phrases are of equal status, all of them are labeled as core sentences.

\begin{table}[!htb]
\footnotesize
\centering
  \begin{tabular}{  p{2cm} | p{11.5cm} }
    \toprule
    
     \textsc{Tregex Pattern} & \cellcolor{Peach!50}{ROOT $<<:$ (S $<$ (\fbox{NP} $\$..$ (VP $<+$(VP) (VP $>$ VP ?$\$..$ \underline{\dotuline{CC}} \& $\$..$ \textbf{VP}))))} \\ \hline
     
  \textsc{Extracted Sentence} & \cellcolor{Peach!25}{\fbox{NP} + \textbf{VP}.} \\ \hline
    
    \textsc{Example Input} & After Pearlman's bankruptcy, \fbox{the company} emerged unscathed \underline{\dotuline{and}} \textbf{was sold to a Canadian company}.\\ \hline
    
    \textsc{Semantic Hierarchy} &
    
    \vspace{0.25cm}
    \begin{minipage}[c]{\linewidth}
\hspace{1cm}
\begin{tikzpicture}[scale=0.9, level distance=2.4cm, sibling distance=0.5cm, every tree node/.style={align=center, transform shape}]
\Tree [
      .\node [style={draw,rectangle}, fill=gray!10] {COORD\\\textit{List}};
            \edge node[midway, left] {core}; [.\node(a){After Pearlman's bankruptcy,\\ the company emerged\\ unscathed.};]
            \edge node[midway, right] {core}; [.\node(b){After Pearlman's bankruptcy,\\ the company was sold\\ to a Canadian company.};]
        ]
]
\end{tikzpicture}

\end{minipage}

    \\ \bottomrule
    
    \end{tabular}
  
  \caption[The transformation rule pattern for disembedding coordinate verb phrases.]{The transformation rule pattern for disembedding coordinate verb phrases (\textsc{Rule \#21}). The pattern in bold represents the part of the input that is extracted from the source and rephrased into a self-contained sentence. The boxed pattern refers to its referent. The underlined pattern (solid line) will be deleted from the remaining part of the input. The pattern that is underlined with a dotted line serves as a cue phrase for the rhetorical relation identification step.}
  \label{examplePatterns_coordinateVPs}

\end{table}

The transformation rule that defines how to perform this process is specified in terms of a Tregex pattern in Table \ref{examplePatterns_coordinateVPs}. 

\subsubsection*{Coordinate Noun Phrase Lists}

In accordance with the objective of splitting complex input sentences into a set of minimal semantic units, we compiled a set of rules for breaking up lists of coordinate noun phrases. The specified transformation rules target patterns of coordinate noun phrases within a parent noun phrase. In order to avoid inadvertently mistaking coordinate noun phrases for appositives, 
 we apply a heuristic that is given by the following regular expression: 
\begin{equation*}
    (NP)(,NP)*,?(and|or)(.+)
\end{equation*}
This pattern is matched only with the topmost noun phrases in subject and object position of the source sentence. By restricting its application to the topmost noun phrases, we compensate for parsing errors that were frequently encountered in deep nested noun phrase structures.



\begin{table}[!htb]
\footnotesize
\centering
\begin{subtable}[h]{\textwidth}
  \centering
    \footnotesize
  \begin{tabular}{  p{2cm} | p{11.5cm} }
    \toprule
  \textsc{Tregex Pattern} & \cellcolor{Peach!50}{ROOT $<<:$ (\fbox{S $<$ (NP $\$..$ (VP} $<<$ (NP=np1 $<$ (NP $?\$..$ \underline{\dotuline{CC}} \& $\$..$ \textbf{NP=np2})))))} \\ \hline
  \textsc{Extracted Sentence} & \cellcolor{Peach!25}{ROOT $<<:$ (\fbox{S $<$ (NP $\$..$ (VP} $<<$ \textbf{(NP=np2)}))).} \\ \hline
    \textsc{Example Input} & \fbox{Demola Aladekomo is} a computer engineer, a technology pioneer, an entrepreneur \underline{\dotuline{and}} \textbf{a philanthropist}. \\ \hline
    \textsc{Semantic Hierarchy} & 
    \vspace{0.25cm}
    \begin{minipage}[c]{\linewidth}
\hspace{0.01cm}
\begin{tikzpicture}[scale=0.9, level distance=2.4cm, sibling distance=0.3cm, every tree node/.style={align=center, transform shape}]
\Tree [
      .\node [style={draw,rectangle}, fill=gray!10] {COORD\\\textit{List}};
            \edge node[midway, left] {core}; [.\node(a){Demola Aladekomo\\ is a computer\\ engineer.};]
            \edge node[midway, left] {core}; [.\node(b){Demola Aladekomo\\ is a technology\\ pioneer.};]
            \edge node[midway, right] {core}; [.\node(c){Demola\\ Aladekomo is\\ an entrepreneur.};]
            \edge node[midway, right] {core}; [.\node(d){Demola\\ Aladekomo is\\ a philantropist.};]
        ]
]
\end{tikzpicture}

\end{minipage}

    \\
    \bottomrule
       \end{tabular}
  \caption{\textsc{Rule \#22}: Coordinate noun phrase lists in object position.}
  \end{subtable}
  \par\bigskip
  
  \end{table}
  
  \begin{table}[!htb]
\ContinuedFloat
  
  \begin{subtable}[h]{\textwidth}
  \centering
    \footnotesize
     \begin{tabular}{  p{2cm} | p{11.5cm} }
     \toprule
      \textsc{Tregex Pattern} & \cellcolor{Peach!50}{ROOT $<<:$ (S $<$ (NP=np1 $<$ (NP $?\$..$ \underline{\dotuline{CC}} \& $\$..$ \textbf{NP=np2}) $\$..$ \fbox{VP}))}\\ \hline
      
      \textsc{Extracted Sentence} & \cellcolor{Peach!25}{ROOT $<<:$ (S $<$ \textbf{(NP=np2)} $\$..$ \fbox{VP}).} \\ \hline
      
    
    \textsc{Example Input} & The intensity of the synthesizer rises before an organ, a bass guitar \underline{\dotuline{and}} \textbf{a piano} \fbox{enter}. \\ \hline
    \textsc{Semantic Hierarchy} & 
    
    \vspace{0.25cm}
    \begin{minipage}[c]{\linewidth}
\hspace{0.8cm}
\begin{tikzpicture}[scale=0.9, level distance=2.4cm, sibling distance=1cm, every tree node/.style={align=center, transform shape}]
\Tree [
      .\node [style={draw,rectangle}, fill=gray!10] {COORD\\\textit{List}};
            \edge node[midway, left] {core}; [.\node(a){The intensity\\ rises before\\ an organ enters.};]
            \edge node[midway, left] {core}; [.\node(b){The intensity\\ rises before\\ a bass guitar enters.};]
            \edge node[midway, right] {core}; [.\node(c){The intensity\\ rises before\\ a piano enters.};]
        ]
]
\end{tikzpicture}

\end{minipage}

    \\ \bottomrule
    \end{tabular}
    \caption{\textsc{Rule \#23}: Coordinate noun phrase lists in subject position.}
    \end{subtable}
  
  \caption[The transformation rule patterns for splitting coordinate noun phrase lists.]{The transformation rule patterns for splitting coordinate noun phrase lists. A pattern in bold represents the part of the input that is extracted from the source and rephrased into a self-contained sentence. The boxed pattern designates its referent. The underlined part (solid line) will be deleted from the remaining part of the input. The pattern that is underlined with a dotted line serves as a cue phrase for the rhetorical relation identification step.}
  \label{examplePatterns_coordinateNPs}

\end{table}


 For each identified entity, a simplified sentence is generated according to the rules depicted in Table \ref{examplePatterns_coordinateNPs}. The signal span corresponds to one of the two coordinating conjunctions \textit{``and''} or \textit{``or''} which joins the individual noun phrases, resulting in a ``List'' relationship. Since the items are linked via a coordinating conjunction, suggesting that they are of equal importance, we tag each split component with a core label. 



\subsubsection*{Participial Phrases}

When two or more actions are carried out simultaneously or immediately one after the other by the same subject, participial phrases are often used to express one of them (e.g., \textit{\uline{``Knowing that he wouldn't be able to buy food on his journey''} he took
large supplies with him.}, meaning \textit{``As he knew \dots''}). Note that participial phrases do not contain a subject of their own. Instead, both the subject and the verb of the phrase are replaced by a participle \cite{martinet1996practical}. Furthermore, a participial phrase may be introduced by adverbial connectors, such as \textit{``although''}, \textit{``after''} or \textit{``when''} \cite{abraham1985field}.

\begin{table}[!htb]
\footnotesize
\centering
\begin{subtable}[h]{\textwidth}
  \centering
    \footnotesize
  \begin{tabular}{  p{2cm} | p{11.5cm}  }
    \toprule
    
     \textsc{Tregex Pattern} & \cellcolor{Peach!50}{ROOT $<<:$ (S $<$ VP \&$<<$ (NP$|$PP $<,$ (\fbox{NP $?\$+$ PP} \& $\$++$ (\underline{/,/} $\$+$ (\textit{\textbf{VP [$<,$ (ADVP$|$PP $\$+$ VBG$|$VBN) $| <,$ VBG$|$VBN]}} \& $?\$+$ \underline{/,/}))))) } \\ \hline
  \textsc{Extracted Sentence} & \cellcolor{Peach!25}{\textit{\fbox{NP $?\$+$ PP} + \textsc{Be} + \textbf{VP [$<,$ (ADVP$|$PP $\$+$ VBG$|$VBN) $| <,$ VBG$|$VBN]}.}} \\ \hline
    
    \textsc{Example Input} & \fbox{The Metox}\underline{,} \textit{\textbf{named after its manufacturer}\underline{,} was a high frequency very sensitive radar receiver.} \\ \hline
    
    \textsc{Semantic Hierarchy} & 
     
    \vspace{0.25cm}
    \begin{minipage}[c]{\linewidth}
\hspace{1.2cm}
\begin{tikzpicture}[scale=0.9, level distance=2.4cm, sibling distance=2cm, every tree node/.style={align=center, transform shape}]
\Tree [
      .\node [style={draw,rectangle}, fill=gray!10] {SUBORD\\\textit{Elaboration\textsubscript{non-defining}}};
            \edge node[midway, left] {core}; [.\node(a){The Metox was a\\ high frequency very\\ sensitive radar receiver.};]
            \edge node[midway, right] {context}; [.\node(b){The Metox was named\\ after its manufacturer.};]
        ]
]
\end{tikzpicture}

\end{minipage}

    \\
    \bottomrule

       \end{tabular}
  \caption{\textsc{Rule \#24}: Non-restrictive embedded participial phrases.}
  \end{subtable}
  \par\bigskip
  
  \end{table}
  
  \begin{table}[!htb]
\ContinuedFloat
  
  \begin{subtable}[h]{\textwidth}
  \centering
    \footnotesize
     \begin{tabular}{  p{2cm} | p{11.5cm} }
     \toprule

     
      \textsc{Tregex Pattern} & \cellcolor{Peach!50}{ROOT $<<:$ (S $<$ VP \&$<<$ (NP|PP $<,$ (\fbox{NP} $\$+$ \textit{\textbf{(VP [$<,$ (ADVP$|$PP $\$+$ VBG$|$VBN) $|$ $<,$ VBG$|$VBN] )}}) \& [$>$ (PP $!>$ S)$|$ $>$ (VP $>$ S)]))} \\ \hline
  \textsc{Extracted Sentence} & \cellcolor{Peach!25}{\textit{\fbox{NP} + \textsc{Be} + \textbf{(VP [$<,$ (ADVP$|$PP $\$+$ VBG$|$VBN) $|$ $<,$ VBG$|$VBN] )}.}} \\ \hline
    
    \textsc{Example Input} & The Muppets at Walt Disney World is \fbox{a film} \textbf{\textit{starring Jim Henson's Muppets at Walt Disney World.}} \\ \hline
    
    \textsc{Semantic Hierarchy} & 
     
    \vspace{0.25cm}
    \begin{minipage}[c]{\linewidth}
\hspace{1.2cm}
\begin{tikzpicture}[scale=0.9, level distance=2.4cm, sibling distance=2cm, every tree node/.style={align=center, transform shape}]
\Tree [
      .\node [style={draw,rectangle}, fill=gray!10] {SUBORD\\\textit{Elaboration\textsubscript{defining}}};
            \edge node[midway, left] {core}; [.\node(a){The Muppets at Walt\\ Disney World is a film.};]
            \edge node[midway, right] {context}; [.\node(b){This film is starring\\ Jim Henson's Muppets\\ at Walt Disney World.};]
        ]
]
\end{tikzpicture}

\end{minipage}

    \\
    \bottomrule

       \end{tabular}
  \caption{\textsc{Rule \#25}: Restrictive postposed participial phrases.}
  \end{subtable}
  \par\bigskip
  \end{table}
  
  \begin{table}[!htb]
\ContinuedFloat
  
  \begin{subtable}[h]{\textwidth}
  \centering
    \footnotesize
     \begin{tabular}{  p{2cm} | p{11.5cm} }
     \toprule

    
    
     \textsc{Tregex Pattern} & \cellcolor{Peach!50}{participialNode = ``(\_\_=node [== S=s $|$ == (PP$|$ADVP $<+$(PP$|$ADVP) S=s)]) : (=s $<:$ (VP $<<,$ VBG|VBN))'';\newline\newline
     ROOT $<<:$ (S $<$ (\fbox{NP} \$.. (VP $<+$(VP) (NP|PP \$.. `` + \textbf{\textit{participialNode}} + ''))))} \\ \hline
  \textsc{Extracted Sentence} & \cellcolor{Peach!25}{\textit{\fbox{NP} + (\textsc{Have}) + \textsc{Be} + \textbf{participialNode}.}} \\ \hline
    
    \textsc{Example Input} & \fbox{He} served as chief judge from 1987 to 1994, \textbf{\textit{assuming senior status on November 2, 1995.}}\\ \hline
    
    \textsc{Semantic Hierarchy} & 
     
    \vspace{0.25cm}
    \begin{minipage}[c]{\linewidth}
\hspace{2.1cm}
\begin{tikzpicture}[scale=0.9, level distance=2.4cm, sibling distance=2cm, every tree node/.style={align=center, transform shape}]
\Tree [
      .\node [style={draw,rectangle}, fill=gray!10] {SUBORD\\\textit{Elaboration\textsubscript{non-defining}}};
            \edge node[midway, left] {core}; [.\node(a){He served as\\ chief judge from\\ 1987 to 1994.};]
            \edge node[midway, right] {context}; [.\node(b){He was assuming\\ senior status on\\ November 2, 1995.};]
        ]
]
\end{tikzpicture}

\end{minipage}

    \\
    \bottomrule

       \end{tabular}
  \caption{\textsc{Rule \#26}: Non-restrictive postposed participial phrases.}
  \end{subtable}
  \par\bigskip
  \end{table}
  
  \begin{table}[!htb]
\ContinuedFloat
  \begin{subtable}[h]{\textwidth}
  \centering
    \footnotesize
     \begin{tabular}{  p{2cm} | p{11.5cm} }
     \toprule

    

     \textsc{Tregex Pattern} & \cellcolor{Peach!50}{participialNode = ``(\_\_=node $[==$ S=s $|$ $==$ (PP$|$ADVP $<+$(PP$|$ADVP) S=s)$]$) : (=s $<:$ (VP $<<,$ VBG$|$VBN))''; \newline\newline
    
    ROOT $<<:$ (S $<$ `` + \textit{\textbf{participialNode}} + '') : (=node $\$..$ (\fbox{NP} $\$..$ VP))}\\ \hline
  \textsc{Extracted Sentence} & \cellcolor{Peach!25}{\textit{\fbox{NP} + (\textsc{Have}) + \textsc{Be} + \textbf{participialNode}.}} \\ \hline
    
    \textsc{Example Input} & 
    \textit{\textbf{\underline{\dotuline{Before}} entering politics}, \fbox{Donald Trump} was a businessman and a television personality.}\\ \hline
    
    \textsc{Semantic Hierarchy} & 
     
    \vspace{0.25cm}
    \begin{minipage}[c]{\linewidth}
\hspace{1.8cm}
\begin{tikzpicture}[scale=0.9, level distance=2.4cm, sibling distance=2cm, every tree node/.style={align=center, transform shape}]
\Tree [
      .\node [style={draw,rectangle}, fill=gray!10] {SUBORD\\\textit{Temporal-After}};
            \edge node[midway, left] {core}; [.\node(a){Donald Trump\\ was a businessman and\\ a television personality.};]
            \edge node[midway, right] {context}; [.\node(b){Donald Trump was\\ entering politics.};]
        ]
]
\end{tikzpicture}

\end{minipage}

    \\
    \bottomrule

    \end{tabular}
    \caption{\textsc{Rule \#27}: Non-restrictive preposed participial phrases.}
    \label{tab:example_participial_adverbial_connector}
    \end{subtable}
    
  \caption[The transformation rule patterns addressing participial phrases.]{The transformation rule patterns addressing participial phrases. A pattern in bold represents the part of the input that is extracted from the source and rephrased into a self-contained sentence. An underlined pattern (solid line) will be deleted from the remaining part of the input. The pattern that is underlined with a dotted line serves as a cue phrase for the rhetorical relation identification step. A boxed pattern designates its referent. An italic pattern will be labelled as a context sentence.}
  \label{examplePatterns_participial}

\end{table}

The transformation patterns that were specified for decomposing participial phrases are listed in Table \ref{examplePatterns_participial}. For each action expressed in the input, a separate simplified sentence is created. In order to generate an output that is grammatically sound, a paraphrasing stage is required where the participle has to be inflected, if necessary, and linked to the noun phrase that represents the subject it replaces.

Providing some additional piece of information about their respective referent, participial phrases take on a role similar to relative clauses in the semantics of a sentence; the only difference is that they lack a relative pronoun. Therefore, they are sometimes referred to as ``reduced relative clauses''. Hence, in analogy to relative clauses, the rephrasings resulting from the extracted participial phrases are labelled as context sentences, while the remaining part of the input is tagged as a core sentence, and the split components are connected via an ``Elaboration'' relation.\footnote{Note that participial phrases introduced by adverbial connectors, such as in the example sentence of Table \ref{tab:example_participial_adverbial_connector}, form an exception of this principle. In such a case, the adverbial connector determines the semantic relationship that holds between the decomposed spans, acting as the cue phrase for identifying the type of rhetorical relation connecting them.} Analogously to relative clauses, we distinguish between \textit{defining} ``Elaboration'' relationships for restrictive participial phrases and \textit{non-defining} ``Elaboration'' relationships for their non-restrictive counterparts.



\subsubsection*{Appositive Phrases}

An appositive is a noun phrase that further characterizes the phrase to which it refers. Similar to relative clauses, appositions can be classified as either restrictive or non-restrictive. Non-restrictive appositives are separate information units, marked by segregation through punctuation \cite{Quir85}, such as in \textit{``His main opponent was \textbf{Mitt Romney}, \underline{the former governor of Massachusetts}.''} 

\begin{table}[!htb]
\footnotesize
\centering
\begin{subtable}[h]{\textwidth}
  \centering
    \footnotesize
  \begin{tabular}{  p{2cm} | p{11.5cm} }
    \toprule

 \textsc{Tregex Pattern} & \cellcolor{Peach!50}{ROOT $<<:$ (S $<$ VP \& $<<$ (\fbox{NP} $\$+$ (\underline{/,/} $\$+$ (\textit{\textbf{NP}} $!\$$ CC $\&$ $?\$+$ \underline{/,/}))))} \\ \hline
  \textsc{Extracted Sentence} & \cellcolor{Peach!25}{\textit{\fbox{NP} + \textsc{Be} + \textbf{NP}.}} \\ \hline
    
    \textsc{Example Input} & \fbox{The president of Lithuania}\underline{,} \textbf{\textit{Antanas Smetona}}\underline{,} proposes armed resistance.\\ \hline
    
    \textsc{Semantic Hierarchy} & 
    
    \vspace{0.25cm}
    \begin{minipage}[c]{\linewidth}
\hspace{1cm}
\begin{tikzpicture}[scale=0.9, level distance=2.4cm, sibling distance=2cm, every tree node/.style={align=center, transform shape}]
\Tree [
      .\node [style={draw,rectangle}, fill=gray!10] {SUBORD\\\textit{Elaboration}};
            \edge node[midway, left] {core}; [.\node(a){The president of Lithuania\\ proposes armed resistance.};]
            \edge node[midway, right] {context}; [.\node(b){Antanas Smetona is the\\ president of Lithuania.};]
        ]
]
\end{tikzpicture}

\end{minipage}

    \\
    \bottomrule

       \end{tabular}
  \caption{\textsc{Rule \#28}: Non-restrictive appositive phrases.}
       \label{tab:non_res_appos_pattern}
  \end{subtable}
  \par\bigskip
  
  \end{table}
  
  \begin{table}[!htb]
\ContinuedFloat
  
  \begin{subtable}[h]{\textwidth}
  \centering
    \footnotesize
     \begin{tabular}{  p{2cm} | p{11.5cm} }
     \toprule


     \textsc{Tregex Pattern} & \cellcolor{Peach!50}{ \textit{\textbf{(((PRP$\backslash\backslash\$|$DT)$\backslash\backslash$ s)*(JJ$\backslash\backslash$
    s)*((NN$|$NNS$|$NNP$|$NNPS)$\backslash\backslash$
    s))+(((CC$|$IN)$\backslash\backslash$
    s)((PRP$\backslash\backslash\$|$DT)$\backslash\backslash$
    s)*(JJ$\backslash\backslash$
    s)*((NN$|$NNS$|$NNP$|$NNPS)$\backslash\backslash$
    s))*}} followed by a \fbox{named entity}}\\ \hline
  \textsc{Extracted Sentence} & \cellcolor{Peach!25}{\textit{\fbox{named entity} + \textsc{Be} + \textbf{regex}.}} \\ \hline
    
    \textsc{Example Input} & The regional government was moved from \textit{\textbf{the old Cossack capital}} \fbox{Novocherkassk} to Rostov. \\ \hline
    
    \textsc{Semantic Hierarchy} & 
    
    \vspace{0.25cm}
    \begin{minipage}[c]{\linewidth}
\hspace{1.1cm}
\begin{tikzpicture}[scale=0.9, level distance=2.4cm, sibling distance=2cm, every tree node/.style={align=center, transform shape}]
\Tree [
      .\node [style={draw,rectangle}, fill=gray!10] {SUBORD\\\textit{Elaboration}};
            \edge node[midway, left] {core}; [.\node(a){The regional government\\ was moved from\\ Novocherkassk to Rostov.};]
            \edge node[midway, right] {context}; [.\node(b){Novocherkassk was\\ the old Cossack capital.};]
        ]
]
\end{tikzpicture}

\end{minipage}

    \\
    \bottomrule

    \end{tabular}
    \caption{\textsc{Rule \#29}: Restrictive appositive phrases.}
    \label{tab:res_appos_rule}
    \end{subtable}
  
  \caption[The transformation rule patterns addressing appositive phrases.]{The transformation rule patterns addressing appositive phrases. A pattern in bold represents the part of the input that is extracted from the source and rephrased into a self-contained sentence. An underlined pattern (solid line) will be deleted from the remaining part of the input. A boxed pattern designates its referent. An italic pattern will be labelled as a context sentence.}
  \label{examplePatterns_appositions}

\end{table}

The pattern for transforming non-restrictive appositions is given in Table \ref{tab:non_res_appos_pattern}. It searches for a noun phrase whose immediate right sister is a comma that in turn has another noun phrase as its direct right sibling.
In order to avoid inadvertently mistaking coordinate noun phrases for appositives, the following heuristic is applied: 
from the phrase that is deemed an appositive by matching the pattern described above, we scan ahead, looking one after the other at its sibling nodes in the parse tree. If a conjunction \textit{``and''} or \textit{``or''} is encountered, the analysis of the appositive is rejected \cite{siddharthan2006syntactic}. In this way, we avoid wrong analyses like: \textit{``Obama has talked about using alcohol, $[_{appos}$ marijuana$]$, and cocaine.''} 


The second type of appositives, restrictive apposition, does not contain punctuation \cite{Quir85}. An example for such a linguistic construct is illustrated in the following sentence: \textit{``Joe Biden was formally nominated by \underline{former President} \textbf{Bill Clinton} as the Democratic Party candidate for vice president.''} The pattern in Table \ref{tab:res_appos_rule} shows the heuristic that was specified for rephrasing and transforming restrictive appositive phrases into a contextual hierarchy. It defines a regular expression that searches for a noun or a proper noun (or a coordinate sequence thereof), optionally with a combination of prepending adjectives, determiners, and possessive pronouns. This string must be followed by a named entity expression.

According to the rules listed in Table \ref{examplePatterns_appositions}, the appositive phrases are extracted from the input and transformed into stand-alone sentences. Representing parenthetical information, they are labelled as context sentences, while the reduced source sentences receive a core tag. 
Since appositions commonly provide additional information about the entity to which they refer, we link the split sentences by an ``Elaboration'' relation.


\subsubsection*{Prepositional Phrases}
\label{patterns:prep_phrases}

A prepositional phrase is composed of a preposition and a complement in the form of a noun phrase (e.g., \textit{``on the table'', ``in terms of money''}), a nominal \textit{wh}-clause (e.g., \textit{``from what he said''}) or a nominal \textit{-ing} clause (e.g., \textit{``by signing a peace treaty''}). They may function as a postmodifier in a noun phrase or an adverbial phrase, or act as a complement of a verb phrase or an adjective \cite{Quir85}. 

\begin{table}[!htb]
\footnotesize
\centering
\begin{subtable}[h]{\textwidth}
  \centering
    \footnotesize
  \begin{tabular}{  p{2cm} | p{11.5cm} }
    \toprule
  
   \textsc{Tregex Pattern} & \cellcolor{Peach!50}{ROOT $<<:$ (S $<+$(S$|$VP) (VP $<$ (\textbf{\textit{PP}} $\$-$ NP$|$PP)) $\& <$ VP)} \\ \hline
  \textsc{Extracted Sentence} & \cellcolor{Peach!25}{\textit{``This'' + \textsc{Be} + \fbox{PP}.}} \\ \hline
    
    \textsc{Example Input} & Brick enabled the construction of permanent buildings \textit{\textbf{in regions of India where the harsher climate precluded the use of mud bricks}}.\\ \hline
    
    \textsc{Semantic Hierarchy} &
    \vspace{0.25cm}
    \begin{minipage}[c]{\linewidth}
\hspace{0.9cm}
\begin{tikzpicture}[scale=0.9, level distance=2.4cm, sibling distance=2cm, every tree node/.style={align=center, transform shape}]
\Tree [
      .\node [style={draw,rectangle}, fill=gray!10] {SUBORD\\\textit{Spatial}};
            \edge node[midway, left] {core}; [.\node(a){Brick enabled the\\ construction of\\ permanent buildings.};]
            \edge node[midway, right] {context}; [.\node(b){This was in regions of India where\\ the harsher climate precluded\\ the use of mud bricks.};]
        ]
]
\end{tikzpicture}

\end{minipage}

    \\    
    \bottomrule

       \end{tabular}
  \caption{\textsc{Rule \#30}: Prepositional phrases that act as complements of verb phrases.}
       \label{tab:rule_prepositional_verbal}
  \end{subtable}
  \par\bigskip
  
  \end{table}
  
  \begin{table}[!htb]
\ContinuedFloat
  
  \begin{subtable}[h]{\textwidth}
  \centering
    \footnotesize
     \begin{tabular}{  p{2cm} | p{11.5cm}  }
     \toprule
      
    
     \textsc{Tregex Pattern} & \cellcolor{Peach!50}{ROOT $<<:$ (S $<,$ (\textbf{\textit{PP}} $?\$+$ \underline{/,/} $\&$ $\$++$ VP))} \\ \hline
  \textsc{Extracted Sentence} & \cellcolor{Peach!25}{\textit{``This'' + \textsc{Be} + \fbox{PP}.}} \\ \hline
    
    \textsc{Example Input} & \textbf{\textit{After his retirement in 1998}}\underline{,} he took charge as director of the Indian Institute of Science. \\ \hline
    
    \textsc{Semantic Hierarchy} &
    
    \vspace{0.25cm}
    \begin{minipage}[c]{\linewidth}
\hspace{1.2cm}
\begin{tikzpicture}[scale=0.9, level distance=2.4cm, sibling distance=2cm, every tree node/.style={align=center, transform shape}]
\Tree [
      .\node [style={draw,rectangle}, fill=gray!10] {SUBORD\\\textit{Temporal}};
            \edge node[midway, left] {core}; [.\node(a){He took charge\\ as director of the\\ Indian Institute of Science.};]
            \edge node[midway, right] {context}; [.\node(b){This was\\ after his retirement\\ in 1998.};]
        ]
]
\end{tikzpicture}

\end{minipage}
    
    \\
    \bottomrule

       \end{tabular}
  \caption{\textsc{Rule \#31}: Preposed prepositional phrases offset by commas.}
       \label{tab:rule_prepositional_preposed}
  \end{subtable}
  \par\bigskip
  \end{table}
  
  \begin{table}
  \ContinuedFloat
  \begin{subtable}[h]{\textwidth}
  \centering
    \footnotesize
     \begin{tabular}{  p{2cm} | p{11.5cm}  }
     \toprule
     
    
     \textsc{Tregex Pattern} & \cellcolor{Peach!50}{ROOT $<<:$ (S $<$ VP $\& <<$ (\underline{/,/} $\$+$ (\textbf{\textit{PP}} $?\$+$ \underline{/,/})))} \\ \hline
  \textsc{Extracted Sentence} & \cellcolor{Peach!25}{\textit{``This'' + \textsc{Be} + \fbox{PP}.}} \\ \hline
    
    \textsc{Example Input} & It later became a Roman town in the province of Africa, \textit{\textbf{before its eventual abandonment around 9th to 10th century}}. \\ \hline
    
    \textsc{Semantic Hierarchy} & 
    
    \vspace{0.25cm}
    \begin{minipage}[c]{\linewidth}
\hspace{1cm}
\begin{tikzpicture}[scale=0.9, level distance=2.4cm, sibling distance=2cm, every tree node/.style={align=center, transform shape}]
\Tree [
      .\node [style={draw,rectangle}, fill=gray!10] {SUBORD\\\textit{Temporal}};
            \edge node[midway, left] {core}; [.\node(a){It later became\\ a Roman town in\\ the province of Africa.};]
            \edge node[midway, right] {context}; [.\node(b){This was before\\ its eventual abandonment\\ around 9th to 10th century.};]
        ]
]
\end{tikzpicture}

\end{minipage}

    \\ \bottomrule

    \end{tabular}
    \caption{\textsc{Rule \#32}: Postposed and embedded prepositional phrases offset by commas.}
    \label{tab:rule_prepositional_postposed}
    \end{subtable}
  
  \caption[The transformation rule patterns addressing prepositional phrases.]{The transformation rule patterns addressing prepositional phrases. A pattern in bold represents the part of the input that is extracted from the source and rephrased into a self-contained sentence. An underlined pattern (solid line) will be deleted from the remaining part of the input. An italic pattern will be labelled as a context sentence.}
  \label{examplePatterns_prepos}

\end{table}

We defined a set of rules for decomposing two variants of prepositional phrases (see Table \ref{examplePatterns_prepos}). First, we address prepositional phrases that are offset by commas, e.g. \textit{``\uline{At the 2004 Democratic National Convention,} Obama delivered the keynote address.''} or \textit{``When they moved to Washington, D.C.\uline{, in January 2009,} the girls started at the Sidwell Friends School.''} (see the patterns in Table \ref{tab:rule_prepositional_preposed} and \ref{tab:rule_prepositional_postposed}) Such linguistic constructs typically provide some piece of background information that may be extracted from the source sentence without corrupting the meaning of the input. Second, we specified a pattern for transforming prepositional phrases that act as complements of verb phrases, such as in \textit{``Obama formally \textbf{announced} his candidacy \uline{in January 2003}.''} or \textit{``Obama \textbf{defeated} John McCain \uline{in the general election}.''} (see the pattern in Table \ref{tab:rule_prepositional_verbal}) Often, such verb phrase modifiers represent optional constituents that contribute no more than some form of additional information which can be eliminated, resulting in a simplified source sentence that is still both meaningful and grammatically sound. 

However, automatically distinguishing optional verb phrase modifiers from those that are required in order to form a syntactically and semantically well-formed sentence is challenging.
For instance, consider the following input: \textit{``Radio France is \textbf{headquartered} \uline{in Paris' 16th arrondissement}.''} 
When separating out the prepositional phrase that modifies the verb phrase, the output (\textit{``Radio France is headquartered.''}) is overly terse. To avoid this, we implemented the following heuristic: prepositional phrases that complement verb phrases are decomposed only if there is another constituent left in the object position of the verb phrase. In that way, we ensure that the resulting output presents a regular subject-predicate-object structure that is still meaningful and at the same time easy to process for downstream Open IE applications.

Representing some piece of background information, the extracted prepositional phrases are labelled as context sentences, while the remaining part of the input is tagged as a core sentence.
Since prepositional phrases commonly express either temporal or spatial information, the rhetorical relation identification step is carried out based on named entities. For this purpose, we iterate over all the tokens contained in the extracted phrase. If we encounter a named entity of the type ``LOCATION'', we join the decomposed sentences by a ``Spatial'' relationship. In case of a named entity of the class ``DATE'', the split sentences are linked through a ``Temporal'' relation.

\subsubsection*{Adjectival and Adverbial Phrases} 
\label{patterns:adjectival_adverbial_phrases}

An adjectival phrase is a phrase whose head is an adjective that is optionally complemented by a number of dependent elements. It further characterizes the noun phrase it is modifying. Similarly, an adverbial phrase consists of an adverb as its head, together with an optional pre- or postmodifying complement \cite{Brin00,Quir85}. The rules for detecting and transforming these types of linguistic constructs are listed in Table \ref{examplePatterns_adjectival} in terms of Tregex patterns.





\begin{table}[!htb]
\footnotesize
\centering
\begin{subtable}[h]{\textwidth}
  \centering
    \footnotesize
  \begin{tabular}{  p{2cm} | p{11.5cm}  }
    \toprule
   \textsc{Tregex Pattern} & \cellcolor{Peach!50}{ROOT $<<:$ (S $<,$ (\textbf{\textit{ADJP$|$ADVP}} $\$+$ (\underline{/,/} $\$++$ VP)))} \\ \hline
  \textsc{Extracted Sentence} & \cellcolor{Peach!25}{\textit{``This'' + \textsc{Be} + \textbf{ADJP$|$ADVP}.}} \\ \hline
    
    \textsc{Example Input} & \textit{\textbf{Meanwhile}}\underline{,} unemployment in France threw skilled workers down to the level of the proletariat. \\ \hline
    
    \textsc{Semantic Hierarchy} &
    \vspace{0.25cm}
    \begin{minipage}[c]{\linewidth}
\hspace{1.8cm}
\begin{tikzpicture}[scale=0.9, level distance=2.4cm, sibling distance=2cm, every tree node/.style={align=center, transform shape}]
\Tree [
      .\node [style={draw,rectangle}, fill=gray!10] {SUBORD\\\textit{Elaboration}};
            \edge node[midway, left] {core}; [.\node(a){Unemployment in France\\ threw skilled workers\\ down to the level\\ of the proletariat.};]
            \edge node[midway, right] {context}; [.\node(b){This was\\ meanwhile.};]
        ]
]
\end{tikzpicture}

\end{minipage}

    \\ \bottomrule
 
       \end{tabular}
  \caption{\textsc{Rule \#33}: Preposed adjectival and adverbial phrases offset by commas.}
  \end{subtable}
  \par\bigskip
  
  \end{table}
  
  \begin{table}[!htb]
\ContinuedFloat
  
  \begin{subtable}[h]{\textwidth}
  \centering
    \footnotesize
     \begin{tabular}{  p{2cm} | p{11.5cm} }
     \toprule
                       
    
     \textsc{Tregex Pattern} & \cellcolor{Peach!50}{ROOT $<<:$ (S $<$ VP \& $<<$ (\underline{/,/} $\$+$ (\textbf{\textit{ADJP$|$ADVP}} $?\$+$ \underline{/,/})))} \\ \hline
  \textsc{Extracted Sentence} & \cellcolor{Peach!25}{\textit{``This'' + \textsc{Be} +  \textbf{ADJP$|$ADVP}.}} \\ \hline
    
    \textsc{Example Input} & Gustafsson lived a normal life until 2004\underline{,} \textit{\textbf{almost 44 years after the event}}.\\ \hline
    
    \textsc{Semantic Hierarchy} &
    \vspace{0.25cm}
    \begin{minipage}[c]{\linewidth}
\hspace{2cm}
\begin{tikzpicture}[scale=0.9, level distance=2.4cm, sibling distance=2cm, every tree node/.style={align=center, transform shape}]
\Tree [
      .\node [style={draw,rectangle}, fill=gray!10] {SUBORD\\\textit{Elaboration}};
            \edge node[midway, left] {core}; [.\node(a){Gustafsson lived\\ a normal life\\ until 2004.};]
            \edge node[midway, right] {context}; [.\node(b){This was almost\\ 44 years after\\ the event.};]
        ]
]
\end{tikzpicture}

\end{minipage}

    \\ \bottomrule

    \end{tabular}
    \caption{\textsc{Rule \#34}: Postposed and embedded adjectival and adverbial phrases offset by commas.}
    \end{subtable}

  \caption[The transformation rule patterns addressing adjectival and adverbial phrases.]{The transformation rule patterns addressing adjectival and adverbial phrases. The pattern in bold represents the part of the input that is extracted from the source and rephrased into a self-contained sentence. The underlined pattern (solid line) will be deleted from the remaining part of the input. The italic pattern will be labelled as a context sentence.}
  \label{examplePatterns_adjectival}

\end{table}

Note that our TS approach is limited to extracting adjectival and adverbial phrases that are offset by commas. Sentences that contain attributive adjectives or adverbial modifiers that are not separated through punctuation from the main clause (e.g., \textit{``He has a \textbf{red} car.''}, \textit{``I \textbf{usually} go to the lake.''}) are not simplified, as the underlying sentence structure typically already presents a regular subject-predicate-object order. Therefore, the sentence is already easy to process, outweighing the risk of introducing errors when attempting to further simplify the input by transforming the adjective or adverb into a self-contained sentence.

Whenever one of the patterns from Table \ref{examplePatterns_adjectival} matches a sentence's phrasal parse tree, the adjectival or adverbial phrase, respectively, is extracted from the input and turned into a stand-alone simplified sentence by prepending the canoncial phrase \textit{``This is/was''}. As it commonly expresses a piece of background information, it is labelled as a context sentence. Containing the key information of the input, the remaining part of the source receives a core tag. Aside from setting up a contextual hierarchy between the split elements, a semantic relationship between them is established. For this purpose, the decomposed sentences are connected with an ``Elaboration'' relation, which is selected from the classes of rhetorical relations due to the fact that adjectival and adverbial phrases usually provide additional details about the event described in the respective main clause.

\subsubsection*{Lead Noun Phrases}

Occasionally, sentences may start with an inserted noun phrase, which generally indicates a temporal expression. Hence, such a phrase usually represents background information that can be eliminated from the main sentence without resulting in a lack of key information. This is achieved by applying the transformation rule specified in Table \ref{tab:rule_leadNP}. Since the information expressed in the leading noun phrase commonly represents a piece of minor background information, the resulting extraction is labelled as a context sentence, while the remaining part from the input is tagged as a core sentence. The two simplified components are then linked through a ``Temporal'' relation. In that way, a semantic hierarchy is established between the decomposed sentences. 


\begin{table}[!htb]
\footnotesize
\centering
  \begin{tabular}{ p{2cm} | p{11.5cm} }
    \toprule
  
  \textsc{Tregex Pattern} & \cellcolor{Peach!50}{ROOT $<<:$ (S $<,$ (\textit{\textbf{NP-TMP|NP}} $\$+$ \underline{(/,/} $\$+$ NP \& $\$++$ VP)))} \\ \hline
  \textsc{Extracted Sentence} & \cellcolor{Peach!25}{\textit{``This'' + \textsc{Be} + \textbf{NP-TMP|NP}.}} \\ \hline
    
    \textsc{Example Input} & \textit{\textbf{Six days later}}\underline{,} NATO took over leadership of the effort.\\ \hline
    
    \textsc{Semantic Hierarchy} &
     
    \vspace{0.25cm}
    \begin{minipage}[c]{\linewidth}
\hspace{1.8cm}
\begin{tikzpicture}[scale=0.9, level distance=2.4cm, sibling distance=2cm, every tree node/.style={align=center, transform shape}]
\Tree [
      .\node [style={draw,rectangle}, fill=gray!10] {SUBORD\\\textit{Temporal}};
            \edge node[midway, left] {core}; [.\node(a){NATO took over\\ leadership of the effort.};]
            \edge node[midway, right] {context}; [.\node(b){This was six\\ days later.};]
        ]
]
\end{tikzpicture}

\end{minipage}

    \\
    \bottomrule
    \end{tabular}
  
  \caption[The transformation rule pattern for splitting and rephrasing leading noun phrases.]{The transformation rule pattern for splitting and rephrasing leading noun phrases (\textsc{Rule \#35}). The pattern in bold represents the part of the input that is extracted from the source and turned into a self-contained sentence. The underlined pattern (solid line) will be deleted from the remaining part of the input. The italic pattern will be labelled as a context sentence.}
  \label{tab:rule_leadNP}

\end{table}

\section*{Appendix B. Annotation Guidelines for the Human Evaluation Task}
\label{app:annotation_guidelines}
Table \ref{fig:guidelines} details the guidelines that were given to the annotators for the human evaluation task.

\begin{table}[!htb]
\centering
\small
\begin{tabular}{l}
  \hline
  \begin{minipage}{13cm}
  \vskip 4pt
    \begin{itemize}
    \item[1.] \textbf{\textsc{Grammaticality (G)}: Is the output fluent and grammatical?}
    \begin{itemize}
        \item[\textbf{5}] The output is meaningful and there are no grammatical mistakes.
        \item[\textbf{4}] One or two minor errors, but the meaning can be easily understood.
        \item[\textbf{3}] Several errors, but it is still possible to understand the meaning.
        \item[\textbf{2}] It is hard to understand the meaning due to many grammatical errors.
        \item[\textbf{1}] The output is so ungrammatical that it is impossible to infer any meaning.
    \end{itemize}
    \end{itemize}
    \vskip 4pt
    \end{minipage}
    \\ \hline
    \begin{minipage}{13cm}
  \vskip 4pt
    \begin{itemize}
    \item[2.] \textbf{\textsc{Meaning Preservation (M)}: Does the output preserve the meaning of the input?}
    \begin{itemize}
        \item[\textbf{5}] The simplification has the same meaning as the original sentence. No piece of information from the input is lost.
        \item[\textbf{4}] \begin{enumerate}
            \item The core meaning is the same as in the original sentence but with some subtle differences.
            \item If there are several output sentences, one is missing some important piece of information from the original sentence but the others fully preserve the original meaning.
        \end{enumerate}
        \item[\textbf{3}] \begin{enumerate}
            \item The simplified sentence(s) contain(s) a part of the relevant information from the original sentence, but another important part of relevant information is missing (there is no spurious information).
            \item Incorrect pronoun/attachment resolution.
        \end{enumerate} 
        \item[\textbf{2}] The simplification has the opposite or very different meaning compared to the original sentence.
        \item[\textbf{1}] Simplified sentence(s) have no coherent meaning (i.e. it is not possible to compare it/them to the original sentence).
    \end{itemize} 
    \end{itemize}
    \vskip 4pt
    \end{minipage}
    \\ \hline
    \begin{minipage}{13cm}
  \vskip 4pt
    \begin{itemize}
    \item[3.] \textbf{\textsc{Structural Simplicity (S)}: Is the output simpler than the input, \textit{ignoring the complexity of the words}?}
    \begin{itemize}
        \item[\textbf{2}] The simplified output is \textit{much simpler} than the original sentence.
        \item[\textbf{1}] The simplified output is \textit{somewhat simpler} than the original sentence.
        \item[\textbf{0}] The simplified output is \textit{equally difficult} as the original sentence.
        \item[\textbf{-1}] The simplified output is \textit{somewhat more difficult} than the original sentence.
        \item[\textbf{-2}] The simplified output is \textit{much more difficult} than the original sentence.
    \end{itemize} 
 \end{itemize}
 \vskip 4pt
 \end{minipage}
 \\ \hline
\end{tabular}
\caption{Annotation guidelines for the human evaluation task.}
\label{fig:guidelines}
\end{table}

\section*{Appendix C. Mapping of Cue Phrases to Rhetorical Relations}
\label{appendix:mapping_cue_phrases}

Table \ref{app:mapping_cue_phrases} lists the full set of cue phrases that serve as lexical features for the identification of rhetorical relations 
when establishing the semantic hierarchy between a pair of split sentences. 
 It further shows to which rhetorical relation each of them is mapped.

\begin{table}[!htb]
\centering
\begin{tabular}{l | p{11.0cm}}
\toprule
    \textsc{Rhet. Relation} & \textsc{Cue Phrases} \\ \hline\hline
     \textbf{Contrast} & although, but, but now, despite, even though, even when, except when, however, instead, rather, still, though, thus, until recently, while, yet \\ \hline
     \textbf{List} & and, in addition, in addition to, moreover \\ \hline
     \textbf{Disjunction} & or \\ \hline
     \textbf{Cause} & largely because, because, since \\ \hline
     \textbf{Result} & as a result, as a result of \\ \hline
     \textbf{Temporal} & after, and after, next, then, before, previously \\ \hline
     \textbf{Background} & as, now, once, when, with, without \\ \hline
     \textbf{Condition} & if, in case, unless, until \\ \hline
     \textbf{Elaboration} & more provocatively, even before, for example, further, recently, since, since now, so, so far, where, whereby, whether \\ \hline
     \textbf{Explanation} & simply because, because of, indeed, so, so that \\
     \bottomrule
\end{tabular}

\caption{Mapping of cue phrases to rhetorical relations.}
\label{app:mapping_cue_phrases}
\end{table}

In addition, \textit{Spatial} and \textit{Temporal} relationships are identified on the basis of named entities, while \textit{Attribution} relations are detected using a pre-defined list of verbs of reported speech and cognition \cite{carlson2001discourse}.

Furthermore, in some cases, the type of relationship that is set between two decomposed spans is selected based on syntactic information. This applies to the following rhetorical relations:

\begin{itemize}
    \item \textit{Purpose} (in case of adverbial clauses of purpose, lexicalized on the preposition \textit{``to''}),
    \item \textit{Elaboration} (in case of appositives, adjectival/adverbial phrases, participial phrases without an adverbial connector and relative clauses that are \textit{not} introduced by the relative pronoun \textit{``where''}),
    \item \textit{Spatial} (in case of relative clauses commencing with the relative pronoun \textit{``where''}) and
    \item \textit{Temporal} (in case of lead noun phrases).
\end{itemize}

\section*{Appendix D. Error Analysis}
\label{app:error_classification}

Using an example sentence, Table \ref{ErrorClassification} illustrates the six types of errors that were identified through qualitative analysis of the simplification patterns.

\begin{table}[!htb]
\small
\centering
  \begin{tabular}{  c | p{2.5cm} || p{5cm} |p{5cm} }
    \toprule
    \textsc{Type} & \textsc{Error} & \textsc{Input} &  \textsc{Output} \\ \bottomrule
    1 & additional parts & The largest populations of Mennonites are in the Democratic Republic of Congo and the United States. &  The largest populations of Mennonites are in the Democratic Republic of Congo. The largest populations of Mennonites are in \textbf{the Democratic Republic of} the United States. \\\hline
   2 & missing parts & Chun was later pardoned by Kim Young-sam on the advice of \textbf{then} President-elect Kim Dae-jung. & Chun was later pardoned by Kim Young-sam on the advice of Kim Dae-jung. Kim Dae-jung was President-elect. \\ \hline
    3 & morpho\-logical errors & Mars has two Moons, Phobos and Deimos. & Mars has two Moons. Phobos and Deimos \textbf{is} two moons. \\ \hline
    4 & wrong split point & Peter Ermakov killed her \textbf{with a gun shot to the left side of her head}. &  Peter Ermakov killed her. \textbf{This was to the left side of her head. This was with a gun shot.} \\\hline
    5 &  wrong referent & Pushkin and \textbf{his wife Natalya Goncharova, whom he married in 1831}, later became regulars of the court society. &  Pushkin and his wife Natalya Goncharova later became regulars of the court society. He married \textbf{Pushkin and his wife Natalya Goncharova} in 1831.\\ \hline
    6 &  wrong order of the syntactic elements & She is the adoptive mother of actor Dylan McDermott, whom she adopted when he was 18. & She is the adoptive mother of actor Dylan McDermott. \textbf{Dylan McDermott she} adopted when he was 18. \\ \bottomrule
    
  \end{tabular} 
  
  \caption{Error classification.}
  \label{ErrorClassification}
\end{table}

\section*{Appendix E. Step-by-Step Example}

In the following, a step-by-step example is given, illustrating the transformation stage on the following complex source sentence:

\begin{quote}
    \textit{``A fluoroscopic study which is known as an upper gastrointestinal series is typically the next step in management, although if volvulus is suspected, caution with non water soluble contrast is mandatory as the usage of barium can impede surgical revision and lead to increased post operative complications.''}
\end{quote}

With the help of the 35 hand-crafted grammar rules, 
the input is recursively transformed into a discourse tree, i.e. a set of hierarchically ordered and semantically interconnected sentences that present a simplified syntax.

\vspace{1cm}

(1) Initialization of the discourse tree:
\begin{figure}[H]
\centering
\begin{tikzpicture}[scale=0.8, level distance=2.4cm, sibling distance=0cm, every tree node/.style={align=center, transform shape}]
\Tree [
  .\node(a){A fluoroscopic study which is known as an upper gastrointestinal series\\ is typically the next step in management, although if volvulus is suspected, \\caution with non water soluble contrast is mandatory as the usage of barium can\\ impede surgical revision and lead to increased post operative complications.};
]
\end{tikzpicture}
\end{figure}

(2) Rule application on the marked leaf:
\begin{figure}[H]
\centering
\begin{tikzpicture}[scale=0.8, level distance=2.4cm, sibling distance=0cm, every tree node/.style={align=center, transform shape}]
\Tree [
  .\node[fill={rgb:orange,1;yellow,2;pink,5}](a){A fluoroscopic study which is known as an upper gastrointestinal series\\ is typically the next step in management, although if volvulus is suspected, \\caution with non water soluble contrast is mandatory as the usage of barium can\\ impede surgical revision and lead to increased post operative complications.};
]
\end{tikzpicture}
\end{figure}

(3) Result of the rule application:
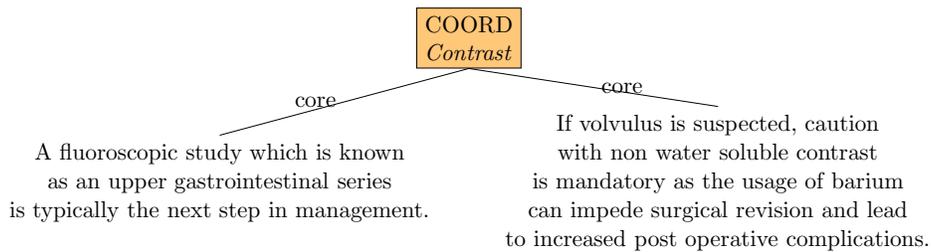
\begin{figure}[H]
\centering
\begin{tikzpicture}[scale=0.8, level distance=2.4cm, sibling distance=1cm, every tree node/.style={align=center, transform shape}]
\Tree [
      .\node [style={draw,rectangle}, fill={rgb:orange,1;yellow,2;pink,5}] {COORD\\\textit{Contrast}};
            \edge node[midway, left] {core}; [.\node(a){A fluoroscopic study which is known\\ as an upper gastrointestinal series\\ is typically the next step in management.};]
            \edge node[midway, right] {core}; [.\node(b){If volvulus is suspected, caution\\ with non water soluble contrast\\ is mandatory as the usage of barium\\ can impede surgical revision and lead\\ to increased post operative complications.};]
        ]
]
\end{tikzpicture}
\end{figure}

(4) Rule application on the marked leaf:
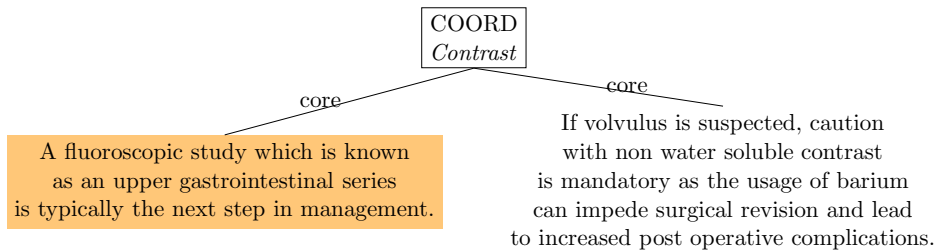
\begin{figure}[H]
\centering
\begin{tikzpicture}[scale=0.8, level distance=2.4cm, sibling distance=1cm, every tree node/.style={align=center, transform shape}]
\Tree [
      .\node [style={draw,rectangle}] {COORD\\\textit{Contrast}};
            \edge node[midway, left] {core}; [.\node[fill={rgb:orange,1;yellow,2;pink,5}](a){A fluoroscopic study which is known\\ as an upper gastrointestinal series\\ is typically the next step in management.};]
            \edge node[midway, right] {core}; [.\node(b){If volvulus is suspected, caution\\ with non water soluble contrast\\ is mandatory as the usage of barium\\ can impede surgical revision and lead\\ to increased post operative complications.};]
        ]
  
]
\end{tikzpicture}
\end{figure}


(5) Result of the rule application:
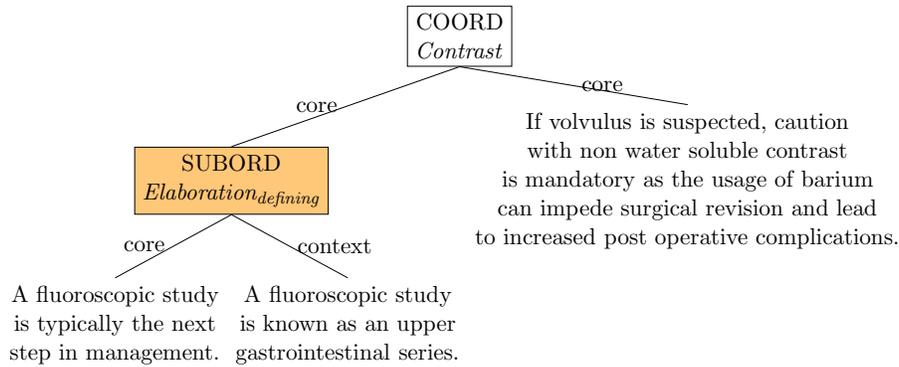
\begin{figure}[H]
\centering
\begin{tikzpicture}[scale=0.8, level distance=2.4cm, sibling distance=0cm, every tree node/.style={align=center, transform shape}]
\Tree [
      .\node [style={draw,rectangle}] {COORD\\\textit{Contrast}};
            \edge node[midway, left] {core}; [
                .\node [style={draw,rectangle}, fill={rgb:orange,1;yellow,2;pink,5}] {SUBORD\\\textit{Elaboration\textsubscript{defining}}};
                    \edge node[midway, left] {core}; [.\node(a){A fluoroscopic study\\ is typically the next \\step in management.};]
                    \edge node[midway, right] {context}; [.\node(b){A fluoroscopic study\\ is known as an upper\\ gastrointestinal series.};]
            ]
            \edge node[midway, right] {core}; [.\node(c){If volvulus is suspected, caution\\ with non water soluble contrast\\ is mandatory as the usage of barium\\ can impede surgical revision and lead\\ to increased post operative complications.};]
        ]
]
\end{tikzpicture}
\end{figure}

(6) Rule application on the marked leaf:
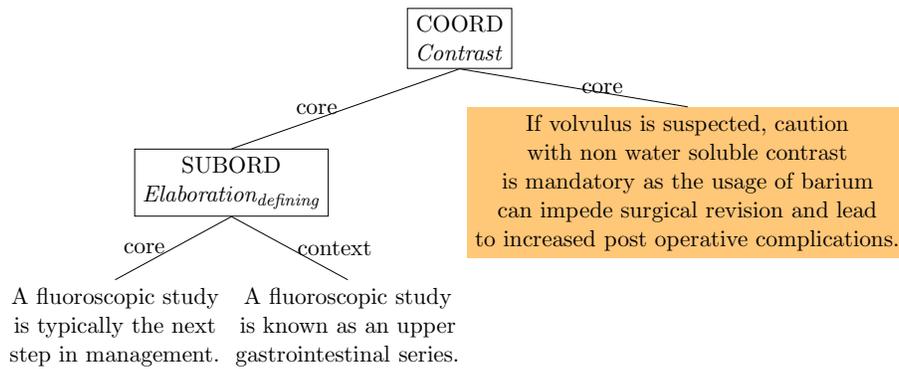
\begin{figure}[H]
\centering
\begin{tikzpicture}[scale=0.8, level distance=2.4cm, sibling distance=0cm, every tree node/.style={align=center, transform shape}]
\Tree [
      .\node [style={draw,rectangle}] {COORD\\\textit{Contrast}};
            \edge node[midway, left] {core}; [
                .\node [style={draw,rectangle}] {SUBORD\\\textit{Elaboration\textsubscript{defining}}};
                    \edge node[midway, left] {core}; [.\node(a){A fluoroscopic study\\ is typically the next \\step in management.};]
                    \edge node[midway, right] {context}; [.\node(b){A fluoroscopic study\\ is known as an upper\\ gastrointestinal series.};]
            ]
            \edge node[midway, right] {core}; [.\node[fill={rgb:orange,1;yellow,2;pink,5}](c){If volvulus is suspected, caution\\ with non water soluble contrast\\ is mandatory as the usage of barium\\ can impede surgical revision and lead\\ to increased post operative complications.};]
        ]
]
\end{tikzpicture}
\end{figure}

(7) Result of the rule application
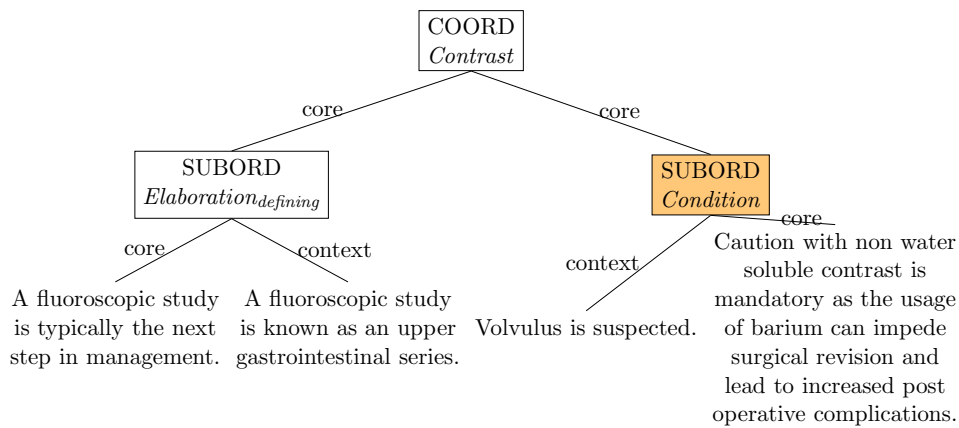
\begin{figure}[H]
\centering
\begin{tikzpicture}[scale=0.8, level distance=2.4cm, sibling distance=0cm, every tree node/.style={align=center, transform shape}]
\Tree [
      .\node [style={draw,rectangle}] {COORD\\\textit{Contrast}};
            \edge node[midway, left] {core}; [
                .\node [style={draw,rectangle}] {SUBORD\\\textit{Elaboration\textsubscript{defining}}};
                    \edge node[midway, left] {core}; [.\node(a){A fluoroscopic study\\ is typically the next \\step in management.};]
                    \edge node[midway, right] {context}; [.\node(b){A fluoroscopic study\\ is known as an upper\\ gastrointestinal series.};]
            ]
            \edge node[midway, right] {core}; [
                .\node [style={draw,rectangle}, fill={rgb:orange,1;yellow,2;pink,5}] {SUBORD\\\textit{Condition}};
                    \edge node[midway, left] {context}; [.\node(c){Volvulus is suspected.};]
                    \edge node[midway, right] {core}; [.\node(d){Caution with non water\\ soluble contrast is\\ mandatory as the usage\\ of barium can impede\\ surgical revision and\\ lead to increased post\\ operative complications.};]
            ]
        ]
]
\end{tikzpicture}
\end{figure}


(8) Rule application on the marked leaf:
\begin{figure}[!ht]
\centering
\begin{tikzpicture}[scale=0.8, level distance=2.4cm, sibling distance=0cm, every tree node/.style={align=center, transform shape}]
\Tree [
       .\node [style={draw,rectangle}] {COORD\\\textit{Contrast}};
            \edge node[midway, left] {core}; [
                .\node [style={draw,rectangle}] {SUBORD\\\textit{Elaboration\textsubscript{defining}}};
                    \edge node[midway, left] {core}; [.\node(a){A fluoroscopic study\\ is typically the next \\step in management.};]
                    \edge node[midway, right] {context}; [.\node(b){A fluoroscopic study\\ is known as an upper\\ gastrointestinal series.};]
            ]
            \edge node[midway, right] {core}; [
                .\node [style={draw,rectangle}] {SUBORD\\\textit{Condition}};
                    \edge node[midway, left] {context}; [.\node(c){Volvulus is suspected.};]
                    \edge node[midway, right] {core}; [.\node[fill={rgb:orange,1;yellow,2;pink,5}](d){Caution with non water\\ soluble contrast is\\ mandatory as the usage\\ of barium can impede\\ surgical revision and\\ lead to increased post\\ operative complications.};]
            ]
        ]
]
\end{tikzpicture}
\end{figure}

(9) Result of the rule application:
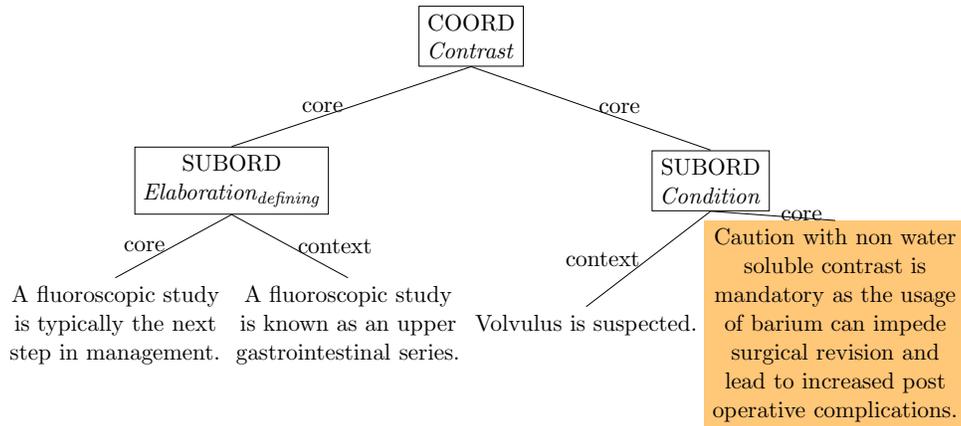
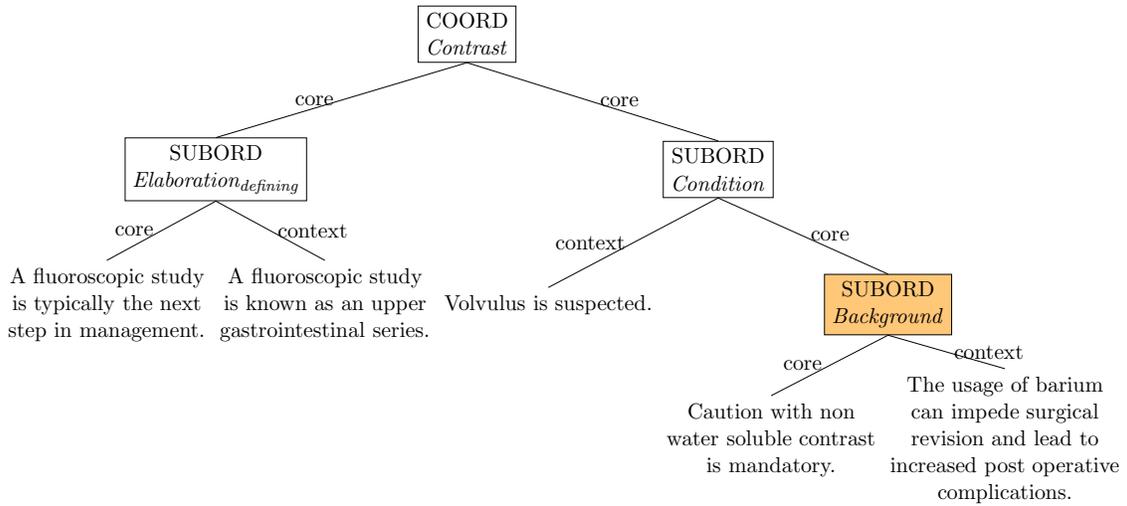
\begin{figure}[H]
\centering
\begin{tikzpicture}[scale=0.75, level distance=2.4cm, sibling distance=0cm, every tree node/.style={align=center, transform shape}]
\Tree [
      .\node [style={draw,rectangle}] {COORD\\\textit{Contrast}};
            \edge node[midway, left] {core}; [
                .\node [style={draw,rectangle}] {SUBORD\\\textit{Elaboration\textsubscript{defining}}};
                    \edge node[midway, left] {core}; [.\node(a){A fluoroscopic study\\ is typically the next \\step in management.};]
                    \edge node[midway, right] {context}; [.\node(b){A fluoroscopic study\\ is known as an upper\\ gastrointestinal series.};]
            ]
            \edge node[midway, right] {core}; [
                .\node [style={draw,rectangle}] {SUBORD\\\textit{Condition}};
                     \edge node[midway, left] {context}; [.\node(c){Volvulus is suspected.};]
                    \edge node[midway, right] {core};[.\node [style={draw,rectangle}, fill={rgb:orange,1;yellow,2;pink,5}] {SUBORD\\\textit{Background}};
                        \edge node[midway, left] {core}; [.\node(d){Caution with non\\ water soluble contrast\\ is mandatory.};]
                        \edge node[midway, right] {context}; [.\node(e){The usage of barium \\can impede surgical\\ revision and lead to\\ increased post operative\\ complications.};]
                        ]
            ]
        ]
]
\end{tikzpicture}
\end{figure}


(10) Rule application on the marked leaf:
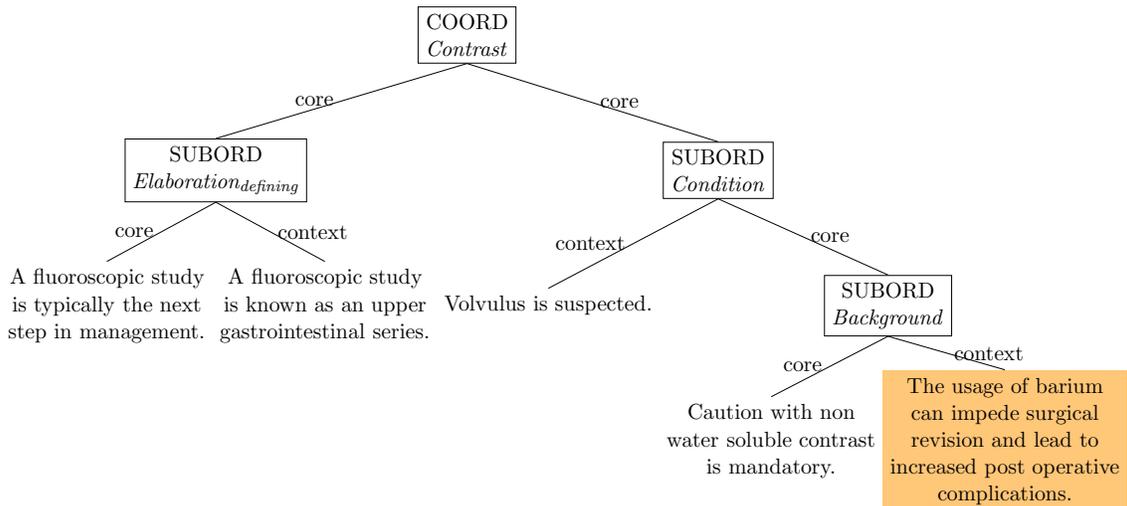
\begin{figure}[!ht]
\centering
\begin{tikzpicture}[scale=0.75, level distance=2.4cm, sibling distance=0cm, every tree node/.style={align=center, transform shape}]
\Tree [
      .\node [style={draw,rectangle}] {COORD\\\textit{Contrast}};
            \edge node[midway, left] {core}; [
                .\node [style={draw,rectangle}] {SUBORD\\\textit{Elaboration\textsubscript{defining}}};
                    \edge node[midway, left] {core}; [.\node(a){A fluoroscopic study\\ is typically the next \\step in management.};]
                    \edge node[midway, right] {context}; [.\node(b){A fluoroscopic study\\ is known as an upper\\ gastrointestinal series.};]
            ]
            \edge node[midway, right] {core}; [
                .\node [style={draw,rectangle}] {SUBORD\\\textit{Condition}};
                     \edge node[midway, left] {context}; [.\node(c){Volvulus is suspected.};]
                    \edge node[midway, right] {core};[.\node [style={draw,rectangle}] {SUBORD\\\textit{Background}};
                        \edge node[midway, left] {core}; [.\node(d){Caution with non\\ water soluble contrast\\ is mandatory.};]
                        \edge node[midway, right] {context}; [.\node[fill={rgb:orange,1;yellow,2;pink,5}](e){The usage of barium \\can impede surgical\\ revision and lead to\\ increased post operative\\ complications.};]
                        ]
            ]
        ]
]
\end{tikzpicture}
\end{figure}

(11) Result of the rule application:
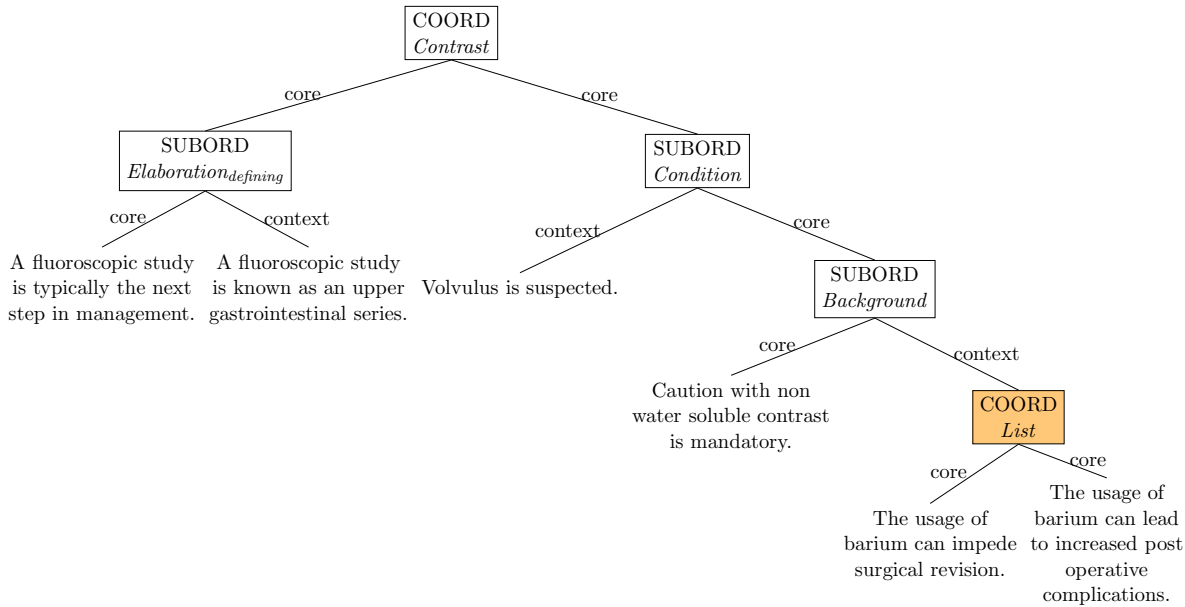
\begin{figure}[!ht]
\centering
\begin{tikzpicture}[scale=0.71, level distance=2.4cm, sibling distance=0cm, every tree node/.style={align=center, transform shape}]
\Tree [
      .\node [style={draw,rectangle}] {COORD\\\textit{Contrast}};
            \edge node[midway, left] {core}; [
                .\node [style={draw,rectangle}] {SUBORD\\\textit{Elaboration\textsubscript{defining}}};
                    \edge node[midway, left] {core}; [.\node(a){A fluoroscopic study\\ is typically the next \\step in management.};]
                    \edge node[midway, right] {context}; [.\node(b){A fluoroscopic study\\ is known as an upper\\ gastrointestinal series.};]
            ]
            \edge node[midway, right] {core}; [
                .\node [style={draw,rectangle}] {SUBORD\\\textit{Condition}};
                     \edge node[midway, left] {context}; [.\node(c){Volvulus is suspected.};]
                    \edge node[midway, right] {core};[.\node [style={draw,rectangle}] {SUBORD\\\textit{Background}};
                        \edge node[midway, left] {core}; [.\node(d){Caution with non\\ water soluble contrast\\ is mandatory.};]
                        \edge node[midway, right] {context}; [.\node [style={draw,rectangle}, fill={rgb:orange,1;yellow,2;pink,5}] {COORD\\\textit{List}};
                            \edge node[midway, left] {core}; [.\node(e){The usage of\\ barium can impede\\ surgical revision.};]
                            \edge node[midway, right] {core}; [.\node(f){The usage of\\ barium can lead\\ to increased post\\ operative\\ complications.};]
                        ]
                    ]
            ]
        ]
]
\end{tikzpicture}
\caption{Final discourse tree of the example sentence.}
\label{example_full_final}
\end{figure}

(12) STOP. No more transformation rule pattern matches.

\vspace{1cm}

When no more transformation pattern matches, the algorithm stops. The final discourse tree for our running example is shown in Figure \ref{example_full_final}.

\section*{Appendix F. Analysis of the Impact of Splitting-only Transformations on SARI}
\label{app:sari}

SARI is the arithmetic mean of n-gram precisions and recalls for the operations \textit{add, keep} and \textit{delete}. It rewards systems for \textit{adding} n-grams that occur in the reference simplifications (but not in the input), for \textit{keeping} n-grams that occur both in the output and the references, and for not over-\textit{deleting} n-grams \cite{alva-manchego2020survey}.

    TS systems that are limited to sentence splitting operations neither add nor delete any n-grams from the input (apart from duplicating phrases when splitting, for instance, coordinate verb or noun phrases, or deleting function words such as relative pronouns when extracting relative clauses). Instead, all input words are typically kept and decomposed into simplified sentences.

    To compute SARI, five sub-scores need to be calculated:
    \begin{enumerate}
        \item precision\textsubscript{\textit{add}}: The precision of \textit{add} is typically high in the case of sentence splitting only, as in most cases no words are added in the simplified output. In some cases, words from the input are duplicated to reconstruct grammatically sound sentences, e.g. when splitting coordinate verb or noun phrases. However, no new words are introduced.
        \item recall\textsubscript{\textit{add}}: The recall of \textit{add} depends on the references. The more words and phrases that do not appear in the input are added in the references, the lower the score will be, when restricted to sentence splitting transformations.
        \item precision\textsubscript{\textit{keep}}: The precision of \textit{keep} depends on the references, too. The fewer input words and phrases are kept in the references, the worse the score will be.
        \item recall\textsubscript{\textit{keep}}: As we do not delete in the sentence splitting setting, the recall of \textit{keep} is typically high. All words that are kept in the references will appear in the simplified output as well, apart from some function words such as conjunctions or relative pronouns.
        \item precision\textsubscript{\textit{delete}}: The precision of \textit{delete} is typically high, since in the case of splitting-only transformations no deletions are performed. (Note that SARI does not consider the recall of \textit{delete}. Instead, for deletion, it only uses precision because over-deleting hurts readability much more significantly than not deleting \cite{xu2016optimizing}.)
    \end{enumerate}

    Hence, when performing sentence splitting operations only, three out of five sub-scores will be high a priori and might therefore distort the resulting SARI score. 
    The sub-scores are better suited to capture various aspects that influence systems' performance when carrying out lexical simplifications and deletion operations. In this case, the SARI score better reflects systems' performance. 

\vskip 0.2in
\bibliography{test}

\begin{thebibliography}{}

\bibitem[\protect\BCAY{Abraham}{Abraham}{1985}]{abraham1985field}
Abraham, R.~G. \BBOP1985\BBCP.
\newblock \BBOQ Field independence-dependence and the teaching of grammar\BBCQ\
\newblock {\Bem Tesol Quarterly}, {\Bem 19\/}(4), 689--702.

\bibitem[\protect\BCAY{Aharoni\ \BBA\ Goldberg}{Aharoni\ \BBA\
  Goldberg}{2018}]{aharoni2018split}
Aharoni, R.\BBACOMMA\  \BBA\ Goldberg, Y. \BBOP2018\BBCP.
\newblock \BBOQ Split and rephrase: Better evaluation and stronger
  baselines\BBCQ\
\newblock In {\Bem Proceedings of the 56th Annual Meeting of the Association
  for Computational Linguistics (Volume 2: Short Papers)}, \BPGS\ 719--724.
  Association for Computational Linguistics.

\bibitem[\protect\BCAY{Alva-Man\-chego, Scarton,\ \BBA\ Specia}{Alva-Man\-chego
  et~al.}{2020}]{alva-manchego2020survey}
Alva-Man\-chego, F., Scarton, C., \BBA\ Specia, L. \BBOP2020\BBCP.
\newblock \BBOQ Data-driven sentence simplification: Survey and benchmark\BBCQ\
\newblock {\Bem Computational Linguistics}, {\Bem 46\/}(1), 135--187.

\bibitem[\protect\BCAY{Angeli, Johnson~Premkumar,\ \BBA\ Manning}{Angeli
  et~al.}{2015}]{Angeli15}
Angeli, G., Johnson~Premkumar, M.~J., \BBA\ Manning, C.~D. \BBOP2015\BBCP.
\newblock \BBOQ Leveraging linguistic structure for open domain information
  extraction\BBCQ\
\newblock In {\Bem Proceedings of the 53rd Annual Meeting of the Association
  for Computational Linguistics and the 7th International Joint Conference on
  Natural Language Processing (Volume 1: Long Papers)}, \BPGS\ 344--354,
  Beijing, China. Association for Computational Linguistics.

\bibitem[\protect\BCAY{Angrosh, Nomoto,\ \BBA\ Siddharthan}{Angrosh
  et~al.}{2014}]{angrosh-etal-2014-lexico}
Angrosh, M., Nomoto, T., \BBA\ Siddharthan, A. \BBOP2014\BBCP.
\newblock \BBOQ Lexico-syntactic text simplification and compression with typed
  dependencies\BBCQ\
\newblock In {\Bem Proceedings of {COLING} 2014, the 25th International
  Conference on Computational Linguistics: Technical Papers}, \BPGS\
  1996--2006, Dublin, Ireland. Dublin City University and Association for
  Computational Linguistics.

\bibitem[\protect\BCAY{Bahdanau, Cho,\ \BBA\ Bengio}{Bahdanau
  et~al.}{2015}]{bahdanau2014neural}
Bahdanau, D., Cho, K., \BBA\ Bengio, Y. \BBOP2015\BBCP.
\newblock \BBOQ Neural machine translation by jointly learning to align and
  translate\BBCQ\
\newblock In Bengio, Y.\BBACOMMA\  \BBA\ LeCun, Y.\BEDS, {\Bem 3rd
  International Conference on Learning Representations, {ICLR} 2015, San Diego,
  CA, USA, May 7-9, 2015, Conference Track Proceedings}.

\bibitem[\protect\BCAY{Bailey, Lierler,\ \BBA\ Susman}{Bailey
  et~al.}{2015}]{PP2015}
Bailey, D., Lierler, Y., \BBA\ Susman, B. \BBOP2015\BBCP.
\newblock \BBOQ Prepositional phrase attachment problem revisited: how verbnet
  can help\BBCQ\
\newblock In {\Bem Proceedings of the 11th International Conference on
  Computational Semantics, {IWCS} 2015, 15-17 April, 2015, Queen Mary
  University of London, London, {UK}}, \BPGS\ 12--22.

\bibitem[\protect\BCAY{Banko, Cafarella, Soderland, Broadhead,\ \BBA\
  Etzioni}{Banko et~al.}{2007}]{Banko07}
Banko, M., Cafarella, M.~J., Soderland, S., Broadhead, M., \BBA\ Etzioni, O.
  \BBOP2007\BBCP.
\newblock \BBOQ Open information extraction from the web\BBCQ\
\newblock In {\Bem Proceedings of the 20th International Joint Conference on
  Artifical Intelligence}, \BPGS\ 2670--2676, San Francisco, CA, USA. Morgan
  Kaufmann Publishers Inc.

\bibitem[\protect\BCAY{Bast\ \BBA\ Haussmann}{Bast\ \BBA\
  Haussmann}{2013}]{bast2013open}
Bast, H.\BBACOMMA\  \BBA\ Haussmann, E. \BBOP2013\BBCP.
\newblock \BBOQ Open information extraction via contextual sentence
  decomposition\BBCQ\
\newblock In {\Bem 2013 IEEE Seventh International Conference on Semantic
  Computing}, \BPGS\ 154--159. IEEE.

\bibitem[\protect\BCAY{Benamara\ \BBA\ Taboada}{Benamara\ \BBA\
  Taboada}{2015}]{benamara2015mapping}
Benamara, F.\BBACOMMA\  \BBA\ Taboada, M. \BBOP2015\BBCP.
\newblock \BBOQ Mapping different rhetorical relation annotations: A
  proposal\BBCQ\
\newblock In {\Bem Proceedings of the Fourth Joint Conference on Lexical and
  Computational Semantics}, \BPGS\ 147--152.

\bibitem[\protect\BCAY{Bernhard, De~Viron, Moriceau,\ \BBA\ Tannier}{Bernhard
  et~al.}{2012}]{bernhard2012question}
Bernhard, D., De~Viron, L., Moriceau, V., \BBA\ Tannier, X. \BBOP2012\BBCP.
\newblock \BBOQ Question generation for french: collating parsers and
  paraphrasing questions\BBCQ\
\newblock {\Bem Dialogue \& Discourse}, {\Bem 3\/}(2), 43--74.

\bibitem[\protect\BCAY{Botha, Faruqui, Alex, Baldridge,\ \BBA\ Das}{Botha
  et~al.}{2018}]{Botha2018}
Botha, J.~A., Faruqui, M., Alex, J., Baldridge, J., \BBA\ Das, D.
  \BBOP2018\BBCP.
\newblock \BBOQ Learning to split and rephrase from wikipedia edit
  history\BBCQ\
\newblock In {\Bem Proceedings of the 2018 Conference on Empirical Methods in
  Natural Language Processing}, \BPGS\ 732--737. Association for Computational
  Linguistics.

\bibitem[\protect\BCAY{Bouayad-Agha, Casamayor, Ferraro, Mille, Vidal,\ \BBA\
  Wanner}{Bouayad-Agha et~al.}{2009}]{bouayad2009improving}
Bouayad-Agha, N., Casamayor, G., Ferraro, G., Mille, S., Vidal, V., \BBA\
  Wanner, L. \BBOP2009\BBCP.
\newblock \BBOQ Improving the comprehension of legal documentation: the case of
  patent claims\BBCQ\
\newblock In {\Bem Proceedings of the 12th International Conference on
  Artificial Intelligence and Law}, \BPGS\ 78--87. ACM.

\bibitem[\protect\BCAY{Brinton}{Brinton}{2000}]{Brin00}
Brinton, L.~J. \BBOP2000\BBCP.
\newblock {\Bem The Structure of Modern English: A linguistic introduction}.
\newblock John Benjamins B.V., Amsterdam, The Netherlands.

\bibitem[\protect\BCAY{Carlson\ \BBA\ Marcu}{Carlson\ \BBA\
  Marcu}{2001}]{carlson2001discourse}
Carlson, L.\BBACOMMA\  \BBA\ Marcu, D. \BBOP2001\BBCP.
\newblock \BBOQ Discourse tagging reference manual\BBCQ\
\newblock {\Bem ISI Technical Report ISI-TR-545}, {\Bem 54}, 56.

\bibitem[\protect\BCAY{Carlson, Okurowski,\ \BBA\ Marcu}{Carlson
  et~al.}{2002}]{carlson2002rst}
Carlson, L., Okurowski, M.~E., \BBA\ Marcu, D. \BBOP2002\BBCP.
\newblock {\Bem RST discourse treebank}.
\newblock Linguistic Data Consortium, University of Pennsylvania.

\bibitem[\protect\BCAY{Carroll, Minnen, Pearce, Canning, Devlin,\ \BBA\
  Tait}{Carroll et~al.}{1999}]{carroll1999simplifying}
Carroll, J., Minnen, G., Pearce, D., Canning, Y., Devlin, S., \BBA\ Tait, J.
  \BBOP1999\BBCP.
\newblock \BBOQ Simplifying text for language-impaired readers\BBCQ\
\newblock In {\Bem Ninth Conference of the European Chapter of the Association
  for Computational Linguistics}.

\bibitem[\protect\BCAY{Cetto, Niklaus, Freitas,\ \BBA\ Handschuh}{Cetto
  et~al.}{2018}]{cetto2018graphene}
Cetto, M., Niklaus, C., Freitas, A., \BBA\ Handschuh, S. \BBOP2018\BBCP.
\newblock \BBOQ Graphene: Semantically-linked propositions in open information
  extraction\BBCQ\
\newblock In {\Bem Proceedings of the 27th International Conference on
  Computational Linguistics}, \BPGS\ 2300--2311. Association for Computational
  Linguistics.

\bibitem[\protect\BCAY{Chandrasekar, Doran,\ \BBA\ Srinivas}{Chandrasekar
  et~al.}{1996}]{Chandrasekar1996}
Chandrasekar, R., Doran, C., \BBA\ Srinivas, B. \BBOP1996\BBCP.
\newblock \BBOQ Motivations and methods for text simplification\BBCQ\
\newblock In {\Bem Proceedings of the 16th Conference on Computational
  Linguistics - Volume 2}, COLING '96, \BPGS\ 1041--1044, Stroudsburg, PA, USA.
  Association for Computational Linguistics.

\bibitem[\protect\BCAY{Cohen}{Cohen}{1968}]{cohen1968weighted}
Cohen, J. \BBOP1968\BBCP.
\newblock \BBOQ Weighted kappa: Nominal scale agreement provision for scaled
  disagreement or partial credit.\BBCQ\
\newblock {\Bem Psychological bulletin}, {\Bem 70\/}(4), 213.

\bibitem[\protect\BCAY{Coster\ \BBA\ Kauchak}{Coster\ \BBA\
  Kauchak}{2011}]{Coster:2011:SEW:2002736.2002865}
Coster, W.\BBACOMMA\  \BBA\ Kauchak, D. \BBOP2011\BBCP.
\newblock \BBOQ Simple english wikipedia: A new text simplification task\BBCQ\
\newblock In {\Bem Proceedings of the 49th Annual Meeting of the Association
  for Computational Linguistics: Human Language Technologies: Short Papers -
  Volume 2}, HLT '11, \BPGS\ 665--669, Stroudsburg, PA, USA. Association for
  Computational Linguistics.

\bibitem[\protect\BCAY{De~Belder\ \BBA\ Moens}{De~Belder\ \BBA\
  Moens}{2010}]{debelder2010text}
De~Belder, J.\BBACOMMA\  \BBA\ Moens, M.-F. \BBOP2010\BBCP.
\newblock \BBOQ Text simplification for children\BBCQ\
\newblock In {\Bem Prroceedings of the SIGIR workshop on accessible search
  systems}, \BPGS\ 19--26. ACM.

\bibitem[\protect\BCAY{Del~Corro\ \BBA\ Gemulla}{Del~Corro\ \BBA\
  Gemulla}{2013}]{DelCorro13}
Del~Corro, L.\BBACOMMA\  \BBA\ Gemulla, R. \BBOP2013\BBCP.
\newblock \BBOQ Clausie: Clause-based open information extraction\BBCQ\
\newblock In {\Bem Proceedings of the 22Nd International Conference on World
  Wide Web}, \BPGS\ 355--366, New York, NY, USA. ACM.

\bibitem[\protect\BCAY{Evans, Or{\u{a}}san,\ \BBA\ Dornescu}{Evans
  et~al.}{2014}]{evans-etal-2014-evaluation}
Evans, R., Or{\u{a}}san, C., \BBA\ Dornescu, I. \BBOP2014\BBCP.
\newblock \BBOQ An evaluation of syntactic simplification rules for people with
  autism\BBCQ\
\newblock In {\Bem Proceedings of the 3rd Workshop on Predicting and Improving
  Text Readability for Target Reader Populations ({PITR})}, \BPGS\ 131--140,
  Gothenburg, Sweden. Association for Computational Linguistics.

\bibitem[\protect\BCAY{Evans\ \BBA\ Or\v{a}san}{Evans\ \BBA\
  Or\v{a}san}{2019}]{evansorasan2019}
Evans, R.\BBACOMMA\  \BBA\ Or\v{a}san, C. \BBOP2019\BBCP.
\newblock \BBOQ Identifying signs of syntactic complexity for rule-based
  sentence simplification\BBCQ\
\newblock {\Bem Natural Language Engineering}, {\Bem 25\/}(1), 69–119.

\bibitem[\protect\BCAY{Evans}{Evans}{2011}]{evans2011comparing}
Evans, R.~J. \BBOP2011\BBCP.
\newblock \BBOQ Comparing methods for the syntactic simplification of sentences
  in information extraction\BBCQ\
\newblock {\Bem Literary and linguistic computing}, {\Bem 26\/}(4), 371--388.

\bibitem[\protect\BCAY{Fader, Soderland,\ \BBA\ Etzioni}{Fader
  et~al.}{2011}]{Fader11}
Fader, A., Soderland, S., \BBA\ Etzioni, O. \BBOP2011\BBCP.
\newblock \BBOQ Identifying relations for open information extraction\BBCQ\
\newblock In {\Bem Proceedings of the 2011 Conference on Empirical Methods in
  Natural Language Processing}, \BPGS\ 1535--1545, Edinburgh, Scotland, UK.
  Association for Computational Linguistics.

\bibitem[\protect\BCAY{Fay}{Fay}{1990}]{collinsgrammar}
Fay, R.\BED. \BBOP1990\BBCP.
\newblock {\Bem Collins Cobuild English Grammar}.
\newblock Collins.

\bibitem[\protect\BCAY{Feng\ \BBA\ Hirst}{Feng\ \BBA\
  Hirst}{2014}]{feng2014linear}
Feng, V.~W.\BBACOMMA\  \BBA\ Hirst, G. \BBOP2014\BBCP.
\newblock \BBOQ A linear-time bottom-up discourse parser with constraints and
  post-editing.\BBCQ\
\newblock In {\Bem ACL (1)}, \BPGS\ 511--521.

\bibitem[\protect\BCAY{Ferr{\'e}s, Marimon, Saggion,\ \BBA\
  AbuRa'ed}{Ferr{\'e}s et~al.}{2016}]{Ferres2016}
Ferr{\'e}s, D., Marimon, M., Saggion, H., \BBA\ AbuRa'ed, A. \BBOP2016\BBCP.
\newblock \BBOQ Yats: Yet another text simplifier\BBCQ\
\newblock In M{\'e}tais, E., Meziane, F., Saraee, M., Sugumaran, V., \BBA\
  Vadera, S.\BEDS, {\Bem Natural Language Processing and Information Systems},
  \BPGS\ 335--342, Cham. Springer International Publishing.

\bibitem[\protect\BCAY{Filippova\ \BBA\ Strube}{Filippova\ \BBA\
  Strube}{2008}]{filippova-strube-2008-sentence}
Filippova, K.\BBACOMMA\  \BBA\ Strube, M. \BBOP2008\BBCP.
\newblock \BBOQ Sentence fusion via dependency graph compression\BBCQ\
\newblock In {\Bem Proceedings of the 2008 Conference on Empirical Methods in
  Natural Language Processing}, \BPGS\ 177--185, Honolulu, Hawaii. Association
  for Computational Linguistics.

\bibitem[\protect\BCAY{Fisher\ \BBA\ Roark}{Fisher\ \BBA\
  Roark}{2007}]{fisher2007utility}
Fisher, S.\BBACOMMA\  \BBA\ Roark, B. \BBOP2007\BBCP.
\newblock \BBOQ The utility of parse-derived features for automatic discourse
  segmentation\BBCQ\
\newblock In {\Bem Annual Meeting-Association for Computational Linguistics},
  \lowercase{\BVOL}~45, \BPG\ 488.

\bibitem[\protect\BCAY{Gardent, Shimorina, Narayan,\ \BBA\
  Perez-Beltrachini}{Gardent et~al.}{2017}]{gardent-etal-2017-webnlg}
Gardent, C., Shimorina, A., Narayan, S., \BBA\ Perez-Beltrachini, L.
  \BBOP2017\BBCP.
\newblock \BBOQ The {W}eb{NLG} challenge: Generating text from {RDF} data\BBCQ\
\newblock In {\Bem Proceedings of the 10th International Conference on Natural
  Language Generation}, \BPGS\ 124--133, Santiago de Compostela, Spain.
  Association for Computational Linguistics.

\bibitem[\protect\BCAY{Gelbukh\ \BBA\ Calvo}{Gelbukh\ \BBA\
  Calvo}{2018}]{Gelbukh2018}
Gelbukh, A.\BBACOMMA\  \BBA\ Calvo, H. \BBOP2018\BBCP.
\newblock {\Bem Prepositional phrase attachment disambiguation}, \BPGS\
  85--110.
\newblock Studies in Computational Intelligence. Springer Verlag, Alemania.

\bibitem[\protect\BCAY{Glava{\v{s}}\ \BBA\ {\v{S}}tajner}{Glava{\v{s}}\ \BBA\
  {\v{S}}tajner}{2015}]{glavas2015}
Glava{\v{s}}, G.\BBACOMMA\  \BBA\ {\v{S}}tajner, S. \BBOP2015\BBCP.
\newblock \BBOQ Simplifying lexical simplification: Do we need simplified
  corpora?\BBCQ\
\newblock In {\Bem Proceedings of the 53rd Annual Meeting of the Association
  for Computational Linguistics and the 7th International Joint Conference on
  Natural Language Processing (Volume 2: Short Papers)}, \BPGS\ 63--68.
  Association for Computational Linguistics.

\bibitem[\protect\BCAY{Gu, Lu, Li,\ \BBA\ Li}{Gu et~al.}{2016}]{Gu2016}
Gu, J., Lu, Z., Li, H., \BBA\ Li, V.~O. \BBOP2016\BBCP.
\newblock \BBOQ Incorporating copying mechanism in sequence-to-sequence
  learning\BBCQ\
\newblock In {\Bem Proceedings of the 54th Annual Meeting of the Association
  for Computational Linguistics (Volume 1: Long Papers)}, \BPGS\ 1631--1640.
  Association for Computational Linguistics.

\bibitem[\protect\BCAY{Heilman\ \BBA\ Smith}{Heilman\ \BBA\
  Smith}{2010}]{heilman2010extracting}
Heilman, M.\BBACOMMA\  \BBA\ Smith, N.~A. \BBOP2010\BBCP.
\newblock \BBOQ Extracting simplified statements for factual question
  generation\BBCQ\
\newblock In {\Bem Proceedings of QG2010: The Third Workshop on Question
  Generation}, \lowercase{\BVOL}~11.

\bibitem[\protect\BCAY{Hernault, Prendinger, Ishizuka, et~al.}{Hernault
  et~al.}{2010}]{hernault2010hilda}
Hernault, H., Prendinger, H., Ishizuka, M., et~al. \BBOP2010\BBCP.
\newblock \BBOQ Hilda: A discourse parser using support vector machine
  classification\BBCQ\
\newblock {\Bem Dialogue \& Discourse}, {\Bem 1\/}(3).

\bibitem[\protect\BCAY{Hershcovich, Abend,\ \BBA\ Rappoport}{Hershcovich
  et~al.}{2017}]{hershcovich2017a}
Hershcovich, D., Abend, O., \BBA\ Rappoport, A. \BBOP2017\BBCP.
\newblock \BBOQ A transition-based directed acyclic graph parser for ucca\BBCQ\
\newblock In {\Bem Proc. of ACL}, \BPGS\ 1127--1138.

\bibitem[\protect\BCAY{Inui, Fujita, Takahashi, Iida,\ \BBA\ Iwakura}{Inui
  et~al.}{2003}]{Inui:2003:TSR:1118984.1118986}
Inui, K., Fujita, A., Takahashi, T., Iida, R., \BBA\ Iwakura, T.
  \BBOP2003\BBCP.
\newblock \BBOQ Text simplification for reading assistance: A project
  note\BBCQ\
\newblock In {\Bem Proceedings of the Second International Workshop on
  Paraphrasing - Volume 16}, PARAPHRASE '03, \BPGS\ 9--16, Stroudsburg, PA,
  USA. Association for Computational Linguistics.

\bibitem[\protect\BCAY{Jonnalagadda, Tari, Hakenberg, Baral,\ \BBA\
  Gonzalez}{Jonnalagadda et~al.}{2009}]{jonnalagadda2009towards}
Jonnalagadda, S., Tari, L., Hakenberg, J., Baral, C., \BBA\ Gonzalez, G.
  \BBOP2009\BBCP.
\newblock \BBOQ Towards effective sentence simplification for automatic
  processing of biomedical text\BBCQ\
\newblock In {\Bem Proceedings of Human Language Technologies: The 2009 Annual
  Conference of the North American Chapter of the Association for Computational
  Linguistics, Companion Volume: Short Papers}, \BPGS\ 177--180. Association
  for Computational Linguistics.

\bibitem[\protect\BCAY{Joty, Carenini,\ \BBA\ Ng}{Joty
  et~al.}{2015}]{joty2015codra}
Joty, S., Carenini, G., \BBA\ Ng, R.~T. \BBOP2015\BBCP.
\newblock \BBOQ {CODRA}: A novel discriminative framework for rhetorical
  analysis\BBCQ\
\newblock {\Bem Computational Linguistics}, {\Bem 41\/}(3), 385--435.

\bibitem[\protect\BCAY{Kamp}{Kamp}{1981}]{Kamp1981-KAMATO-2}
Kamp, H. \BBOP1981\BBCP.
\newblock \BBOQ A theory of truth and semantic representation, 277-322, jag
  groenendijk, tmv janssen and mbj stokhof, eds\BBCQ\
\newblock In Groenendijk, J.\BED, {\Bem Formal Methods in the Study of
  Language}. U of Amsterdam.

\bibitem[\protect\BCAY{Kibrik\ \BBA\ Krasavina}{Kibrik\ \BBA\
  Krasavina}{2005}]{kibrik2005corpus}
Kibrik, A.~A.\BBACOMMA\  \BBA\ Krasavina, O.~N. \BBOP2005\BBCP.
\newblock \BBOQ A corpus study of referential choice: The role of rhetorical
  structure\BBCQ.
\newblock Citeseer.

\bibitem[\protect\BCAY{Knott\ \BBA\ Dale}{Knott\ \BBA\
  Dale}{1994}]{knott1994using}
Knott, A.\BBACOMMA\  \BBA\ Dale, R. \BBOP1994\BBCP.
\newblock \BBOQ Using linguistic phenomena to motivate a set of coherence
  relations\BBCQ\
\newblock {\Bem Discourse processes}, {\Bem 18\/}(1), 35--62.

\bibitem[\protect\BCAY{Lafferty, McCallum,\ \BBA\ Pereira}{Lafferty
  et~al.}{2001}]{Lafferty2001}
Lafferty, J.~D., McCallum, A., \BBA\ Pereira, F. C.~N. \BBOP2001\BBCP.
\newblock \BBOQ Conditional random fields: Probabilistic models for segmenting
  and labeling sequence data\BBCQ\
\newblock In {\Bem Proceedings of the Eighteenth International Conference on
  Machine Learning}, ICML ’01, \BPG\ 282–289, San Francisco, CA, USA.
  Morgan Kaufmann Publishers Inc.

\bibitem[\protect\BCAY{Levy\ \BBA\ Andrew}{Levy\ \BBA\ Andrew}{2006}]{Levy2006}
Levy, R.\BBACOMMA\  \BBA\ Andrew, G. \BBOP2006\BBCP.
\newblock \BBOQ Tregex and tsurgeon: tools for querying and manipulating tree
  data structures\BBCQ\
\newblock In {\Bem Proceedings of the fifth international conference on
  Language Resources and Evaluation}, \BPGS\ 2231--2234.

\bibitem[\protect\BCAY{Li, Li,\ \BBA\ Hovy}{Li et~al.}{2014}]{li2014recursive}
Li, J., Li, R., \BBA\ Hovy, E.~H. \BBOP2014\BBCP.
\newblock \BBOQ Recursive deep models for discourse parsing.\BBCQ\
\newblock In {\Bem EMNLP}, \BPGS\ 2061--2069.

\bibitem[\protect\BCAY{Mallinson\ \BBA\ Lapata}{Mallinson\ \BBA\
  Lapata}{2019}]{mallinson2019controllable}
Mallinson, J.\BBACOMMA\  \BBA\ Lapata, M. \BBOP2019\BBCP.
\newblock \BBOQ Controllable sentence simplification: Employing syntactic and
  lexical constraints\BBCQ.

\bibitem[\protect\BCAY{Mann\ \BBA\ Thompson}{Mann\ \BBA\
  Thompson}{1988}]{mann1988rhetorical}
Mann, W.~C.\BBACOMMA\  \BBA\ Thompson, S.~A. \BBOP1988\BBCP.
\newblock \BBOQ Rhetorical structure theory: Toward a functional theory of text
  organization\BBCQ\
\newblock {\Bem Text-Interdisciplinary Journal for the Study of Discourse},
  {\Bem 8\/}(3), 243--281.

\bibitem[\protect\BCAY{Marcu}{Marcu}{1997}]{marcu1997rhetorical}
Marcu, D. \BBOP1997\BBCP.
\newblock \BBOQ The rhetorical parsing of natural language texts\BBCQ\
\newblock In {\Bem Proceedings of the 35th Annual Meeting of the Association
  for Computational Linguistics and Eighth Conference of the European Chapter
  of the Association for Computational Linguistics}, \BPGS\ 96--103.
  Association for Computational Linguistics.

\bibitem[\protect\BCAY{Marcu}{Marcu}{2000}]{marcu2000rhetorical}
Marcu, D. \BBOP2000\BBCP.
\newblock \BBOQ The rhetorical parsing of unrestricted texts: A surface-based
  approach\BBCQ\
\newblock {\Bem Computational linguistics}, {\Bem 26\/}(3), 395--448.

\bibitem[\protect\BCAY{Martinet\ \BBA\ Thomson}{Martinet\ \BBA\
  Thomson}{1996}]{martinet1996practical}
Martinet, A.\BBACOMMA\  \BBA\ Thomson, A. \BBOP1996\BBCP.
\newblock {\Bem A Practical English Grammar. 4th edition}.
\newblock Oxford University Press.

\bibitem[\protect\BCAY{Mausam}{Mausam}{2016}]{Mausam16}
Mausam \BBOP2016\BBCP.
\newblock \BBOQ Open information extraction systems and downstream
  applications\BBCQ\
\newblock In {\Bem Proceedings of the Twenty-Fifth International Joint
  Conference on Artificial Intelligence, {IJCAI} 2016, New York, NY, USA, 9-15
  July 2016}, \BPGS\ 4074--4077.

\bibitem[\protect\BCAY{Mausam, Schmitz, Soderland, Bart,\ \BBA\ Etzioni}{Mausam
  et~al.}{2012}]{Mausam12}
Mausam, Schmitz, M., Soderland, S., Bart, R., \BBA\ Etzioni, O. \BBOP2012\BBCP.
\newblock \BBOQ Open language learning for information extraction\BBCQ\
\newblock In {\Bem Proceedings of the 2012 Joint Conference on Empirical
  Methods in Natural Language Processing and Computational Natural Language
  Learning}, \BPGS\ 523--534, Jeju Island, Korea. Association for Computational
  Linguistics.

\bibitem[\protect\BCAY{Miltsakaki, Prasad, Joshi,\ \BBA\ Webber}{Miltsakaki
  et~al.}{2004}]{miltsakaki2004penn}
Miltsakaki, E., Prasad, R., Joshi, A.~K., \BBA\ Webber, B.~L. \BBOP2004\BBCP.
\newblock \BBOQ The penn discourse treebank.\BBCQ\
\newblock In {\Bem LREC}.

\bibitem[\protect\BCAY{Mitkov\ \BBA\ Saggion}{Mitkov\ \BBA\
  Saggion}{2018}]{SaggionsurveyTextSimplification}
Mitkov, R.\BBACOMMA\  \BBA\ Saggion, H. \BBOP2018\BBCP.
\newblock \BBOQ Text simplification\BBCQ.

\bibitem[\protect\BCAY{Miwa, S{\ae}tre, Miyao,\ \BBA\ Tsujii}{Miwa
  et~al.}{2010}]{Miwa2010simplificationRE}
Miwa, M., S{\ae}tre, R., Miyao, Y., \BBA\ Tsujii, J. \BBOP2010\BBCP.
\newblock \BBOQ Entity-focused sentence simplification for relation
  extraction\BBCQ\
\newblock In {\Bem Proceedings of the 23rd International Conference on
  Computational Linguistics (Coling 2010)}, \BPGS\ 788--796. Coling 2010
  Organizing Committee.

\bibitem[\protect\BCAY{Narayan\ \BBA\ Gardent}{Narayan\ \BBA\
  Gardent}{2014}]{narayan2014hybrid}
Narayan, S.\BBACOMMA\  \BBA\ Gardent, C. \BBOP2014\BBCP.
\newblock \BBOQ Hybrid simplification using deep semantics and machine
  translation\BBCQ\
\newblock In {\Bem Proceedings of the 52nd Annual Meeting of the Association
  for Computational Linguistics (Volume 1: Long Papers)}, \lowercase{\BVOL}~1,
  \BPGS\ 435--445.

\bibitem[\protect\BCAY{Narayan\ \BBA\ Gardent}{Narayan\ \BBA\
  Gardent}{2016}]{Narayan2016}
Narayan, S.\BBACOMMA\  \BBA\ Gardent, C. \BBOP2016\BBCP.
\newblock \BBOQ Unsupervised sentence simplification using deep semantics\BBCQ\
\newblock In {\Bem Proceedings of the 9th International Natural Language
  Generation conference}, \BPGS\ 111--120. Association for Computational
  Linguistics.

\bibitem[\protect\BCAY{Narayan, Gardent, Cohen,\ \BBA\ Shimorina}{Narayan
  et~al.}{2017}]{Narayan2017}
Narayan, S., Gardent, C., Cohen, S.~B., \BBA\ Shimorina, A. \BBOP2017\BBCP.
\newblock \BBOQ Split and rephrase\BBCQ\
\newblock In {\Bem Proceedings of the 2017 Conference on Empirical Methods in
  Natural Language Processing}, \BPGS\ 606--616. Association for Computational
  Linguistics.

\bibitem[\protect\BCAY{Niklaus, Cetto, Freitas,\ \BBA\ Handschuh}{Niklaus
  et~al.}{2018}]{niklaus-etal-2018-survey}
Niklaus, C., Cetto, M., Freitas, A., \BBA\ Handschuh, S. \BBOP2018\BBCP.
\newblock \BBOQ A survey on open information extraction\BBCQ\
\newblock In {\Bem Proceedings of the 27th International Conference on
  Computational Linguistics}, \BPGS\ 3866--3878, Santa Fe, New Mexico, USA.
  Association for Computational Linguistics.

\bibitem[\protect\BCAY{Niklaus, Cetto, Freitas,\ \BBA\ Handschuh}{Niklaus
  et~al.}{2019a}]{niklaus-etal-2019-dissim}
Niklaus, C., Cetto, M., Freitas, A., \BBA\ Handschuh, S. \BBOP2019a\BBCP.
\newblock \BBOQ {D}is{S}im: A discourse-aware syntactic text simplification
  framework for {E}nglish and {G}erman\BBCQ\
\newblock In {\Bem Proceedings of the 12th International Conference on Natural
  Language Generation}, \BPGS\ 504--507, Tokyo, Japan. Association for
  Computational Linguistics.

\bibitem[\protect\BCAY{Niklaus, Cetto, Freitas,\ \BBA\ Handschuh}{Niklaus
  et~al.}{2019b}]{niklaus-etal-2019-transforming}
Niklaus, C., Cetto, M., Freitas, A., \BBA\ Handschuh, S. \BBOP2019b\BBCP.
\newblock \BBOQ Transforming complex sentences into a semantic hierarchy\BBCQ\
\newblock In {\Bem Proceedings of the 57th Annual Meeting of the Association
  for Computational Linguistics}, \BPGS\ 3415--3427, Florence, Italy.
  Association for Computational Linguistics.

\bibitem[\protect\BCAY{Niklaus, Cetto, Freitas,\ \BBA\ Handschuh}{Niklaus
  et~al.}{2023}]{niklaus2023canonical}
Niklaus, C., Cetto, M., Freitas, A., \BBA\ Handschuh, S. \BBOP2023\BBCP.
\newblock \BBOQ A canonical context-preserving representation for open ie:
  Extracting semantically typed relational tuples from complex sentences\BBCQ\
\newblock {\Bem Know.-Based Syst.}, {\Bem 268\/}(C).

\bibitem[\protect\BCAY{Niklaus, Freitas,\ \BBA\ Handschuh}{Niklaus
  et~al.}{2022}]{niklaus-etal-2022-shallow}
Niklaus, C., Freitas, A., \BBA\ Handschuh, S. \BBOP2022\BBCP.
\newblock \BBOQ Shallow discourse parsing for open information extraction and
  text simplification\BBCQ\
\newblock In {\Bem Proceedings of the 3rd Workshop on Computational Approaches
  to Discourse}, \BPGS\ 64--76, Gyeongju, Republic of Korea and Online.
  International Conference on Computational Linguistics.

\bibitem[\protect\BCAY{Nisioi, {\v{S}}tajner, Ponzetto,\ \BBA\ Dinu}{Nisioi
  et~al.}{2017}]{nisioi2017exploring}
Nisioi, S., {\v{S}}tajner, S., Ponzetto, S.~P., \BBA\ Dinu, L.~P.
  \BBOP2017\BBCP.
\newblock \BBOQ Exploring neural text simplification models\BBCQ\
\newblock In {\Bem Proceedings of the 55th Annual Meeting of the Association
  for Computational Linguistics (Volume 2: Short Papers)}, \lowercase{\BVOL}~2,
  \BPGS\ 85--91.

\bibitem[\protect\BCAY{Paetzold\ \BBA\ Specia}{Paetzold\ \BBA\
  Specia}{2016}]{Paetzold:2016:ULS:3016387.3016433}
Paetzold, G.~H.\BBACOMMA\  \BBA\ Specia, L. \BBOP2016\BBCP.
\newblock \BBOQ Unsupervised lexical simplification for non-native
  speakers\BBCQ\
\newblock In {\Bem Proceedings of the Thirtieth AAAI Conference on Artificial
  Intelligence}, AAAI'16, \BPGS\ 3761--3767. AAAI Press.

\bibitem[\protect\BCAY{Papineni, Roukos, Ward,\ \BBA\ Zhu}{Papineni
  et~al.}{2002}]{papineni2002bleu}
Papineni, K., Roukos, S., Ward, T., \BBA\ Zhu, W.-J. \BBOP2002\BBCP.
\newblock \BBOQ Bleu: a method for automatic evaluation of machine
  translation\BBCQ\
\newblock In {\Bem Proceedings of the 40th annual meeting on association for
  computational linguistics}, \BPGS\ 311--318. Association for Computational
  Linguistics.

\bibitem[\protect\BCAY{Prasad, Miltsakaki, Dinesh, Lee, Joshi, Robaldo,\ \BBA\
  Webber}{Prasad et~al.}{2007}]{prasad2007penn}
Prasad, R., Miltsakaki, E., Dinesh, N., Lee, A., Joshi, A., Robaldo, L., \BBA\
  Webber, B.~L. \BBOP2007\BBCP.
\newblock \BBOQ The penn discourse treebank 2.0 annotation manual\BBCQ.

\bibitem[\protect\BCAY{Quirk, Greenbaum, Leech,\ \BBA\ Svartvik}{Quirk
  et~al.}{1985}]{Quir85}
Quirk, R., Greenbaum, S., Leech, G., \BBA\ Svartvik, J. \BBOP1985\BBCP.
\newblock {\Bem A Comprehensive Grammar of the English Language}.
\newblock Longman, London.

\bibitem[\protect\BCAY{Saggion, \v{S}tajner, Bott, Mille, Rello,\ \BBA\
  Drndarevic}{Saggion et~al.}{2015}]{Saggion:2015:MSI:2775084.2738046}
Saggion, H., \v{S}tajner, S., Bott, S., Mille, S., Rello, L., \BBA\ Drndarevic,
  B. \BBOP2015\BBCP.
\newblock \BBOQ Making it simplext: Implementation and evaluation of a text
  simplification system for spanish\BBCQ\
\newblock {\Bem ACM Trans. Access. Comput.}, {\Bem 6\/}(4), 14:1--14:36.

\bibitem[\protect\BCAY{Saha\ \BBA\ Mausam}{Saha\ \BBA\
  Mausam}{2018}]{Swarnadeep2018}
Saha, S.\BBACOMMA\  \BBA\ Mausam \BBOP2018\BBCP.
\newblock \BBOQ Open information extraction from conjunctive sentences\BBCQ\
\newblock In {\Bem Proceedings of the 27th International Conference on
  Computational Linguistics}, \BPGS\ 2288--2299. Association for Computational
  Linguistics.

\bibitem[\protect\BCAY{Schrimpf}{Schrimpf}{2018}]{schrimpf-2018-using}
Schrimpf, N.~M. \BBOP2018\BBCP.
\newblock \BBOQ Using rhetorical topics for automatic summarization\BBCQ\
\newblock In {\Bem Proceedings of the Society for Computation in Linguistics
  ({SC}i{L}) 2018}, \BPGS\ 125--135.

\bibitem[\protect\BCAY{See, Liu,\ \BBA\ Manning}{See et~al.}{2017}]{See2017}
See, A., Liu, P.~J., \BBA\ Manning, C.~D. \BBOP2017\BBCP.
\newblock \BBOQ Get to the point: Summarization with pointer-generator
  networks\BBCQ\
\newblock In {\Bem Proceedings of the 55th Annual Meeting of the Association
  for Computational Linguistics (Volume 1: Long Papers)}, \BPGS\ 1073--1083.
  Association for Computational Linguistics.

\bibitem[\protect\BCAY{Shardlow}{Shardlow}{2014}]{shardlow2014survey}
Shardlow, M. \BBOP2014\BBCP.
\newblock \BBOQ A survey of automated text simplification\BBCQ\
\newblock {\Bem International Journal of Advanced Computer Science and
  Applications}, {\Bem 4\/}(1), 58--70.

\bibitem[\protect\BCAY{Siddharthan}{Siddharthan}{2002}]{siddharthan2002architecture}
Siddharthan, A. \BBOP2002\BBCP.
\newblock \BBOQ An architecture for a text simplification system\BBCQ\
\newblock In {\Bem Language Engineering Conference, 2002. Proceedings}, \BPGS\
  64--71. IEEE.

\bibitem[\protect\BCAY{Siddharthan}{Siddharthan}{2006}]{siddharthan2006syntactic}
Siddharthan, A. \BBOP2006\BBCP.
\newblock \BBOQ Syntactic simplification and text cohesion\BBCQ\
\newblock {\Bem Research on Language and Computation}, {\Bem 4\/}(1), 77--109.

\bibitem[\protect\BCAY{Siddharthan}{Siddharthan}{2014}]{siddharthan2014survey}
Siddharthan, A. \BBOP2014\BBCP.
\newblock \BBOQ A survey of research on text simplification\BBCQ\
\newblock {\Bem ITL-International Journal of Applied Linguistics}, {\Bem
  165\/}(2), 259--298.

\bibitem[\protect\BCAY{Siddharthan\ \BBA\ Mandya}{Siddharthan\ \BBA\
  Mandya}{2014}]{Siddharthan2014}
Siddharthan, A.\BBACOMMA\  \BBA\ Mandya, A. \BBOP2014\BBCP.
\newblock \BBOQ Hybrid text simplification using synchronous dependency
  grammars with hand-written and automatically harvested rules\BBCQ\
\newblock In {\Bem Proceedings of the 14th Conference of the European Chapter
  of the Association for Computational Linguistics}, \BPGS\ 722--731.
  Association for Computational Linguistics.

\bibitem[\protect\BCAY{Siddharthan, Nenkova,\ \BBA\ McKeown}{Siddharthan
  et~al.}{2004}]{siddharthan2004syntactic}
Siddharthan, A., Nenkova, A., \BBA\ McKeown, K. \BBOP2004\BBCP.
\newblock \BBOQ Syntactic simplification for improving content selection in
  multi-document summarization\BBCQ\
\newblock In {\Bem Proceedings of the 20th international conference on
  Computational Linguistics}, \BPG\ 896. Association for Computational
  Linguistics.

\bibitem[\protect\BCAY{Socher, Bauer, Manning,\ \BBA\ Ng}{Socher
  et~al.}{2013}]{Socher2013}
Socher, R., Bauer, J., Manning, C.~D., \BBA\ Ng, A.~Y. \BBOP2013\BBCP.
\newblock \BBOQ {Parsing With Compositional Vector Grammars}\BBCQ\
\newblock In {\Bem {ACL}}.

\bibitem[\protect\BCAY{{\v{S}}tajner\ \BBA\ Glava{\v{s}}}{{\v{S}}tajner\ \BBA\
  Glava{\v{s}}}{2017}]{stajner2017leveraging}
{\v{S}}tajner, S.\BBACOMMA\  \BBA\ Glava{\v{s}}, G. \BBOP2017\BBCP.
\newblock \BBOQ Leveraging event-based semantics for automated text
  simplification\BBCQ\
\newblock {\Bem Expert systems with applications}, {\Bem 82}, 383--395.

\bibitem[\protect\BCAY{{\v{S}}tajner\ \BBA\ Popovic}{{\v{S}}tajner\ \BBA\
  Popovic}{2016}]{stajner2016can}
{\v{S}}tajner, S.\BBACOMMA\  \BBA\ Popovic, M. \BBOP2016\BBCP.
\newblock \BBOQ Can text simplification help machine translation?\BBCQ\
\newblock In {\Bem Proceedings of the 19th Annual Conference of the European
  Association for Machine Translation}, \BPGS\ 230--242.

\bibitem[\protect\BCAY{{\v{S}}tajner\ \BBA\ Popovic}{{\v{S}}tajner\ \BBA\
  Popovic}{2018}]{stajner2018improvingMT}
{\v{S}}tajner, S.\BBACOMMA\  \BBA\ Popovic, M. \BBOP2018\BBCP.
\newblock \BBOQ Improving machine translation of english relative clauses with
  automatic text simplification\BBCQ\
\newblock In {\Bem Proceedings of the First Workshop on Automatic Text
  Adaptation (ATA)}.

\bibitem[\protect\BCAY{Stanovsky\ \BBA\ Dagan}{Stanovsky\ \BBA\
  Dagan}{2016}]{Stanovsky2016benchmark}
Stanovsky, G.\BBACOMMA\  \BBA\ Dagan, I. \BBOP2016\BBCP.
\newblock \BBOQ Creating a large benchmark for open information
  extraction\BBCQ\
\newblock In {\Bem Proceedings of the 2016 Conference on Empirical Methods in
  Natural Language Processing (EMNLP)}, \BPG\ (to appear), Austin, Texas.
  Association for Computational Linguistics.

\bibitem[\protect\BCAY{Sulem, Abend,\ \BBA\ Rappoport}{Sulem
  et~al.}{2018a}]{sulemBLEU2018}
Sulem, E., Abend, O., \BBA\ Rappoport, A. \BBOP2018a\BBCP.
\newblock \BBOQ Bleu is not suitable for the evaluation of text
  simplification\BBCQ\
\newblock In {\Bem Proceedings of the 2018 Conference on Empirical Methods in
  Natural Language Processing}, \BPGS\ 738--744. Association for Computational
  Linguistics.

\bibitem[\protect\BCAY{Sulem, Abend,\ \BBA\ Rappoport}{Sulem
  et~al.}{2018b}]{sulemsemantic}
Sulem, E., Abend, O., \BBA\ Rappoport, A. \BBOP2018b\BBCP.
\newblock \BBOQ Semantic structural evaluation for text simplification\BBCQ\
\newblock In {\Bem Proceedings of the 2018 Conference of the North American
  Chapter of the Association for Computational Linguistics: Human Language
  Technologies, Volume 1 (Long Papers)}, \BPGS\ 685--696. Association for
  Computational Linguistics.

\bibitem[\protect\BCAY{Sulem, Abend,\ \BBA\ Rappoport}{Sulem
  et~al.}{2018c}]{sulemSystem}
Sulem, E., Abend, O., \BBA\ Rappoport, A. \BBOP2018c\BBCP.
\newblock \BBOQ Simple and effective text simplification using semantic and
  neural methods\BBCQ\
\newblock In {\Bem Proceedings of the 56th Annual Meeting of the Association
  for Computational Linguistics (Volume 1: Long Papers)}, \BPGS\ 162--173.
  Association for Computational Linguistics.

\bibitem[\protect\BCAY{Suter, Ebling,\ \BBA\ Volk}{Suter
  et~al.}{2016}]{suter2016rule}
Suter, J., Ebling, S., \BBA\ Volk, M. \BBOP2016\BBCP.
\newblock \BBOQ Rule-based automatic text simplification for german\BBCQ\
\newblock In {\Bem 13th Conference on Natural Language Processing (KONVENS
  2016)}. s.n.

\bibitem[\protect\BCAY{Taboada\ \BBA\ Das}{Taboada\ \BBA\
  Das}{2013}]{Taboada13}
Taboada, M.\BBACOMMA\  \BBA\ Das, D. \BBOP2013\BBCP.
\newblock \BBOQ Annotation upon annotation: Adding signalling information to a
  corpus of discourse relations.\BBCQ\
\newblock {\Bem D\&D}, {\Bem 4\/}(2), 249--281.

\bibitem[\protect\BCAY{Vickrey\ \BBA\ Koller}{Vickrey\ \BBA\
  Koller}{2008}]{Vickrey2008}
Vickrey, D.\BBACOMMA\  \BBA\ Koller, D. \BBOP2008\BBCP.
\newblock \BBOQ Sentence simplification for semantic role labeling\BBCQ\
\newblock In {\Bem Proceedings of ACL-08: HLT}, \BPGS\ 344--352. Association
  for Computational Linguistics.

\bibitem[\protect\BCAY{Xu, Callison-Burch,\ \BBA\ Napoles}{Xu
  et~al.}{2015}]{Xu2015newsela}
Xu, W., Callison-Burch, C., \BBA\ Napoles, C. \BBOP2015\BBCP.
\newblock \BBOQ Problems in current text simplification research: New data can
  help\BBCQ\
\newblock {\Bem Transactions of the Association for Computational Linguistics},
  {\Bem 3}, 283--297.

\bibitem[\protect\BCAY{Xu, Napoles, Pavlick, Chen,\ \BBA\ Callison-Burch}{Xu
  et~al.}{2016}]{xu2016optimizing}
Xu, W., Napoles, C., Pavlick, E., Chen, Q., \BBA\ Callison-Burch, C.
  \BBOP2016\BBCP.
\newblock \BBOQ Optimizing statistical machine translation for text
  simplification\BBCQ\
\newblock {\Bem Transactions of the Association for Computational Linguistics},
  {\Bem 4}, 401--415.

\bibitem[\protect\BCAY{Zhang\ \BBA\ Lapata}{Zhang\ \BBA\
  Lapata}{2017}]{Zhang2017}
Zhang, X.\BBACOMMA\  \BBA\ Lapata, M. \BBOP2017\BBCP.
\newblock \BBOQ Sentence simplification with deep reinforcement learning\BBCQ\
\newblock In {\Bem Proceedings of the 2017 Conference on Empirical Methods in
  Natural Language Processing}, \BPGS\ 584--594. Association for Computational
  Linguistics.

\bibitem[\protect\BCAY{Zhu, Bernhard,\ \BBA\ Gurevych}{Zhu
  et~al.}{2010}]{zhu2010monolingual}
Zhu, Z., Bernhard, D., \BBA\ Gurevych, I. \BBOP2010\BBCP.
\newblock \BBOQ A monolingual tree-based translation model for sentence
  simplification\BBCQ\
\newblock In {\Bem Proceedings of the 23rd International Conference on
  Computational Linguistics}, \BPGS\ 1353--1361. Association for Computational
  Linguistics.

\end{thebibliography}
\bibliographystyle{theapa}

\end{document}